\documentclass[11pt]{article}
\usepackage{graphicx} 
\usepackage{natbib}
\usepackage{multirow}
\usepackage{caption}
\usepackage{siunitx}
\usepackage{algorithm}
\usepackage[noend]{algorithmic}
\usepackage{booktabs} 
\usepackage{fullpage}
\usepackage{authblk}
\usepackage{mathtools}
\usepackage{amsmath, amssymb, amsfonts, amsthm}
\usepackage{hyperref}
\usepackage{tikz}
\usepackage{bbding}
\usepackage{subcaption} 
\usepackage{gensymb}
\usepackage{complexity}
\usepackage{float}
\usepackage{xcolor}

\usepackage{xcolor}
\newtheorem{theorem}{Theorem}
\newtheorem{proposition}[theorem]{Proposition}
\newtheorem{definition}{Definition}

\title{Scalable Dynamic Distributed Constraint Optimization with Metareasoning and Application to Continual Satellite Operations}

\author[1,2]{Itai Zilberstein\thanks{Correspondence to \texttt{izilbers@cs.cmu.edu}. \copyright 2026 The authors. All rights reserved. Government sponsorship acknowledged.}}
\author[2]{Steve Chien}

\affil[1]{Department of Computer Science, Carnegie Mellon University, Pittsburgh, PA}
\affil[2]{Jet Propulsion Laboratory, California Institute of Technology, Pasadena, CA}

\begin{document}

\maketitle
\thispagestyle{empty}

\begin{abstract}
    \textit{Dynamic distributed constraint optimization problems (DDCOPs)} provide a general framework for coordinating autonomous agents in changing environments. However, existing DDCOP formulations do not adequately address settings where optimization and execution overlap, resources persist across time, and agents operate under limited computational and communication constraints. We extend the DDCOP model to address these challenges by introducing an execution-aware formulation, together with new algorithms and frameworks for efficiently computing solutions. We develop a general framework for metareasoning in DDCOPs, enabling agents to determine when the estimated benefit of recomputing solutions outweighs its computational cost. We further extend the \textit{neighborhood stochastic search algorithm} to the dynamic setting, introducing \textit{dynamic incremental neighborhood stochastic search (D-NSS)}, a scalable decomposition-based DDCOP algorithm that efficiently repairs previous solutions in response to problem dynamics. We apply our methods to the real-world application of large-scale satellite scheduling. Deploying autonomy to satellites requires efficient computation and communication in the face of highly dynamic environments. We demonstrate that D-NSS stabilizes to high-quality solutions, outperforming standard DDCOP baselines in solution quality, computation time, and message volume, while our metareasoning framework successfully balances resource conservation with utility. These methods will support the NASA FAME mission, the largest in-space demonstration of distributed multi-agent AI to date.   
\end{abstract}

\newpage
\setcounter{page}{1}

\section{Introduction}

\textit{Dynamic distributed constraint optimization problems (DDCOPs)} provide a general framework for continual decision making in decentralized multi-agent systems operating in changing environments. DDCOPs have been applied to a wide range of multi-agent systems, including sensor networks~\citep{zivan2015distributed}, smart grids~\citep{smartgrid}, UAV coordination~\citep{uavs-coord}, and satellite constellations~\citep{zilberstein2025decentralized}, where agents must optimize a shared objective despite possessing only local information and limited communication. As autonomous systems continue to grow in scale and complexity, DDCOPs offer a natural framework for continual distributed decision making.

Despite their broad applicability, existing DDCOP formulations do not adequately model an important class of dynamic optimization problems. Standard DDCOPs treat a dynamic problem as a sequence of static optimization problems, assuming that solving each optimization problem independently is sufficient. However, many real-world systems violate this assumption. Optimization frequently overlaps execution, meaning plans may become obsolete before they are carried out. Resources such as energy, memory, and computation persist, coupling decisions over time. Computation and communication also consume valuable resources, requiring agents to reason not only about \emph{what} decisions to make, but also \emph{when} it is worthwhile to recompute them. These challenges motivate new formulations and algorithms for dynamic distributed optimization.

Large-scale satellite constellation scheduling is one important application that embodies these challenges. Modern Earth-observing constellations consist of hundreds or thousands of spacecraft capable of autonomous onboard planning and inter-satellite communication \citep{newspace,eo1,nos2024,zilberstein-isairas-2024,dt-spaceops-2025,rabideau2025planning}. Such capabilities support observation campaigns that require time-sensitive and coordinated measurements, including measuring transient science phenomena and global monitoring of natural disasters (\textit{e.g.}, wildfires, floods, and volcanic activities). Without consistent observation and the ability to react promptly to dynamic events, key measurements of these processes will be missed. For example, to be actionable for ground responders, wildfire monitoring often requires updates every 30 minutes \citep{wildfireworkshop}. 

Unlike static scheduling problems, observation campaigns evolve over time. Scheduling and execution occur concurrently, and spacecraft operate under strict computational, communication, and memory constraints. Satellites must continually determine both how to adapt their schedules and whether replanning is worthwhile given its associated cost.

We extend the DDCOP framework to address these limitations through a new execution-aware formulation. Rather than modeling dynamic optimization as a sequence of independent, static optimization problems, we treat continual optimization as a coupled process in which the current solutions affect future execution and optimization.

\subsection{Our contributions}

Our work extends DDCOPs to model and solve continual optimization problems in which optimization and execution overlap, resources persist across time, and computation itself incurs non-negligible cost. We develop new formulations, algorithms, and theoretical frameworks that enable scalable dynamic distributed optimization while explicitly reasoning about the cost of replanning. We evaluate these methods on the challenging real-world problem of large-scale satellite constellation scheduling.

We first introduce the \textit{dynamic multi-satellite constellation observation scheduling problem (DCOSP)}, an execution-aware extension of DDCOPs, that builds on the \textit{static} model introduced for satellite scheduling in prior work~\citep{zilberstein2025decentralized}. Unlike standard DDCOPs, DCOSP explicitly models overlapping scheduling and execution horizons through a novel optimality condition based on completed rather than scheduled tasks. 

We then introduce a general framework for metareasoning in DDCOPs. Rather than assuming that agents should recompute solutions every time the environment changes, our framework enables agents to reason about when replanning is worthwhile by explicitly modeling the trade-off between solution quality and computational cost. We formalize an ideal value of replanning and develop a practical decentralized metareasoning heuristic.

Finally, we present \textit{dynamic incremental neighborhood stochastic search (D-NSS)}, an efficient decomposition-based algorithm for DDCOPs. D-NSS extends the \textit{neighborhood stochastic search (NSS)} algorithm to the dynamic setting. NSS achieves efficiency by decomposing the global problem into smaller sub-problems. D-NSS builds upon this foundation, repairing previous solutions to efficiently manage problem dynamics. To benchmark our online solutions, we also develop an omniscient algorithm to compute optimal solutions to DCOSP by reducing it to a DCOP using hindsight knowledge. 

We evaluate our methods on real-world large-scale satellite scheduling scenarios containing hundreds of agents and millions of decision variables. Our methods compute high-quality solutions while substantially reducing computation time and communication compared to representative DDCOP baselines. Notably, DCOSP and D-NSS will be leveraged in the largest in-space demonstration of multi-agent AI to date. The NASA FAME mission involves over 60 participating spacecraft that will dynamically coordinate actions to observe Earth phenomena \citep{fame-spaceops-2025}. 

A preliminary version of this work also appears as an extended abstract in the International Conference on Autonomous Agents and Multiagent Systems \citep{Zilberstein26:Large}.

\subsection{Related work}

\paragraph{Dynamic distributed constraint optimization} \textit{Distributed constraint optimization problems (DCOP)} have modeled numerous applications, including mobile sensor teams \citep{ca-dcop-mst}, smart grids \citep{smartgrid}, and satellite scheduling \citep{zilberstein2025decentralized}. However, solutions to distributed constraint optimization problems tend to be intensive in computation and communication, making deployment to agents with limited computation challenging. Solving a DCOP optimally is \NP-hard \citep{adopt}, meaning complete algorithms have exponential complexities \citep{syncbb,adopt,gershman2009asynchronous,dpop,mailler2004solving}.  Incomplete DCOP algorithms improve efficiency, yet typically rely on agents communicating with all neighboring agents in the constraint graph, resulting in large complexities when constraint graphs are fully connected \citep{mgm,dsa,maxsum,dgibs}. The  NSS algorithm, which iteratively improves sub-problem solutions, has been shown to solve static large-scale distributed satellite observation scheduling with limited computation and communication \citep{zilberstein2025decentralized}. We extend the NSS algorithm to the \textit{dynamic} DCOP setting to perform scalable and effective dynamic observation scheduling.

\textit{Dynamic distributed constraint optimization problems (DDCOP)} \citep{lass2008dynamic} extend DCOPs to capture evolving problem states.  A standard DDCOP is conceptualized as a sequence of $T$ static DCOPs where an optimal solution is obtained by solving each of the $T$ DCOPs optimally. DDCOP solutions are inherently online algorithms as a system reacts to changes. Most prior work has focused on developing dynamic variants of existing DCOP algorithms rather than reconsidering the underlying DDCOP abstraction itself~\citep{mailler2005comparing,faltings2005superstabilizing,khanna2009efficient,billiau2012sbdo,zivan2015distributed,yeoh2015incremental,dynamicdsa}. 
 
These assumptions become problematic in continual optimization domains where optimization and execution overlap. When solving static DCOPs, it is possible to assume that solutions are found prior to the execution horizon. However, DDCOPs cannot always make this assumption as dynamics frequently occur during execution. This is particularly relevant for satellite operations; utility is obtained by completing an observation, not simply scheduling one.  If problem changes render a scheduled task obsolete before execution, maintaining that schedule is suboptimal. In addition, due to the consumption of finite resources, the initial state of any subsequent DCOP is directly dependent on the outcomes of previous solutions.

\paragraph{Metareasoning}

Metareasoning refers to an agent's reasoning about its own computation and decision-making processes \citep{russell1991principles}. 
Many autonomous agents operate under strict computational constraints, and therefore it is necessary to reason about the action of planning and scheduling. Models of bounded rationality \citep{zilberstein2011metareasoning} have modeled both the agent's knowledge about the external world and its internal computational state. In these scenarios, the selection of actions is determined by their estimated value of computation. An agent should only perform a replanning action if the estimated improvement in the utility of the new solution outweighs the cost of planning \citep{russell1991right}. The formal complexity of such decisions varies, and can range from easily solvable instances to \NP-complete \citep{Conitzer03:Definition}. Metareasoning in planning and scheduling has been used for many different types of high-level decision making such as when to replan \citep{krebsbach2009deliberation,cserna2017planning,budd2024stop}, algorithm selection \citep{lieder2014algorithm}, hyper-parameter selection and algorithm configuration \citep{schede2022survey,budd2024stop}, and resource allocation for planning and stopping criteria \citep{dean1988analysis,zilberstein1996using,alexander2008controlling,hansen2001monitoring}.  

These same techniques have been researched for MAS in addition to single-agent problems \citep{Sandholm93:Implementation,Sandholm95:Issues,raja2007framework,sarne2008effective,rubinstein201114,carlin2012bounded,cheng2013multiagent,langlois2020metareasoning,carrillo2021communication}. For MAS, there is the added complexity of reasoning about when to exchange messages with other agents. Metareasoning in MAS can occur at the agent-level (independently deciding whether to participate in planning) or the system-level (reasoning about the system holistically). Both paradigms present challenges. System-level metareasoning often requires additional layers of communication and computation to achieve consensus, a process that may not be resource-efficient. Conversely, agent-level metareasoning can lead to disjointed decisions, such as some agents planning while others do not, which significantly increases algorithmic complexity.

While metareasoning has been applied to domains like task and motion planning \citep{sung2024effort}, game playing \citep{ulam2008combining}, and generic scheduling \citep{krebsbach2009deliberation,sarne2008effective}, previous model-based research has largely concentrated on \textit{Markov Decision Processes (MDPs)} and decentralized MDPs. Within DDCOPs, the work of~\citet{Yedidsion1:Eexplorative} used \textit{function metareasoning} to adapt the factor graph used by the max-sum algorithm rather than to determine whether invoking a DDCOP solver is worthwhile. Despite this extensive literature, metareasoning has not previously been incorporated into the DDCOP framework in a general way to decide whether recomputing a solution is worthwhile given its estimated benefit and computational cost.

\paragraph{Satellite observation scheduling} Satellite observation scheduling is typically framed as an optimization problem that involves geometric reasoning, downlink scheduling, and constraint-based task allocation. The majority of research efforts and operational work has focused on centralized solutions to satellite observation scheduling \citep{globus,augenstein2016optimal,nag2018scheduling,he2018improved,shah2019scheduling,aeossp,squillaci2021managing,boerkoel2021efficientsatellite,eddy2021maximum,squillaci2023scheduling,sat-task-plan-large-areas,Barrault25:Hybridizing}. However, these approaches are often insufficient for dynamic, time-sensitive scenarios due to inherent ground communication latencies and susceptibility to single-point failures. 

There is comparatively limited work on decentralized scheduling approaches, and these primarily focus on static problems. Examples include auction-based methods \citep{picardauction,phillipsauction} and heuristic search-based methods \citep{shreya,bonnet2007collaboration,bonnet2008coordination,zilberstein2025decentralized}. We build on prior work on \textit{static} satellite operations that introduced the \textit{multi-satellite constellation observation scheduling problem (COSP)}. COSP already includes several challenges for static DCOP solvers, which DCOSP inherits. COSP assumes agents are aware of the existence of other agents but lack detailed access to their capabilities or state information (meaning an agent does not know its exact neighborhood in the constraint graph). This assumption ensures COSP and DCOSP remain viable in scenarios with intermittent connectivity, limited bandwidth, or security-mandated communication restrictions. Because standard DCOP solutions typically demand extensive computation and communication or require the exact neighborhood, they are unsuitable for COSP. Many existing DDCOP algorithms inherit assumptions that do not hold for COSP and, by extension, DCOSP.

\section{Problem definition}

In this section, we present the formal definition of DCOSP. To provide necessary context, we begin by outlining the static formulation of the application, COSP.

\subsection{Multi-satellite constellation observation scheduling}

The \textit{multi-satellite constellation observation scheduling problem (COSP)} is a static formulation of the satellite observation scheduling problem.

\begin{definition}[Multi-satellite constellation observation scheduling problem \cite{zilberstein2025decentralized}]
The multi-satellite constellation observation scheduling problem (COSP) is a $6$-tuple $\langle H, A, R, \mathcal{X}, D, C \rangle$. The main components are 

\begin{itemize}
    \item $H$: the scheduling horizon,
    \item $A$: the set of agents,
    \item $R$: the set of observation requests,
    \item $\mathcal{X}$: the set of variables,
    \item $D$: the set of downlinks, and
    \item $C$: the set of constraints.
\end{itemize}

Each variable is controlled by an agent and represents a task that can be scheduled to satisfy a single request in $R$. The goal of the optimization problem is to maximize the number of requests satisfied while not violating the constraints of any agent. 
\end{definition}
The details of the above sets in the definition are as follows.
\begin{itemize}
    \item $H = [h_s, h_e]$: the scheduling horizon.
    
    \item $A$: the set of agents in which each agent is a satellite in the constellation. 
    \item $\mathcal{T}$: the set of point targets on Earth defined by a latitude and longitude.
    \item $R$: the set of requests where each request is defined by the target to observe, $\tau \in \mathcal{T}$, and when in the scheduling horizon to observe, $h \subset H$.  Note that we index elements of a request $r$, such as the horizon, with the notation $h(r)$ and use this notation consistently for other variables. 
    \item $S = \bigcup_{a \in A} S_{a}$: the set of tasks (also referred to as observations) where each $S_{a}$ corresponds to the tasks of agent $a$. A task $s \in S_{a}$ is defined by the request being satisfied, $r \in R$, the interval required to schedule the task, $h \subset h(r)$,  and the data volume required to take the observation, $m \in \mathbb{R}^+$.
    \item $\mathcal{X} = \bigcup_{a \in A} \mathcal{X}_{a}$: the set of Boolean decision variables where each $\mathcal{X}_{a}$ corresponds to the variables of agent $a$. For each $s \in S_{a}$ we define the Boolean decision variable $x \in \mathcal{X}_{a}$ where $x = 1$ iff agent $a$ schedules task $s$. 
    \item  $D=\bigcup_{a \in A} D_{a}$: the set of downlinks where each $D_{a}$ corresponds to the downlinks of agent $a$. A downlink is defined by the maximum data volume downlinked, $m \in \mathbb{R}^+$, and the time interval of the downlink, $h \subset H$. We assume that all downlinks are mandatory. 
    \item $C = \bigcup_{a \in A} C_{a}$: the set of constraints for each agent. Each agent is constrained by processing and data volume. An agent cannot execute two tasks at once and tasks cannot overlap with downlinks. An agent must also never exceed its memory capacity and all observations acquired must be downlinked at the earliest opportunity. Formally,  
    \[ C_{a} = C_{D_{a}} \cup C_{S_{a}}. \]
    We define
    \[ C_{D_a} = \bigcup_{d \in D_{a}} c_{d} \]
    where
    \[ c_{d} = \sum_{s \in S_{a}^{d} } x(s) \cdot m(s) \leq \textsc{min}( m(a), m(d) ).\]
    The set $S_{a}^{d} $ contains the possible tasks for which the soonest downlink window in the future is $d$. The value $m(a)$ denotes the memory capacity of agent $a$. We define
    \[ C_{S_{a}} = \bigcup_{s, s' \in S_{a}, ~s\neq s' } c_{s, s'} \]
    where
    \[ c_{s, s'} = \left[ x(s) \cdot x(s') + \mathbb{I}(h(s) \cap h(s') \neq \emptyset) \leq 1 \right] .\]
\end{itemize}

The objective of COSP is to maximize the number of requests satisfied subject to the constraints. A request is satisfied if a single observation for that request is completed. An optimal assignment of variables $\mathcal{X}^*$ is defined as

\[ \mathcal{X}^* = \arg\max_{\mathcal{X}} \mathcal{F}(\mathcal{X})\]
where 
\[ \mathcal{F}(\mathcal{X}) = \sum_{r \in R} \left[ 1 - \prod_{x \in \mathcal{X}_{r}} (1-x)\right].\] 

Here, $\mathcal{X}_{r}$ is the set of variables such that $x = x(s)$ and $r(s) = r$. 

COSP has been shown to be a challenging problem for DCOP methods. Typical COSP instances have hundreds of agents and thousands of requests, leading to millions of decision variables. In addition, the constraint graph of COSP has high degrees on the order of $\Omega(|A| \cdot |R|)$. The constraint graph is also assumed to be only locally known to an agent. An agent $a$ is oblivious to all tasks $s \notin S_a$ and variables $x \notin \mathcal{X}_a$. This is not consistent with standard DCOPs in which agents know all neighboring variables/agents in the constraint graph \citep{fioretto2018distributed}. In COSP, it is assumed that agents know the existence of all other agents but have no knowledge of the variables of other agents. These factors make existing DCOP approaches that rely on agents communicating with neighboring agents in the constraint graph both computationally challenging due to the high degrees in the graph and inapplicable since we cannot assume agents know which agents they share constraints with.  

\subsection{Dynamic multi-satellite constellation observation scheduling}

We now present DCOSP, which constitutes an execution-aware DDCOP, and discuss how previous challenges from COSP transfer to DCOSP. 

A DDCOP consists of a set of sequential DCOPs. We define DCOSP similarly as a set of COSP instances. Let $\delta_t$ be the COSP instance at time $t$. We then define the DCOSP, $\delta$, as $\delta = \{ \delta_t\}_{t=0}^T$. We assume that the agents have no prior knowledge of when or how the problem might change and must act reactively. We refer to the requests and variables of $\delta_t$ as $R^{\delta_t}$ and $\mathcal{X}^{\delta_t}$. Note that $\delta_i$ depends on $\delta_j$ for $j < i$ since these prior DCOPs will determine the starting state of $\delta_i$. For example, resource expenditure affects both current and future solutions. We assume that there is a globally known horizon for a DCOSP instance, $h(\delta) = [h_s(\delta), h_e(\delta)]$ and that the horizon of each COSP instance is $h(\delta_t) =[h_s(\delta_t), h_e(\delta)]$ where $h_s(\delta_t) \in h(\delta)$. 

\begin{definition}[Dynamic multi-satellite constellation observation scheduling problem]
The dynamic multi-satellite constellation observation scheduling problem (DCOSP) is a $3$-tuple \\ $\langle H, \delta, \mathcal{F} \rangle$. The main components are 

\begin{itemize}
    \item $H$: the global scheduling horizon,
    \item $\delta = \{ \delta_t\}_{t=0}^T$: the set of COSP instances, and 
    \item $\mathcal{F} : 2^\mathcal{X} \rightarrow \mathbb{R}$: the utility function where $\mathcal{X}$ is the set of all variables across the individual COSPs. 
\end{itemize}
\end{definition}

The utility of DCOSP diverges from a standard DDCOP. In a typical DDCOP, the utility is defined as the sum of the utility functions of the individual DCOPs, which means that an optimal solution is obtained by solving each DCOP perfectly in sequence. However, this formulation does not adequately capture DCOSP utility. We define the utility of DCOSP to be the number of requests that are satisfied, where satisfaction is determined by an observation for a request being \textit{executed}. This utility rewards completing a task rather than just scheduling one. Due to the online nature of DCOSP, the scheduling horizon overlaps the execution horizon. Therefore, scheduling an observation does not guarantee that it will be executed. Consider that a task for request $r$ is scheduled at time $t_0$ in $\delta_{t_0}$ to be executed at time $t_i$. If the task is then unscheduled at some $\delta_{t_j}$ where $0 < j< i$  then $r$  will not be satisfied despite having a task scheduled in COSP instance $\delta_{t_0}$. 

\begin{figure}[t!]
    \centering
    \includegraphics[width=0.5\linewidth]{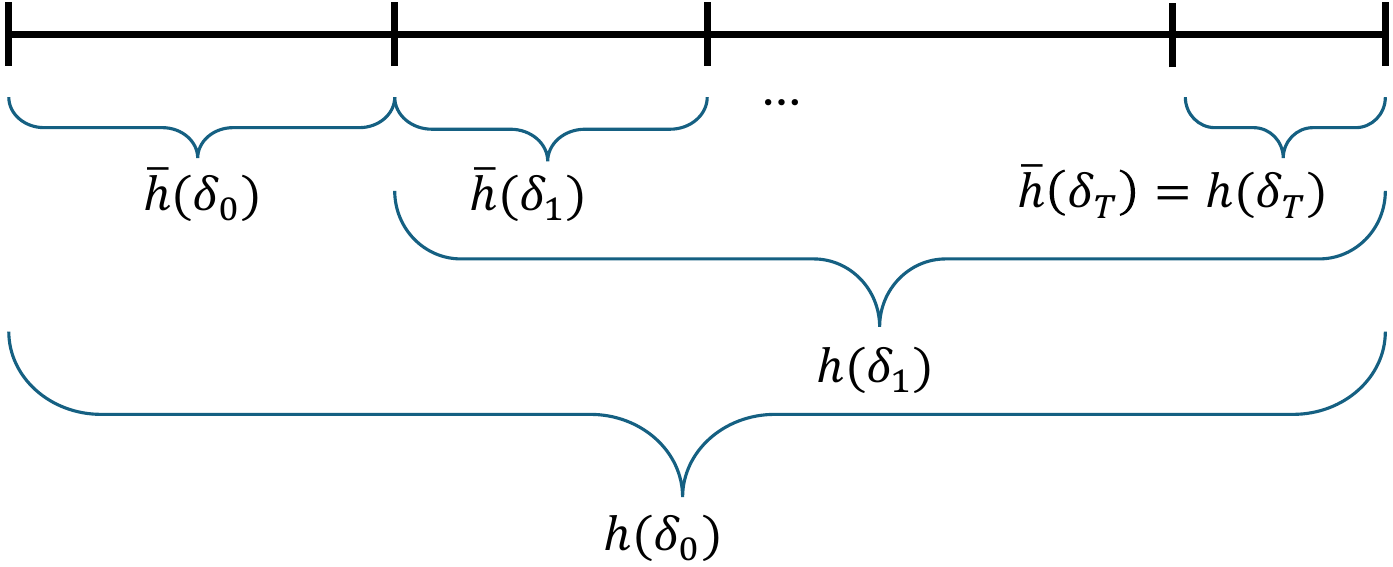}
    \caption{We illustrate $\Bar{h}(\delta_t)$; the line spans the entire horizon of the DCOSP $\delta$ where $\Bar{h}(\delta_t)$ shows the unknown interval the problem is static as defined by $\delta_t$.} 
    \label{fig:hbar}
\end{figure}

To formally define this utility, we provide some useful definitions. Let $\Bar{h}(\delta_t)$ be the unknown execution horizon of $\delta_t$. We visually show $\Bar{h}(\delta_t)$ in Figure \ref{fig:hbar}. This is the horizon for which the problem is static and is defined by $\delta_t$. Formally,

\[ \Bar{h}(\delta_t) = \begin{cases} 
      [h_s(\delta_t), h_s(\delta_{t+1})] & \text{if} ~t < T \\
      [h_s(\delta_t), h_e(\delta_{t})] & \text{else (i.e. $t= T)$}.
   \end{cases} \]

We then define the proposition $\mathsf{executed}(x)$ as

\[ \mathsf{executed}(x) = x \cdot \mathbb{I}[ \exists t ~x = x(s) \wedge h(s) \cap \Bar{h}(\delta_t) \neq \emptyset].\]

This quantity equals one if and only if $x = 1$, meaning task $s$ was scheduled, and the horizon of $s$ occurred during the time when the problem is static as defined by $\delta_t$. In other words, the task is scheduled at the moment it is required to execute. Using this proposition, we can define the utility of DCOSP for an assignment of variables $\mathcal{X}^\delta$ over all time steps: 

\[ \mathcal{F}(\mathcal{X}^\delta) = \sum_{r \in R^\delta } \left[1 - \prod_{x \in \mathcal{X}_r^\delta} (1- \mathsf{executed}(x) ) \right]  \] 

where $R^\delta= \bigcup_{t=0}^T R^{\delta_t}$ and $\mathcal{X}_r^\delta = \bigcup_{t=0}^T \mathcal{X}_r^{\delta_t}$. $R^\delta$ is the set of all requests in DCOSP $\delta$ and $\mathcal{X}_r^\delta$ are all tasks for request $r$ over $\delta$. This utility function strictly rewards completed observations.

The same challenges of solving COSP are multiplied when solving DCOSP. A single instance of COSP is computationally challenging for most DCOP methods. Solutions that are linear in the maximum degree of the constraint graph suffer due to COSP instances having degrees $\Omega(|A| \cdot |R|)$. We again assume that the constraint graph is only partially known to an agent. Therefore, solutions to DCOSP need to conform to the assumption that cross-agent edges in the constraint graph are unknown while being efficient in computation and communication. 

\subsection{DCOSP utility}

We present two examples that illustrate the difference in optimality between a standard DDCOP and DCOSP. These examples highlight the challenges of resource-constrained agents in dynamic settings and overlapping scheduling and execution horizons. These also serve as counterexample proofs that an optimal solution to DCOSP is not equivalent to an optimal solution to a standard DDCOP. In these examples, we consider a toy problem with agents that are coordinating tasks to complete. 

\subsubsection{Example 1: persistent resource constraints}
Consider a scenario where each agent possesses a persistent, non-rechargeable battery. Completing a single task fully depletes an agent's battery.

\begin{figure*}[b!]
   
  \begin{subfigure}{0.45\textwidth}
    \includegraphics[width=\linewidth]{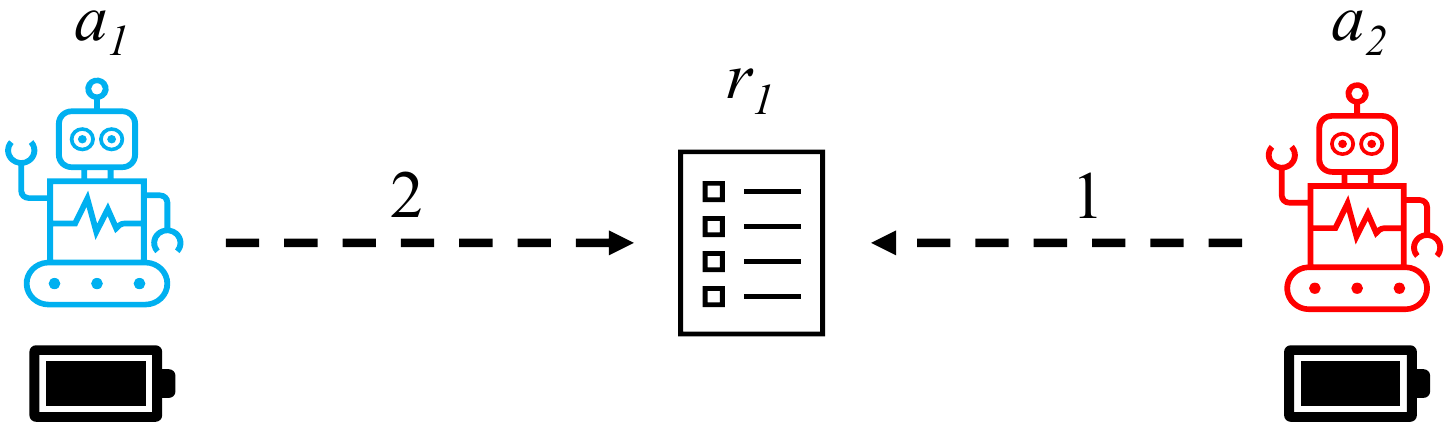}
    \caption{DCOP at time $t_0$. Agent $a_1$ can complete task $r_1$ and receive a reward of 2, while agent $a_2$ can complete task $r_1$ for a reward of 1. Both agents have a full battery. The optimal solution is for $a_1$ to complete $r_1$.} \label{fig:ex1a}
  \end{subfigure}%
  \hspace*{\fill}   
  \begin{subfigure}{0.45\textwidth}
    \includegraphics[width=\linewidth]{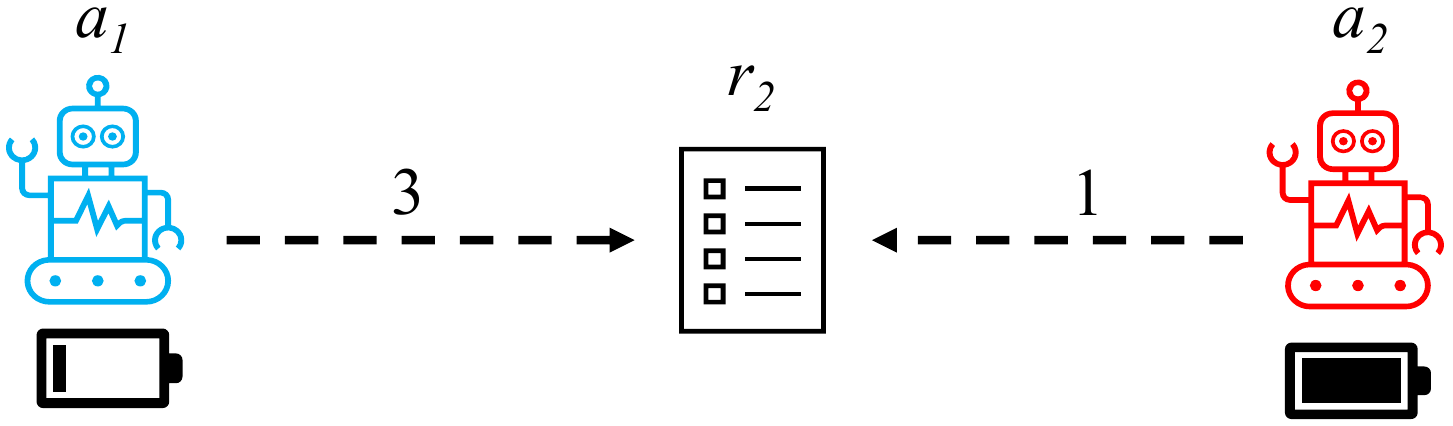}
    \caption{DCOP at time $t_1$. Agent $a_1$ could complete task $r_2$ and receive a reward of 3, however, it has no battery after completing task $r_1$ at time $t_0$. The optimal solution is for $a_2$ to complete $r_2$.} \label{fig:ex1b}
  \end{subfigure}%

\caption{This example highlights how persistent resource constraints affect the utility of a DDCOP. If the agents act optimally at each sequential time step, they will obtain less utility than if they act sub-optimally at the start.}
 \label{fig:example1}
\end{figure*}

As shown in Figure \ref{fig:example1}, agent $a_1$ and agent $a_2$ begin by each having a full battery. At time $t_0$, a single task $r_1$ is introduced that either $a_1$ or $a_2$ can complete. If $a_1$ completes $r_1$, a reward of $2$ is achieved, and if $a_2$ completes $r_1$, a reward of $1$ is achieved. An optimal solution to this first DCOP in the DDCOP is clearly for $a_1$ to complete $r_1$. The consequence is $a_1$ expends all of its battery. Now, at time $t_1$, a new task, $r_2$, is introduced. The reward if $a_1$  does $r_2$ is $3$ and the reward if  $a_2$ does $r_2$ is $1$. However, the constraints of this DCOP dictate that $a_1$ cannot do $r_2$ because it does not have battery charge. Therefore, the optimal solution to this second DCOP is for $a_2$ to do $r_2$. Therefore, the total reward the agents achieve over these two time steps is $3$. Clearly, if $a_2$ had done $r_1$ at $t_0$ and $a_1$ had done $r_2$ at $t_1$, a larger total reward would have been achieved. However, to achieve this, the DCOP at time $t_0$ would need to be solved sub-optimally.

\subsubsection{Example 2: scheduling vs. execution}
For this example, we only need to consider a scenario with a single agent. Agent $a_1$ is again constrained by the same battery as in Example 1. At time $t_0$, $a_1$ must decide between \textit{scheduling} task $r_1$ or task $r_2$. It can only schedule one task. The reward for completing task $r_1$ is $2$, while the reward for completing task $r_2$ is $1$. However, the time $r_1$ would be executed is the next day whereas $r_2$ would be executed immediately. Without considering the execution, $a_1$ will schedule $r_1$ since it has a higher reward. However, prior to the next day, at time $t_1$, $r_1$ is removed from the problem. Now, $a_1$ will receive no reward despite having $r_1$ scheduled at time $t_0$ since the time to complete $r_2$ has passed. The resulting total reward is $0$. However, had $a_1$ prioritized $r_2$ to complete, which is a suboptimal solution (when just solving the scheduling problem) at time $t_0$, a total reward of $1$ would have been achieved. 

\begin{figure*}[t!]
   
  \begin{subfigure}{0.45\textwidth}
    \centering
    \includegraphics[width=0.6\linewidth]{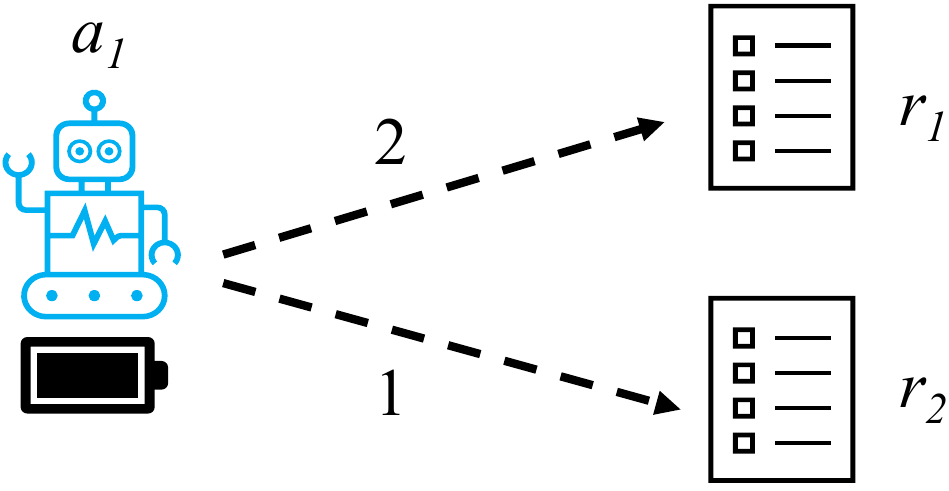}
    \caption{DCOP at time $t_0$. Agent $a_1$ decides between scheduling task $r_1$ for a reward of 2 and task $r_2$ for a reward of 1. Task $r_1$ will be scheduled the next day in the future, whereas task $r_2$ would be scheduled for execution immediately. The optimal solution is for $a_1$ to schedule $r_1$.} \label{fig:ex2a}
  \end{subfigure}%
  \hspace*{\fill}   
  \begin{subfigure}{0.45\textwidth}
  \centering
    \includegraphics[width=0.6\linewidth]{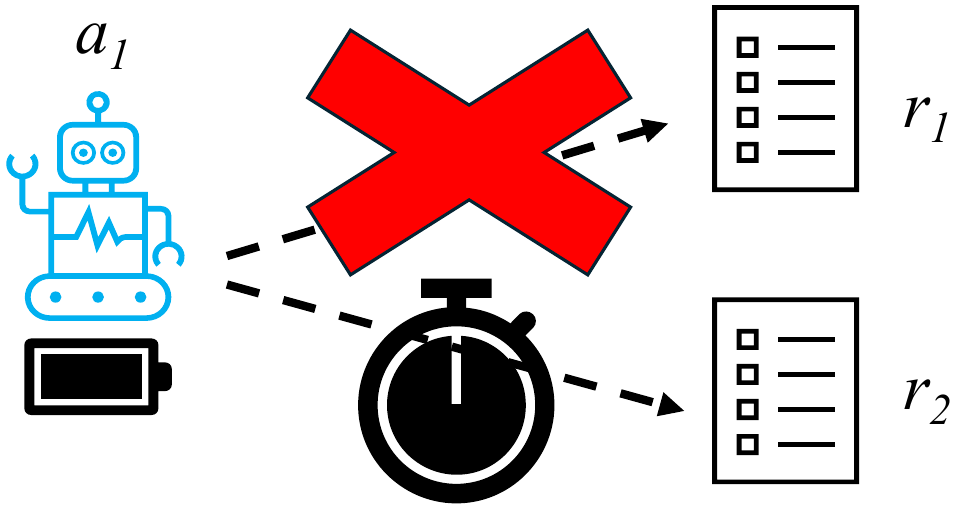}
    \caption{DCOP at time $t_1$. Task $r_1$ is removed from the current DCOP and the time to complete task $r_2$ has expired. Agent $a_1$ now receives no utility despite having previously scheduled $r_1$. Had $a_1$ prioritized $r_2$ with the sooner execution time, a utility of $1$ could have been achieved.} \label{fig:ex2b}
  \end{subfigure}%

\caption{This example highlights how receiving a reward for scheduling vs. execution affects the utility of a DDCOP. If the agent acts optimally at each sequential time step, it will obtain less utility than if it acted sub-optimally at the start.}
 \label{fig:example2}
\end{figure*}

\section{Metareasoning}

We introduce a general framework for metareasoning in DDCOPs and
instantiate it for DCOSP. Because DDCOP algorithms operate online, invoking
a solver is itself a consequential action. Planning consumes computation and
communication, and in settings where planning overlaps execution, it may
also prevent agents from completing otherwise feasible tasks. Agents should
therefore reason not only about how to update their decisions, but also about
whether the benefit of replanning justifies its cost.

This trade-off is particularly important in satellite operations. Spacecraft share limited onboard computation and memory with other flight software, and inter-satellite communication consumes bandwidth and may also incur energy or financial costs. Time spent planning or communicating may overlap observation opportunities and directly reduce mission utility.

\subsection{A metareasoning framework for DDCOPs}

Consider a DDCOP $\delta=\{\delta_t\}_{t=0}^{T}$.
At each dynamic event, an agent may either invoke its DDCOP solver or retain
its current solution after projecting it onto the new problem instance. We
represent the information available to agent $a$ at time $t$ by a local
information state $\mathcal{I}_{a,t}$. This state may include the current
DCOP $\delta_t$, the agent's previous solution, its locally known
neighborhood, the duration of planning, and any estimates of solver
performance or resource expenditure.

\begin{definition}[Metareasoning oracle]
A metareasoning oracle for agent $a$ is a function
\[
    \mathcal{M}_a:\mathfrak{I}_a\rightarrow\{0,1\},
\]
where $\mathfrak{I}_a$ is the space of local information states available to
agent $a$. The value $\mathcal{M}_a(\mathcal{I}_{a,t})=1$ indicates that
agent $a$ invokes its DDCOP solver at time $t$, while
$\mathcal{M}_a(\mathcal{I}_{a,t})=0$ indicates that it does not.
\end{definition}

Let $\Pi_{\delta_t}(\mathcal{X}_{a,t-1})$ denote the projection of agent $a$'s previous solution onto the variables and constraints that remain active in $\delta_t$. At minimum, this operation
removes deleted variable assignments in the current problem instance. The projection is performed
regardless of whether the agent invokes the solver.

Algorithm~\ref{alg:meta} presents the resulting metareasoning framework.
Given a DDCOP solver $\lambda_a$, the agent first projects its previous
solution onto the current problem and then decides whether further
recomputation is worthwhile.

\begin{algorithm*}[t!]
\caption{DDCOP metareasoning framework for agent $a$}
\label{alg:meta}
\begin{flushleft}
\textbf{Input}: Local information state $\mathcal{I}_{a,t}$,
previous solution $\mathcal{X}_{a,t-1}$,
current DCOP $\delta_t$, DDCOP solver $\lambda_a$ \\
\textbf{Output}: Solution $\mathcal{X}_{a,t}$ for agent $a$
\end{flushleft}
\begin{algorithmic}[1]
    \STATE
    $\widetilde{\mathcal{X}}_{a,t-1}
    \gets
    \Pi_{\delta_t}(\mathcal{X}_{a,t-1})$
    \IF{$\mathcal{M}_a(\mathcal{I}_{a,t})=1$}
        \STATE
        $\mathcal{X}_{a,t}
        \gets
        \lambda_a(\delta_t,\widetilde{\mathcal{X}}_{a,t-1})$
    \ELSE
        \STATE
        $\mathcal{X}_{a,t}
        \gets
        \widetilde{\mathcal{X}}_{a,t-1}$
    \ENDIF
    \RETURN $\mathcal{X}_{a,t}$
\end{algorithmic}
\end{algorithm*}

The framework is independent of the particular DDCOP solver. However, agent-level metareasoning requires the solver to tolerate partial participation or to provide a coordination mechanism for determining which agents participate in a replanning episode. Algorithms such as DSA and NSS can naturally proceed when some agents do not send updates, whereas complete algorithms based on fixed pseudo-trees may fail or lose their guarantees without full participation. Extending such methods to support independent metareasoning decisions remains an open problem. A conventional DDCOP in
which agents always replan is recovered by setting $\mathcal{M}_a(\mathcal{I}_{a,t})=1$ for every agent and time step.

\subsection{Ideal value of replanning}
\label{sec:meta}

Deciding whether to replan requires comparing the potential improvement in
solution quality with the opportunity cost incurred while planning. We first
define an ideal value of replanning that assumes access to an optimal
post-replanning solution. This quantity provides a normative benchmark; the
realized benefit of invoking an incomplete solver may be smaller.

Let $t\in H$ denote the time at which replanning begins, and let
$h_{\mathrm{plan}}(t)\subseteq H$ denote the interval during which planning
prevents task execution. In models that ignore planning time, we set
$h_{\mathrm{plan}}(t)=\emptyset$.

For a task variable $x$, define
\[
    \mathsf{available}(x,t)
    =
    x\cdot
    \mathbb{I}
    \left[
        h(s(x))\cap h_{\mathrm{plan}}(t)=\emptyset
    \right].
\]
This quantity equals one if task $x$ is scheduled and its execution does not
overlap the planning interval.

For a current solution $\mathcal{X}_{t-1}$, its utility before replanning is
\[
    \mathcal{F}(\mathcal{X}_{t-1})
    =
    \sum_{r\in R_t}
    \left[
        1-
        \prod_{x\in\mathcal{X}_{r,t-1}}(1-x)
    \right].
\]

For any candidate solution $\mathcal{X}_t$, define its execution-aware
utility after accounting for the planning interval as
\[
    \widehat{\mathcal{F}}_t(\mathcal{X}_t)
    =
    \sum_{r\in R_t}
    \left[
        1-
        \prod_{x\in\mathcal{X}_{r,t}}
        \bigl(1-\mathsf{available}(x,t)\bigr)
    \right].
\]

\begin{definition}[Ideal value of replanning]
Let $\mathcal{X}_t^*$ be an optimal solution to the current
subproblem subject to the planning interval $h_{\mathrm{plan}}(t)$. The ideal
value of replanning at time $t$ is
\[
    V_{\mathrm{replan}}^*
    (\mathcal{X}_{t-1},t)
    =
    \widehat{\mathcal{F}}_t(\mathcal{X}_t^*)
    -
    \mathcal{F}(\mathcal{X}_{t-1}).
\]
Under an optimal replanning procedure, replanning is beneficial whenever
\[
    V_{\mathrm{replan}}^*
    (\mathcal{X}_{t-1},t)>0.
\]
\end{definition}

For a particular solver $\lambda$, the realized value of invoking the solver
can instead be written as
\[
    V_{\mathrm{replan}}^{\lambda}
    (\mathcal{X}_{t-1},t)
    =
    \widehat{\mathcal{F}}_t
    \left(
        \lambda(\delta_t,\mathcal{X}_{t-1})
    \right)
    -
    \mathcal{F}(\mathcal{X}_{t-1}).
\]
If $\lambda$ is stochastic, the solver-specific value is
$\mathbb{E}[V_{\mathrm{replan}}^{\lambda}]$. For every feasible solver
output,
\[
    V_{\mathrm{replan}}^{\lambda}
    (\mathcal{X}_{t-1},t)
    \leq
    V_{\mathrm{replan}}^*
    (\mathcal{X}_{t-1},t).
\]
The ideal value therefore upper bounds the realized value of any incomplete
solver.

We next decompose the ideal value into an opportunity for improvement and
an opportunity cost of planning. Define
\[
    \mathcal{C}_{\mathrm{replan}}
    (\mathcal{X}_{t-1},t)
    =
    \mathcal{F}(\mathcal{X}_{t-1})
    -
    \widehat{\mathcal{F}}_t(\mathcal{X}_{t-1}).
\]
This is the utility lost when the current schedule is retained but tasks
overlapping the planning interval cannot be executed.

Define the opportunity for improvement as
\[
    \mathcal{O}_{\mathrm{replan}}^*
    (\mathcal{X}_{t-1},t)
    =
    \widehat{\mathcal{F}}_t(\mathcal{X}_t^*)
    -
    \widehat{\mathcal{F}}_t(\mathcal{X}_{t-1}).
\]
This is the maximum improvement obtainable after accounting for the same
planning interval.

\begin{proposition}[Opportunity-cost decomposition]
The ideal value of replanning satisfies
\[
    V_{\mathrm{replan}}^*
    (\mathcal{X}_{t-1},t)
    =
    \mathcal{O}_{\mathrm{replan}}^*
    (\mathcal{X}_{t-1},t)
    -
    \mathcal{C}_{\mathrm{replan}}
    (\mathcal{X}_{t-1},t).
\]
\end{proposition}

\begin{proof}
By substitution,
\begin{align*}
\mathcal{O}_{\mathrm{replan}}^*
-
\mathcal{C}_{\mathrm{replan}}
&=
\left[
    \widehat{\mathcal{F}}_t(\mathcal{X}_t^*)
    -
    \widehat{\mathcal{F}}_t(\mathcal{X}_{t-1})
\right] \\
&\quad -
\left[
    \mathcal{F}(\mathcal{X}_{t-1})
    -
    \widehat{\mathcal{F}}_t(\mathcal{X}_{t-1})
\right] \\
&=
\widehat{\mathcal{F}}_t(\mathcal{X}_t^*)
-
\mathcal{F}(\mathcal{X}_{t-1}) \\
&=
V_{\mathrm{replan}}^*
(\mathcal{X}_{t-1},t).
\end{align*}
\end{proof}

Since the DCOSP objective sums unit values, the planning cost can equivalently be
written as
\[
\mathcal{C}_{\mathrm{replan}}
(\mathcal{X}_{t-1},t)
=
\sum_{r\in R_t}
\left[
\prod_{x\in\mathcal{X}_{r,t-1}}
\bigl(1-\mathsf{available}(x,t)\bigr)
-
\prod_{x\in\mathcal{X}_{r,t-1}}
(1-x)
\right].
\]
Intuitively, $\mathcal{C}_{\mathrm{replan}}$ counts requests that are satisfied by the current
schedule but for which all scheduled observations are blocked by replanning.

\subsection{Local metareasoning}

An individual agent generally cannot evaluate the global quantities above.
In particular, it may not know the global current utility or the optimal
post-replanning solution. We therefore define local quantities over a
subproblem defined by an agent's neighborhood $\mathcal{N}$.

Let $A_{\mathcal{N},t}$ and $R_{\mathcal{N},t}$ denote the agents and
requests in subproblem $\mathcal{N}$, and let
$\mathcal{X}_{\mathcal{N},t-1}$ denote its current assignment. We use
$\mathcal{F}_{\mathcal{N}}$ and
$\widehat{\mathcal{F}}_{\mathcal{N},t}$ to denote the restrictions of the
utility functions to requests in $R_{\mathcal{N},t}$.

\begin{definition}[Ideal local value of replanning]
The ideal local value of replanning for subproblem $\mathcal{N}$ is
\[
    V_{\mathrm{replan},\mathcal{N}}^*
    (\mathcal{X}_{\mathcal{N},t-1},t)
    =
    \widehat{\mathcal{F}}_{\mathcal{N},t}
    (\mathcal{X}_{\mathcal{N},t}^*)
    -
    \mathcal{F}_{\mathcal{N}}
    (\mathcal{X}_{\mathcal{N},t-1}),
\]
where $\mathcal{X}_{\mathcal{N},t}^*$ is an optimal feasible assignment for
the local subproblem after accounting for the planning interval.
\end{definition}

Computing this value is generally intractable because it requires solving a
DCOP optimally. We therefore distinguish estimates that upper- or
lower-bound the ideal local value.

\begin{definition}[Optimistic and pessimistic local estimates]
An estimate $\overline{V}_{\mathcal{N},t}$ is optimistic if
\[
    \overline{V}_{\mathcal{N},t}
    \geq
    V_{\mathrm{replan},\mathcal{N}}^*,
\]
and an estimate $\underline{V}_{\mathcal{N},t}$ is pessimistic if
\[
    \underline{V}_{\mathcal{N},t}
    \leq
    V_{\mathrm{replan},\mathcal{N}}^*.
\]
\end{definition}

Optimistic estimates bias an agent toward replanning, whereas pessimistic
estimates bias it toward retaining its current solution. In DCOSP, we prefer
a replanning-biased heuristic because unnecessarily skipping a beneficial
replanning episode may permanently forfeit observation utility.

\subsection{A practical local metareasoning heuristic}
\label{sec:meta-heuristic}

We construct an efficiently computable local heuristic by separately
estimating the opportunity for improvement and the cost of planning.

Agent $a$ can compute its own planning cost exactly, by computing the number of requests locally scheduled that are blocked by planning. We refer to this quantity as $\mathcal{C}_{a,t}$. 
Motivated by the similar orbital geometry of satellites within the same GND neighborhood, as defined by the decomposition used in Section~\ref{sec:dnss}, we estimate the neighborhood-wide planning cost as
\[
    \widehat{\mathcal{C}}_{\mathcal{N},t}^{\,a}
    =
    |A_{\mathcal{N},t}|\cdot \mathcal{C}_{a,t}.
\]

We can also upper bound the optimal post-replanning utility of a local subproblem by the number of requests in that subproblem:
\[
    \widehat{\mathcal{F}}_{\mathcal{N},t}
    (\mathcal{X}_{\mathcal{N},t}^*)
    \leq
    |R_{\mathcal{N},t}|.
\]
We therefore estimate the opportunity for improvement by
\[
    \widehat{ \mathcal{O}}_{\mathcal{N},t}^{\,a}
    =
    |R_{\mathcal{N},t}|
    -
    \widehat{\mathcal{F}}_{\mathcal{N},t}
    (\mathcal{X}_{\mathcal{N},t-1}).
\]

The practical local value estimate used by agent $a$ is
\[
    \widehat{V}_{\mathcal{N},t}^{\,a}
    =
    \widehat{\mathcal{O}}_{\mathcal{N},t}^{\,a}
    -
    \widehat{\mathcal{C}}_{\mathcal{N},t}^{\,a}.
\]
The corresponding metareasoning oracle is
\[
    \mathcal{M}_a(\mathcal{I}_{a,t})
    =
    \mathbb{I}
    \left[
        \widehat{V}_{\mathcal{N},t}^{\,a}>0
    \right].
\]

Because $|R_{\mathcal{N},t}|$ upper bounds the optimal post-replanning utility, $\widehat{ \mathcal{O}}_{\mathcal{N},t}^{\,a}$ upper bounds the ideal local opportunity for improvement.
However, the uniform cost estimate may either overestimate or underestimate
the true neighborhood cost. As a result,
$\widehat{V}_{\mathcal{N},t}^{\,a}$ is not guaranteed to be optimistic in
the formal sense defined above. It is a practical
local metareasoning heuristic rather than an optimistic oracle.

The heuristic is motivated by the correlated observation opportunities of
satellites in the same orbital neighborhood. It is inexpensive for each
agent to compute and requires no additional optimization. As shown in
Section~\ref{sec:results}, it preserves solution quality while saving computational resources across the evaluated scenarios.
\section{Algorithms}

In this section, we present various algorithms for solving DCOSP. First, we show how to construct an optimal solution to DCOSP using omniscient knowledge of problem dynamics. We then present the \textit{dynamic incremental neighborhood stochastic search (D-NSS)} algorithm, which is a scalable, incomplete DDCOP algorithm. Finally, we establish several baseline approaches, including dynamic variations and adaptations of the DSA algorithm.

\subsection{Obtaining an optimal solution to DCOSP}\label{sec:opt}

There is no clear online algorithm to compute an optimal DCOSP solution. This is due to agents having no prior knowledge of problem dynamics and solving each individual COSP instance optimally not necessarily constituting an optimal DCOSP solution. However, we can obtain an optimal solution by collapsing DCOSP into a single DCOP. This requires hindsight knowledge and is therefore unavailable online.

We can collapse a DCOSP $\delta$ to a DCOP $\delta'$ to reason only about the observations that matter. The following are the key set constructions of $\delta'$.
\begin{align*}
S' &= \{ s\in S^\delta ~|~ \exists_t ~ h(s) \cap \Bar{h}(\delta_t) \neq \emptyset\} \text{ and}\\
\mathcal{X}' & = \{ x\in \mathcal{X}^\delta ~|~ \exists_t ~x = x(s) \wedge h(s) \cap \Bar{h}(\delta_t) \neq \emptyset\}.
\end{align*}

\begin{proposition}\label{prop:opt}
   An optimal solution to the above construction of a DCOP $\delta'$ induces an optimal solution to the DCOSP $\delta$.
\end{proposition}
\begin{proof}
Let $\mathbf{x}$ be any feasible assignment for the DCOSP $\delta$, and
let $\mathbf{x}'$ denote its restriction to the variables
$\mathcal{X}'$. By construction,
\[
\mathsf{executed}(x)=
\begin{cases}
x, & x\in\mathcal{X}',\\
0, & x\notin\mathcal{X}'.
\end{cases}
\]
Variables outside $\mathcal{X}'$ therefore contribute no utility. For every
request $r\in R$, each such variable contributes a factor of
$1-\mathsf{executed}(x)=1$ to the corresponding product. Hence,
\begin{align*}
\mathcal{F}(\mathbf{x})
&=
\sum_{r\in R}
\left[
1-
\prod_{x\in\mathcal{X}_r}
\bigl(1-\mathsf{executed}(x)\bigr)
\right] \\
&=
\sum_{r\in R}
\left[
1-
\prod_{x\in\mathcal{X}'_r}
(1-x)
\right] \\
&=
\mathcal{F}(\mathbf{x}').
\end{align*}

It remains to show that this equivalence preserves feasibility. The
constraints of $\delta'$ are the constraints of $\delta$ restricted to
the tasks and variables in $\mathcal{S}'$ and $\mathcal{X}'$, and restricting any feasible DCOSP assignment $\mathbf{x}$ to
$\mathcal{X}'$ yields a feasible assignment $\mathbf{x}'$ for $\delta'$. In addition, let $\mathbf{x}'$ be any feasible assignment for $\delta'$.
We extend it to an assignment $\mathbf{x}$ for $\delta$ by setting
$x=0$ for every $x\in\mathcal{X}\setminus\mathcal{X}'$. The preceding equality then gives $\mathcal{F}(\mathbf{x})=\mathcal{F}(\mathbf{x}')$.
\end{proof}

Solving $\delta'$ optimally results in an optimal solution to the DCOSP $\delta$. By definition, $\delta'$ is a static COSP instance. Therefore, we can employ any complete algorithm to obtain an optimal solution to $\delta$ by solving $\delta'$. We reiterate that this optimal way of solving a DCOSP instance relies on constructing a COSP instance using omniscient knowledge and is impossible to deploy online. This method serves to benchmark incomplete online approaches. However, even solving a single DCOP is \NP-hard, and DCOSP is a generalization of it. Therefore, for large DCOSP instances, we use an incomplete solver to obtain a near-optimal solution since any exponential time search is infeasible. 

\subsection{Dynamic incremental neighborhood stochastic search}
\label{sec:dnss}
Due to the scale of DCOSP and the computational constraints of satellites, we require DDCOP algorithms that are both efficient and can reason about the dynamic nature of the problem. D-NSS (Algorithm \ref{alg:DNSS}) extends NSS to address the drawbacks of general DDCOP algorithms. NSS is an iterative algorithm where at each iteration agents stochastically update their variable assignments based on the assignments of agents they communicate with. This iterative procedure is shared with other algorithms such as DSA. However, NSS relies on a decomposition heuristic $\Upsilon: A \times S \rightarrow \{0,1\}$ to generate a subproblem $\mathcal{N}$ for neighborhoods of agents to solve. This subproblem is a smaller DCOP consisting of requests $R_\mathcal{N}$ and agents $A_\mathcal{N}$. D-NSS continually computes subproblems, repairs local solutions between problem instances, and reasons about prior scheduling and execution in the search and repair phases.

\begin{algorithm}[b!]
\caption{Dynamic Incremental Neighborhood Stochastic Search (D-NSS) for agent $a$}\label{alg:DNSS}
\begin{flushleft}
\textbf{Input}: $sched, R, A, S_{a}, C_{a}, \Upsilon, maxIters$ \\
\textbf{Output}: Schedule for agent $a$
\end{flushleft}
\begin{algorithmic}[1]
        \STATE$\mathcal{N} = \textsc{computeSubProblem}(a, A, R, S_{a}, \Upsilon)$
        \STATE $sched = \textsc{repair}(sched, R_\mathcal{N}, S_{a}, C_{a})$
        \STATE $sched = \textsc{D-NSS-Search}(sched, R_\mathcal{N}, A_\mathcal{N}, S_{a}, C_{a}, maxIters)$
       \RETURN $sched$
\end{algorithmic}
\end{algorithm}

We leverage the \textit{geometric neighborhood decomposition (GND)} heuristic to create these subproblems every time dynamics occur \citep{zilberstein2025decentralized}. GND efficiently allocates requests to neighborhoods of agents based on the orbital geometry of the constellation and has been shown to effectively partition COSP instances. For completeness, we provide an outline of GND. GND was first introduced with the NSS algorithm and we refer the reader to prior work for a full presentation of the heuristic \citep{zilberstein2025decentralized}. 

GND leverages the orbital geometry of the satellite constellation to hierarchically partition requests to neighborhoods of agents. We consider the set of orbital planes that define a satellite constellation. An orbital plane is a fixed orbit around Earth that many satellites may follow at various spacing. For every request, an agent computes the  supply from all orbital planes and adds these to estimate the total number of agents with overflights for the request. The supply can also be thought of as an estimate of degrees in the constraint graph. Iterating through requests in ascending order of supply, a request gets assigned to the $n$ neighborhoods with the highest ratio of supply to temporal conflicts. Temporal conflicts are counts of other requests already allocated to a neighborhood that overlap in time with a given request. Finally, within a neighborhood, requests are further subdivided to agents based on biases towards specific tiles on Earth. GND with $n$ degrees of incompleteness is denoted GND($n$). 

\begin{algorithm}[t!]
\caption{\textsc{repair} for agent $a$}\label{alg:repair}
\begin{flushleft}
\textbf{Input}: $sched, R, S_{a}, C_{a}$\\
\textbf{Output}: Repaired Schedule for agent $a$
\end{flushleft}
\begin{algorithmic}[1]
        \FOR{$s \in sched$}
            \IF{$r(s) \notin R$}
                \STATE \textsc{removeFromSchedule($sched, s$)}
            \ENDIF
        \ENDFOR
        \STATE \textbf{shuffle} $S_{a}$
        \FOR{$s \in S_{a}$}
            \IF{\text{$s$ satisfies $C_{a} ~\wedge~ \nexists s' \in sched : r(s') = r(s)$}}
                \STATE \textsc{addToSchedule($sched, s$)}
            \ENDIF
        \ENDFOR
      \RETURN $sched$
\end{algorithmic}
\end{algorithm}

\begin{algorithm}[t!]
\caption{\textsc{D-NSS-Search} for agent $a$}\label{alg:NSS-search}
\begin{flushleft}
\textbf{Input}: $sched, R_\mathcal{N}, A_\mathcal{N}, S_{a}, C_{a}, maxIters$ \\
\textbf{Output}: Schedule for agent $a$
\end{flushleft}
\begin{algorithmic}[1]    
      \STATE $ i \gets 0$
      \WHILE{$i < maxIters \wedge$ not converged}
            \STATE {$com\_out, com\_in$ $= \textsc{message}(A_\mathcal{N},$ $sched$)}
            \STATE{\textbf{shuffle} $R_\mathcal{N}$}
            \FOR{$r \in R_\mathcal{N}$} 
                \STATE assigned $ = \textsc{stochasticUpdate}(r, sched, com\_in)$
                \IF{assigned $= \textsc{true}$}
                     \STATE $sched$ $= \textsc{schedule}(r, sched, S_{a})$
                \ENDIF    
            \ENDFOR
            \STATE $i \gets i+1$
      \ENDWHILE
      \STATE \textbf{return} $sched$
\end{algorithmic}
\end{algorithm}

D-NSS restarts the search phase when changes are initiated in the problem. A key procedure is the \textsc{repair} function that repairs previously computed solutions to leverage assignments of unchanging variables shown in Algorithm \ref{alg:repair}. Repairing consists of removing all tasks that are no longer in the current problem instance and greedily inserting new tasks in random order. Note that we can also remove from an agent's schedule all requests that have been previously executed by agents in the same neighborhood. One benefit of the D-NSS algorithm is that it can leverage different repair procedures. We use the random repair function for two main reasons. Random initialization is the standard for DSA and NSS variants in static domains \citep{dsa} as it promotes diverse solutions  and D-NSS is designed to be as lightweight as possible. Computation-intensive repair procedures counteract the efficiency of the algorithm. We show later on that performing random repair improves the quality and efficiency of D-NSS on DCOSP instances.

After each agent repairs its solution, all agents synchronously begin the iteration phase, \textsc{D-NSS-Search}, to fine-tune the repaired solutions. \textsc{D-NSS-Search} extends the search phase of NSS to account for tasks that have just been scheduled versus ones that have been executed.  We detail the following sub-procedures.

\begin{itemize}
    \item $\textsc{computeSubProblem}(a, A, R, S_{a}, \Upsilon)$. This function computes the neighborhood of agent $a$ and the subset of requests for that neighborhood using the decomposition heuristic $\Upsilon$. We use $\Upsilon =$ GND. GND produces neighborhoods that are reflexive and transitive. 

    \item $\textsc{message}(A_\mathcal{N}, sched$). This function defines the message exchange between agents in a neighborhood. Each agent $a$ sends to each other agent in  $A_\mathcal{N} \setminus\{a\}$ the subset of $R_\mathcal{N}$ that it has executed a task for already and the subset that it has scheduled in the previous iteration via the variable $com\_out$. The resulting data structure $com\_in$ contains the satisfaction information for the neighborhood.  
    \item $\textsc{stochasticUpdate}(r, sched, com\_in)$. This function computes the assignment of an agent and a request based on the neighborhood's communication. An assignment refers to if an agent should attempt to schedule a specific request. Let $W$ be the count of agents that scheduled or executed $r$ in the previous iteration, the probability $P_u$ be a hyperparameter, $\textsf{executed}(r)$ be the predicate that $r$ was executed already, and $\textsf{assigned}(a,r)$ be the predicate that $a$ is assigned to $r$. An agent computes the probability of assigning to $r$ in the next iteration, $P(a, r| com\_in)$ using the update scheme from Table \ref{tab:update}. For example, according to $com\_in$, if request $r$ has not been executed, agent $a$ is not assigned to it, and $W \ge 1$, agent $a$ will always remain unassigned to $r$. This update scheme extends the static NSS update schemes to account for dynamic scheduling and execution. If a task for a request has already been executed, then all agents should unassign. Note that $\textsf{executed}(r) \implies W\ge1$. 

\begin{table}[h]
    \centering
    \begin{tabular}{@{}lccc@{}}
        \toprule
         & $\mathsf{executed}(r)$ & $\neg \mathsf{executed}(r)$ & $\neg \mathsf{executed}(r) $ \\ 
         & & $\neg \mathsf{assigned}(a,r)$ & $  \mathsf{assigned}(a,r)$ \\ 
        \midrule
        $W = 0$     & N/A & 1 & $1-P_u$ \\
        $W \geq 1$  & 0   & 0 & $1/W$   \\
        \bottomrule
    \end{tabular}
    \caption{Stochastic assignment update scheme for agent $a$. The table values denote the probability that agent $a$ assigns to request $r$ based on $com\_in$, $P(a, r| com\_in)$.}
    \label{tab:update}
\end{table}

    \item $\textsc{schedule}(r, sched, S_{a})$.  This function tries to schedule a task for request $r$ given the current schedule.  If a task for $r$ satisfies $C_{a}$ it is inserted into the schedule. Otherwise, the scheduler may remove a single task from the schedule to satisfy the constraints. The task with the closest start time to the new task is used as a heuristic for removal. Agent $a$ remains assigned to a removed task and can attempt to re-schedule it in subsequent iterations. Task removal enables the search phase to overcome local maxima.  Note that vanilla DCOSP considers all requests of equal priority and there is no temporal flexibility in the start or end times of tasks. However, DCOSP and the scheduling procedure are amenable to these extensions.  
     
\end{itemize}

Without prior knowledge of the problem dynamics, it is difficult for any online algorithm to reason about which scheduled tasks will be executed. Prioritizing requests that are earlier in the horizon may be more likely to persist until execution and yield reward. However, expending resources early on in the horizon results in fewer resources available to handle problem dynamics later. Although D-NSS currently focuses on reactive repair, it can be extended to integrate proactive scheduling through predictive models, which is a subject of future work. 

We introduce variables to analyze the complexity of D-NSS. Let $L$ be the maximum size of an agent's schedule, $A_\mathcal{N}$ be the largest set of agents in a sub-problem, and $R_\mathcal{N}$ be the largest set of requests in a sub-problem. In general $L \ll |R_\mathcal{N}| \ll |R|$ since $L$ is constrained by resources and time. Sub-problem computation via GND has a time complexity of $O(|R_\mathcal{N}|)$ and uses no communication. The \textsc{repair} procedure is individually computed by an agent and takes $O(|R_\mathcal{N}| \log L)$ time and again uses no communication. D-NSS inherits from NSS a computation and communication complexity of $O(|A_\mathcal{N}|{\cdot}|R_\mathcal{N}|)$ during an iteration. 

\begin{proposition}\label{prop:dnss-complexity}
    D-NSS requires an agent to send $O(|A_\mathcal{N}|)$ messages each of size $O(|R_\mathcal{N}|)$ and perform $O(|A_\mathcal{N}|{\cdot}|R_\mathcal{N}|)$ operations per iteration each time a new solution is computed. 
\end{proposition}
\begin{proof}
    Subproblem computation via GND has a time complexity of $O(|R_\mathcal{N}|)$ and uses no communication. The \textsc{repair} procedure is individually computed by an agent and takes $O(|R_\mathcal{N}| \log L)$ time and again uses no communication. D-NSS inherits from NSS a computation and communication complexity of $O(|A_\mathcal{N}|{\cdot}|R_\mathcal{N}|)$ during an iteration.  
\end{proof}

\subsection{Baseline algorithms}

We evaluate several baseline algorithms in addition to D-NSS. The naive alternative to D-NSS is to recompute a solution from scratch every time problem dynamics occur. We refer to this procedure as 0-NSS. 0-NSS has the same theoretical complexities as D-NSS.  We also evaluate dynamic variations of the \textit{Distributed Stochastic Search Algorithm} (DSA) \citep{dsa,shreya}. D-DSA uses Algorithm \ref{alg:repair} to repair DSA solutions. Likewise, 0-DSA runs DSA from scratch every time the problem changes. These DSA variants have a complexity of $O(|A|{\cdot}|R|)$ per iteration. Dynamic DSA is one of the most lightweight DDCOP solvers and can be deployed without violating the constraint graph assumptions of DCOSP. Other DDCOP algorithms such as a dynamic variation of MGM \citep{mgm}, would also incur a computation and communication complexity of $O(|A|{\cdot}|R|)$ per iteration. Due to the scale of the problem instances and the constraints of satellite computation and communication, we are unable to evaluate any exponential-time algorithms.

We include two baseline non-communication-reliant algorithms: random and greedy. Both of these algorithms construct a schedule for an agent without reasoning about other agents. These algorithms transition schedules after the dynamics occur by removing redundant tasks.  The random solver randomly inserts tasks into an agent's schedule while the greedy algorithm orders tasks by their start time and iteratively inserts them into a schedule in a single pass. Although simple, a greedy solver is comparable to deployed planners on spacecraft \citep{m2020-planner-astra2022,Chien25:Dynamic}. We summarize the per-agent complexity of the DDCOP algorithms in Table \ref{tab:complexity}. It is assumed that all agents know the request set. Therefore, the fully decentralized algorithms incur no communication. In the table, $T$ is the number of dynamics of the DDCOP and $k$ is the number of iterations of iterative algorithms.

Finally, we use the previously outlined construction to obtain optimal and near-optimal solutions. For small problems, we obtain an optimal solution by constructing the static COSP instance and solving it with a centralized branch and bound. For large problems, computing an optimal solution is not computationally feasible so we lower bound the optimal solution using \textit{squeaky wheel optimization (SWO)} \citep{swo}, an incomplete centralized solver.

\begin{table}[t!]
    \centering
    \begin{tabular}{@{}lll@{}}
        \toprule
        \textbf{Algorithm} & \textbf{Computation} & \textbf{Communication}  \\
        \midrule
        Random & $T {\cdot} |R| {\cdot} \log L $ & N/A \\
        Greedy & $T {\cdot}  |R| {\cdot} \log |R| $ & N/A \\
        0-NSS \& D-NSS & $T {\cdot} k  {\cdot} |R_\mathcal{N}|  {\cdot} |A_\mathcal{N}| $ & $T {\cdot} k {\cdot}|R_\mathcal{N}|  {\cdot}|A_\mathcal{N}| $ \\
        0-DSA \& D-DSA & $T {\cdot} k  {\cdot} |R| {\cdot}  |A| $ & $T {\cdot} k {\cdot}|R| {\cdot} |A| $ \\
        \bottomrule
    \end{tabular}
   \caption{$O(\cdot)$ complexity of distributed algorithms per agent.}
    \label{tab:complexity}
\end{table}

\section{Results}
\label{sec:results}
In this section, we detail the experimental setup and the results of our algorithms and metareasoning framework.
\subsection{Setup}
We outline the experimental setup, including the constellations, dynamic observation campaigns, and hyperparameters.

\subsubsection{Satellite constellations}\label{constellation}
\begin{figure}[b!]
    \centering
    \includegraphics[width=0.35\linewidth]{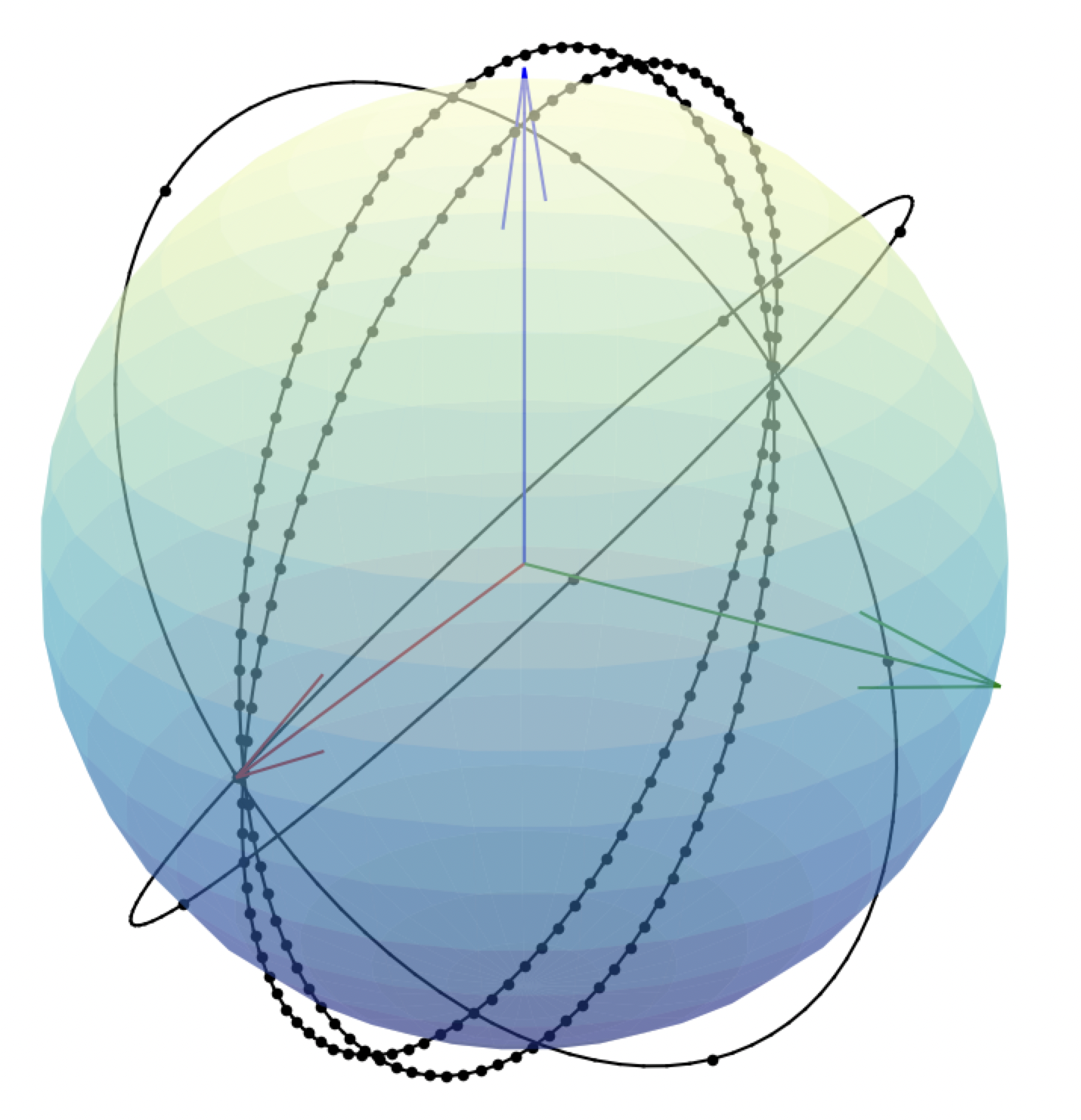}
    \includegraphics[width=0.35\linewidth]{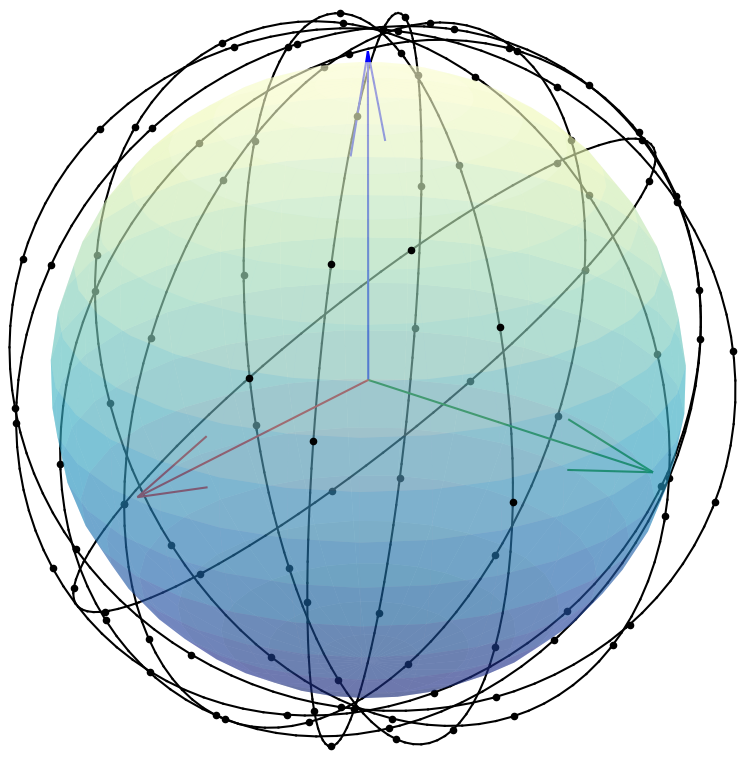}
    \caption{The satellite constellations: Planet (left) and Walker (right). Dots show satellites in an orbital plane.} 
    \label{fig:constellation}
\end{figure}

We evaluate two constellations modeled after operational low Earth orbit constellations \citep{planet}. The Planet constellation is modeled on the Dove constellation from Planet Labs. This constellation is composed of two near sun-synchronous orbital planes at $95^\circ$ inclinations each composed of $95$ satellites with an additional two orbital planes at $52^{\circ}$ inclinations each with $5$ satellites. The Walker constellation is motivated by the Skysat constellation from Planet Labs.  This constellation has $6$ orbital planes with $14$ satellites each at an $88^{\circ}$ inclination and an overlay of $2$ orbital planes at a $51.6^{\circ}$ inclination with $12$ satellites.  Each satellite has a memory capacity of $125$ GB and a single sensor that can slew to $60^\circ$  and $45^\circ$ off-nadir for the Planet and Walker constellations respectively.  Figure \ref{fig:constellation} depicts these two constellations.

\subsubsection{Downlinks}
A $62.5$ MB/s constant bit stream models the satellite downlink during visibility periods with a ground station. We incorporate two ground stations: the ASF Near Space Network Satellite Tracking Ground Station and the Guam Remote Ground Terminal System. 

\subsubsection{Dynamic observation request campaigns}
The target set, $\mathcal{T}$, is $634$ globally distributed cities. Dynamic campaigns consist of periodic requests of these targets. For large problem instances, a periodicity is uniformly sampled from the range $[5, 12]$. A target with periodicity $p$ is requested to be observed once within $p$ evenly spaced intervals. For small problem instances, we fix the periodicity at $3$. The scheduling horizon is set to be one day. However, the start of the horizon is randomly initialized. An observation's memory consumption is sampled from a normal distribution with mean $50$ MB and standard deviation $10$ MB. The time interval required to execute an observation is $63$s, accounting for imaging, slewing, and processing. 

To generate dynamics, we select a parameter $v$. This determines the number of changes during the horizon. The  time of the dynamics is uniformly distributed over the final $1 - \frac{2}{3\cdot v}$ of the scheduling horizon. This ensures that dynamics do not occur directly after the start of the horizon. The set of starting requests is randomly selected from the request set and initialized to one-third of the total size. We select from the remaining requests a random set of $\frac{2}{3\cdot v}$ to add. From the active requests, we randomly remove $\frac{1}{3\cdot v}$ of them. We enforce that once a request is removed, it is not added back, and requests are not changed after their execution horizon starts. We sample $v \in [3,5]$ uniformly.

\subsection{Hyperparameters and execution environment}

The stochastic update of D-NSS and D-DSA relies on the probability, $P_u$ that determines when an agent should unassign from a task.  We fix this value at $0.7$, as published in previous work \citep{shreya,zilberstein2025decentralized}. The $maxIters$ parameter is set to $20$. For the GND heuristic, we set all hyperparameters as described in previous work, and specifically use the GND(2) heuristic \citep{zilberstein2025decentralized}. We use a random seed of $2005$ for the initial scenario generation. When evaluating $N$ scenarios, we increment the seed for each subsequent scenario. The random seed used in the $\textsc{repair}$ procedure is initialized to $1$. The NSS and DSA algorithms are seeded with a value of $1234$, and the random seed for GND is set to $2$. The random scheduling algorithm uses a seed of $2023$. All experiments are executed using Java 19 on a MacBook Pro 16 laptop with an M2 Max processor (12-core CPU and 38-core GPU) and 64 GB of RAM.

\subsection{Evaluating D-NSS}

We first evaluate D-NSS in the absence of metareasoning (\textit{e.g.}, with an oracle always returning 1). 

\subsubsection{Results on small problem instances}
We are able to obtain an optimal solution for small problem instances using the omniscient offline algorithm. We solve this DCOP with a centralized branch and bound to obtain an optimal schedule for each satellite. For the Planet constellation, we solve campaigns of up to $500$ requests. For the Walker constellation, this increases to $1000$ requests. In addition to having fewer agents, the Walker constellation geometry under-constrains small problems, making finding optimal solutions faster. An optimal solver does not consistently terminate for larger problems.  We report the average gap in satisfaction percentage to the optimal solution for $10$ dynamic small problem instances in Tables \ref{tab:gaptooptimalplanet} and \ref{tab:gaptooptimalwalker}. 

These results support the theoretical analysis of D-NSS and demonstrate near-optimal performance. D-NSS outperforms all baselines, including D-DSA, 0-DSA, and 0-NSS, and does so while using less computation and communication. Notably, compared to DSA variants, D-NSS finds better solutions using an order of magnitude less computation and up to two orders of magnitude less communication. This is due to both the lower theoretical complexity per iteration and the faster convergence of D-NSS.

\begin{table}[t!]
    \small
    \centering
    \begin{tabular}{@{}lccc@{}}
        \toprule
        \textbf{Algorithm} & \textbf{Optimality Gap (\%)} & \textbf{Time (ms)} & \textbf{Messages (KB)} \\
        \midrule
        Random & 2.530 & $<$1 & 0 \\
        Greedy & 8.373 & $<$1 & 0 \\
        \midrule
        \textbf{D-NSS} & \textbf{1.867} & \textbf{$<$1} & \textbf{7.3} \\
        0-NSS & 2.590 & $<$1 & 13.2 \\
        D-DSA & 4.217 & 5.3 & 980.6 \\
        0-DSA & 3.795 & 4.7 & 718.4 \\
        \bottomrule
    \end{tabular}
    \caption{Results on 10 dynamic problems for the Planet constellation (up to 500 requests).}
    \label{tab:gaptooptimalplanet}
\end{table}

\begin{table}[t!]
    \small
    \centering
    \begin{tabular}{@{}lccc@{}}
        \toprule
        \textbf{Algorithm} & \textbf{Optimality Gap (\%)} & \textbf{Time (ms)} & \textbf{Messages (KB)} \\
        \midrule
        Random & 14.945 & $<$1 & 0 \\
        Greedy & 15.604 & $<$1 & 0 \\
        \midrule
        \textbf{D-NSS} & \textbf{0.142} & \textbf{5.2} & \textbf{240.2} \\
        0-NSS & 0.480 & 6.1 & 400.0 \\
        D-DSA & 1.165 & 58.2 & 10,459.5 \\
        0-DSA & 1.215 & 54.0 & 7,789.0 \\
        \bottomrule
    \end{tabular}
    \caption{Results on 10 dynamic problems for the Walker constellation (up to 1000 requests).}
    \label{tab:gaptooptimalwalker}
\end{table}

\subsubsection{Results on large problem instances}

\begin{figure*}[t!]
   
  \begin{subfigure}{0.4\textwidth}
    \includegraphics[width=\linewidth]{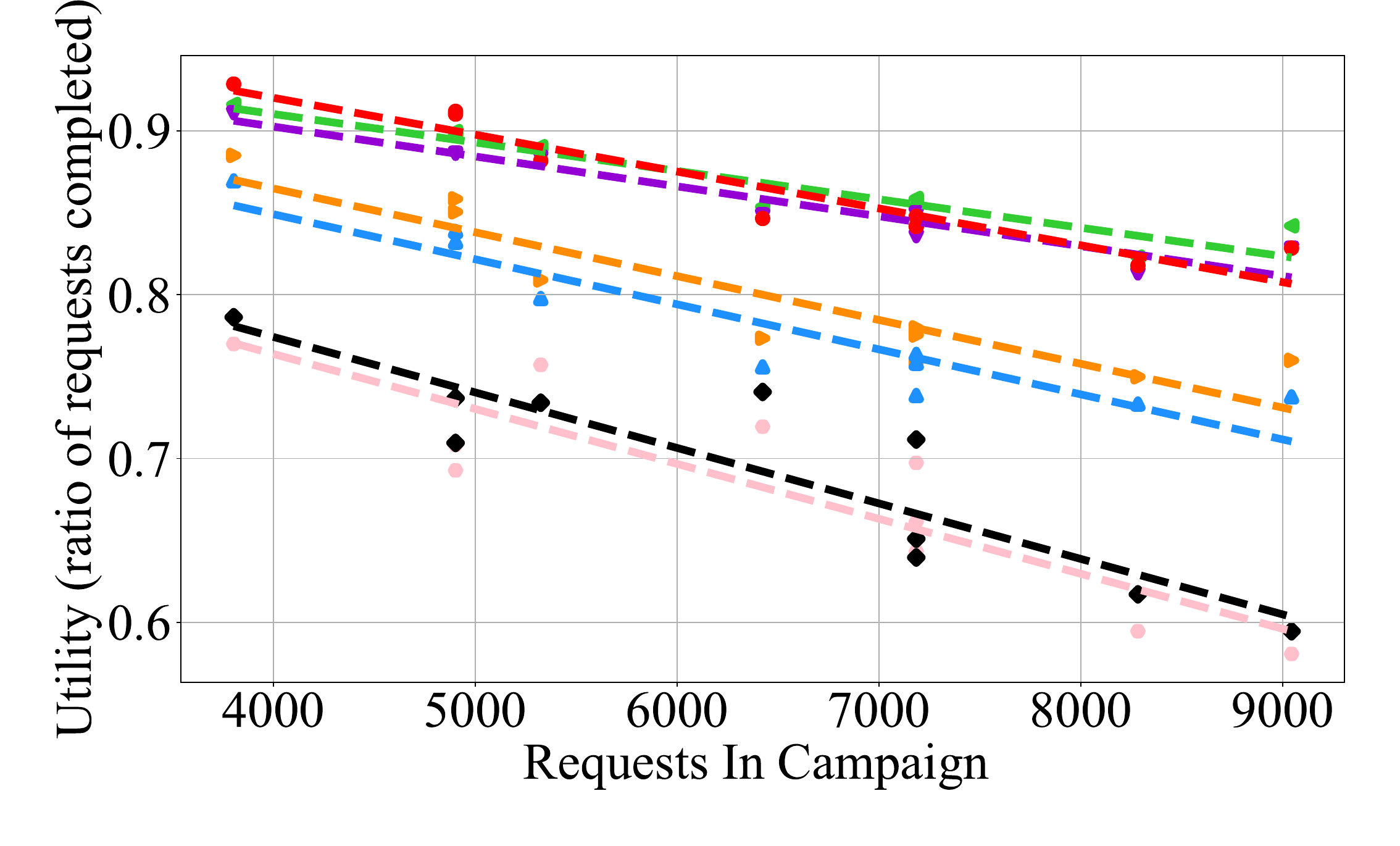}
    \caption{Planet satisfaction \% as a function of problem size.} \label{fig:1a}
  \end{subfigure}%
  \hspace*{\fill}   
  \begin{subfigure}{0.4\textwidth}
    \includegraphics[width=\linewidth]{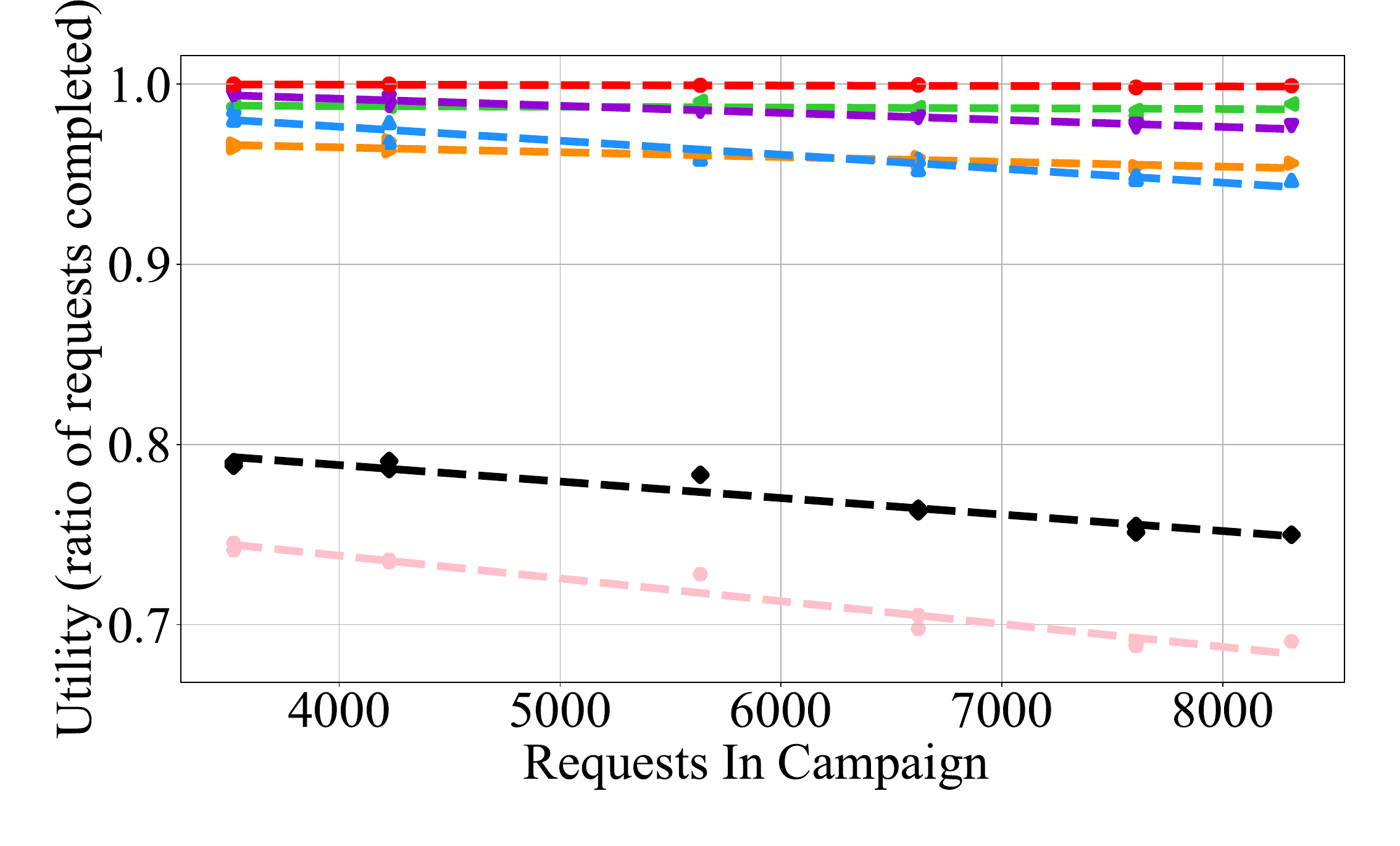}
    \caption{Walker satisfaction \% as a function of problem size.} \label{fig:1b}
  \end{subfigure}%
  
  \begin{subfigure}{0.4\textwidth}
    \includegraphics[width=\linewidth]{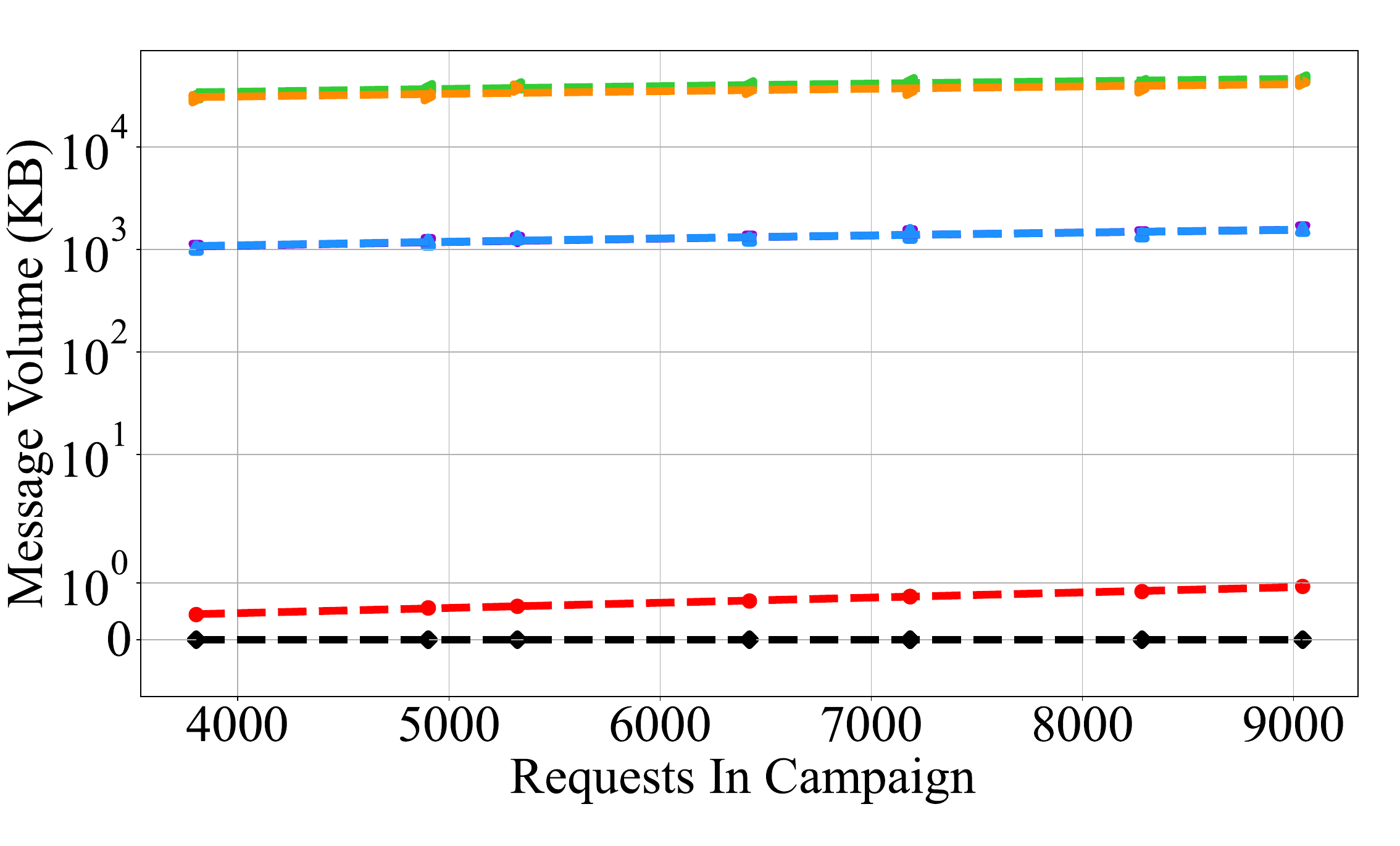}
    \caption{Planet message volume as a function of problem size.} \label{fig:1c}
  \end{subfigure}%
    \hspace*{\fill}   
  \begin{subfigure}{0.4\textwidth}
    \includegraphics[width=\linewidth]{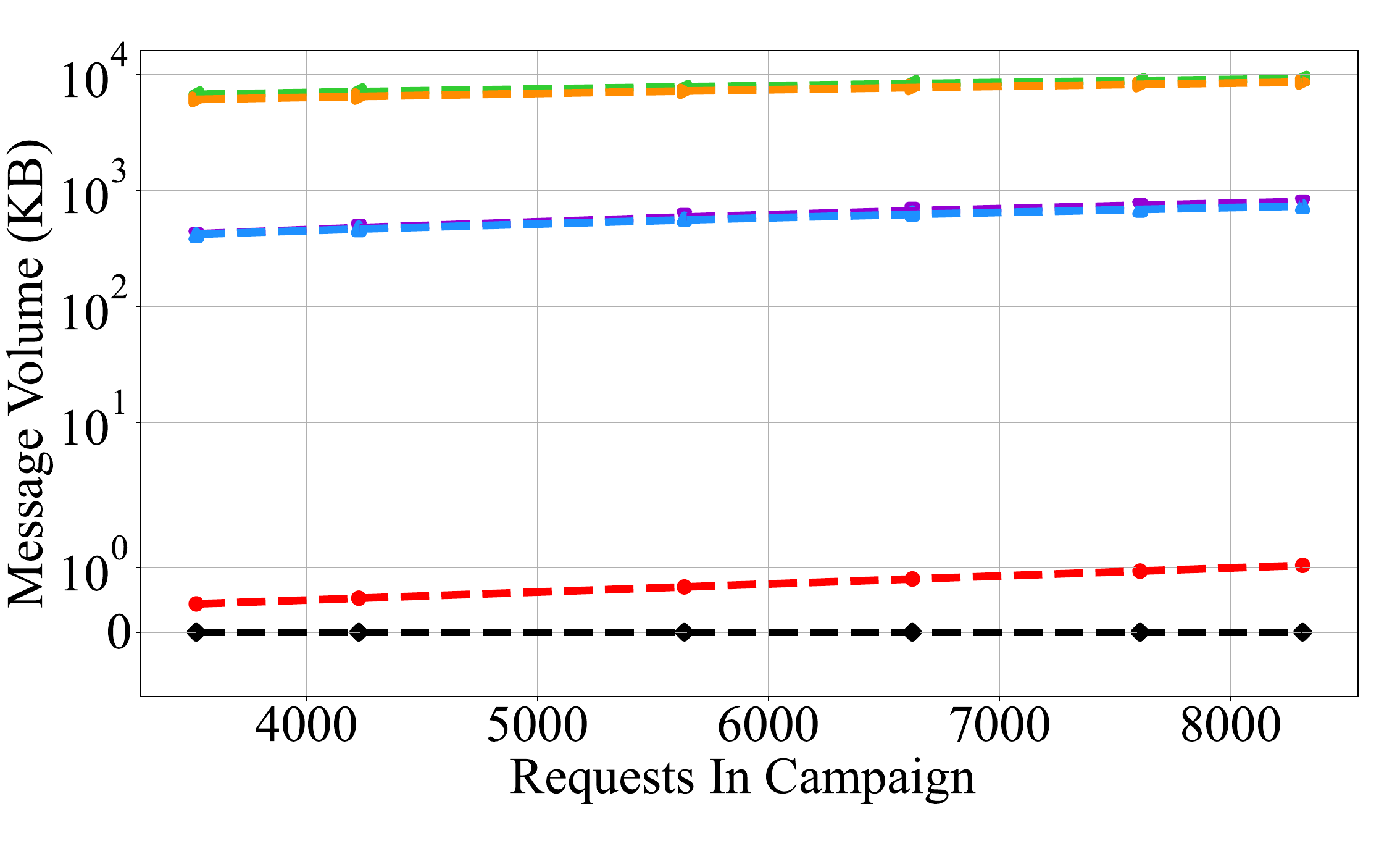}
    \caption{Walker message volume as a function of problem size.} \label{fig:1d}
  \end{subfigure}%

  \begin{subfigure}{0.4\textwidth}
    \includegraphics[width=\linewidth]{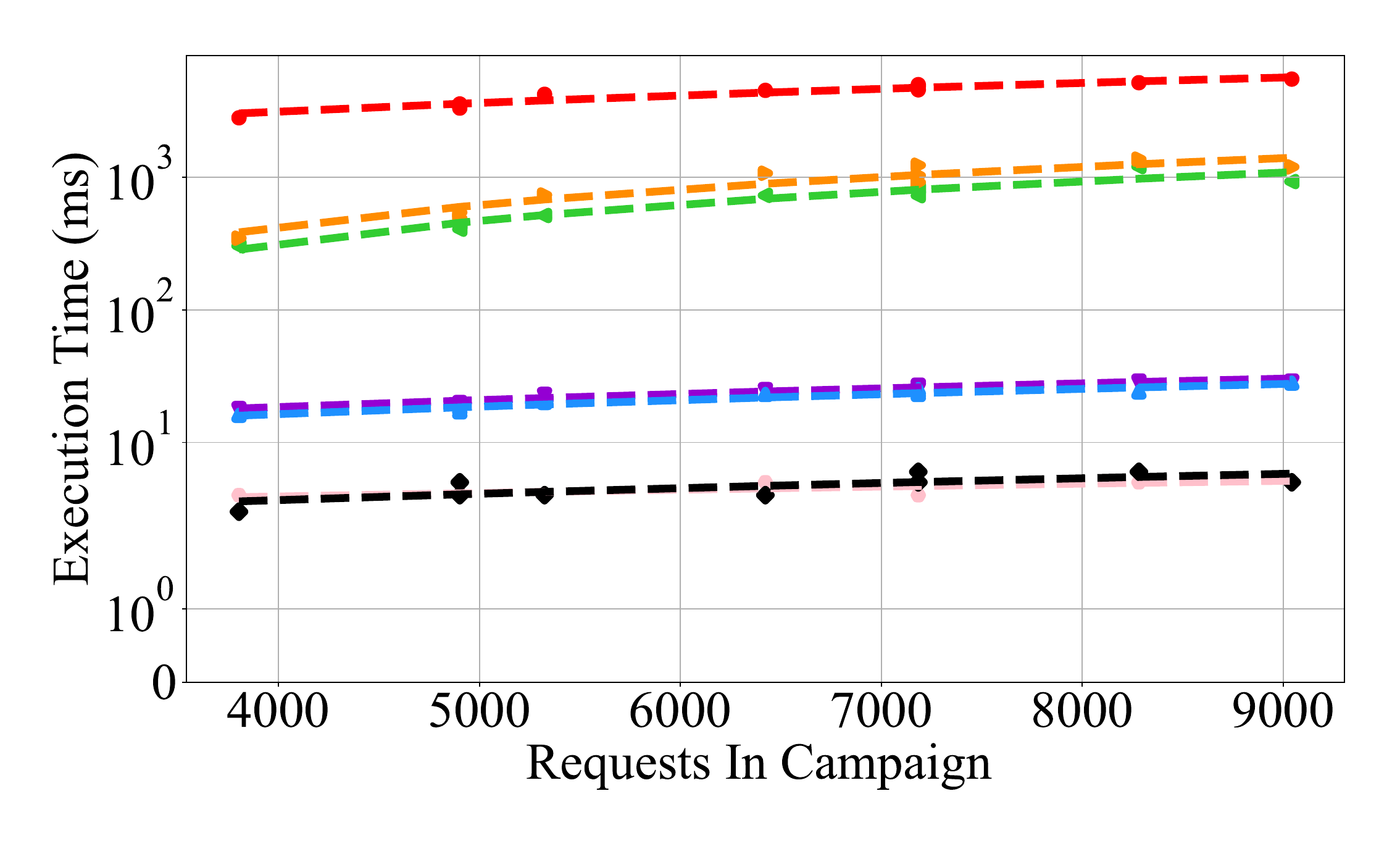}
    \caption{Planet execution time as a function of problem size.} \label{fig:1e}
  \end{subfigure}
      \hspace*{\fill}   
    \hspace*{\fill}   
  \begin{subfigure}{0.4\textwidth}
    \includegraphics[width=\linewidth]{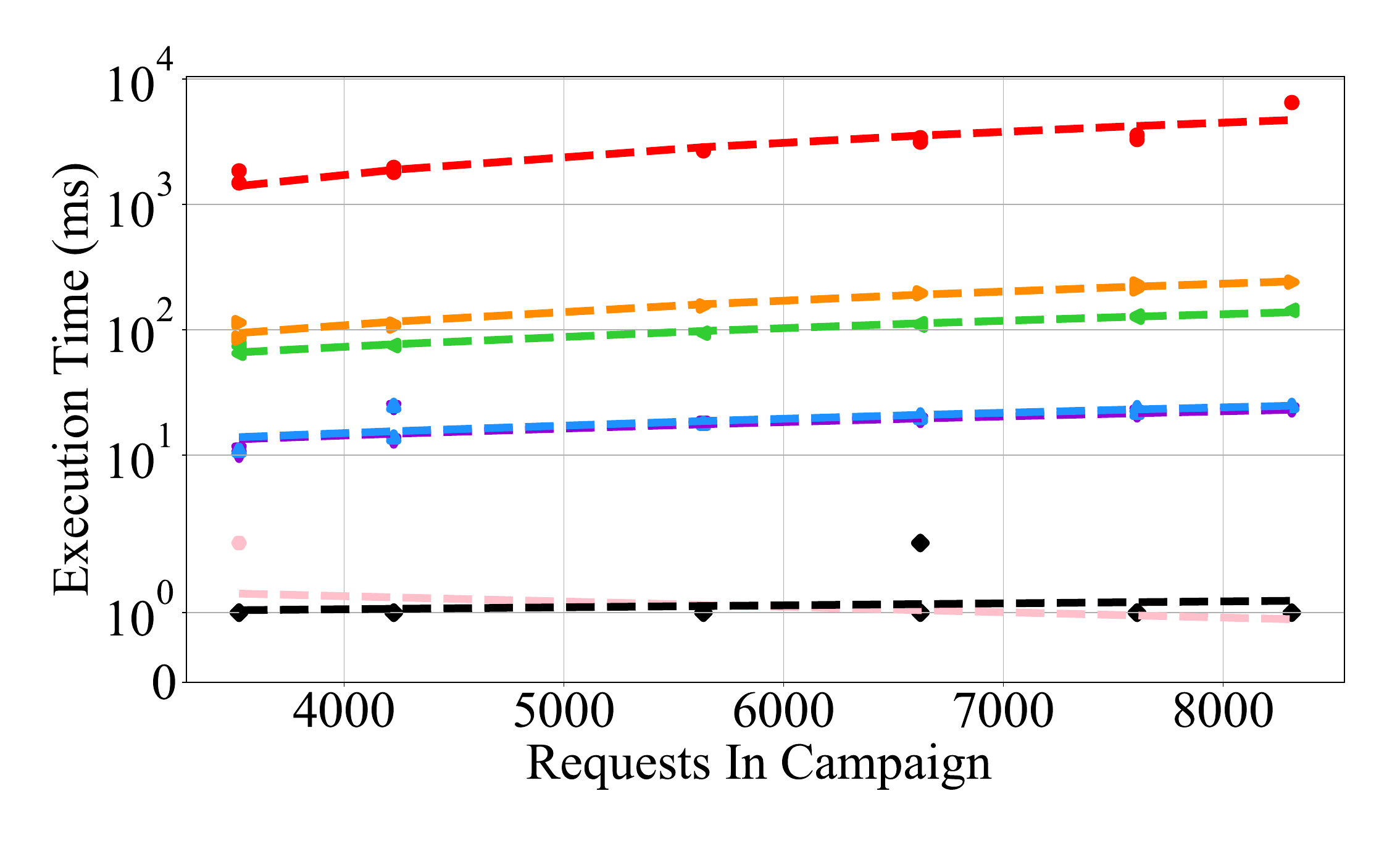}
    \caption{Walker execution time as a function of problem size.} \label{fig:1f}
  \end{subfigure}
  
\begin{center}
    \includegraphics[width=0.65\linewidth]{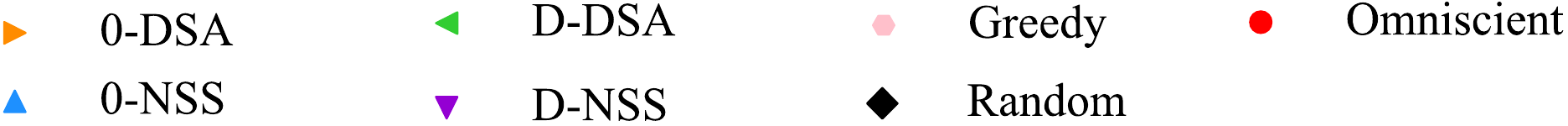}    
\end{center}

\caption{Results of large-scale simulations for the Planet (left column) and Walker (right column) constellations. We report the average satisfaction, total message volume, and per-agent runtime. Note the log scale for message volume and runtime.}
 \label{fig:largeresults}
\end{figure*}

\begin{figure*}
      \begin{subfigure}{0.4\textwidth}
    \includegraphics[width=\linewidth]{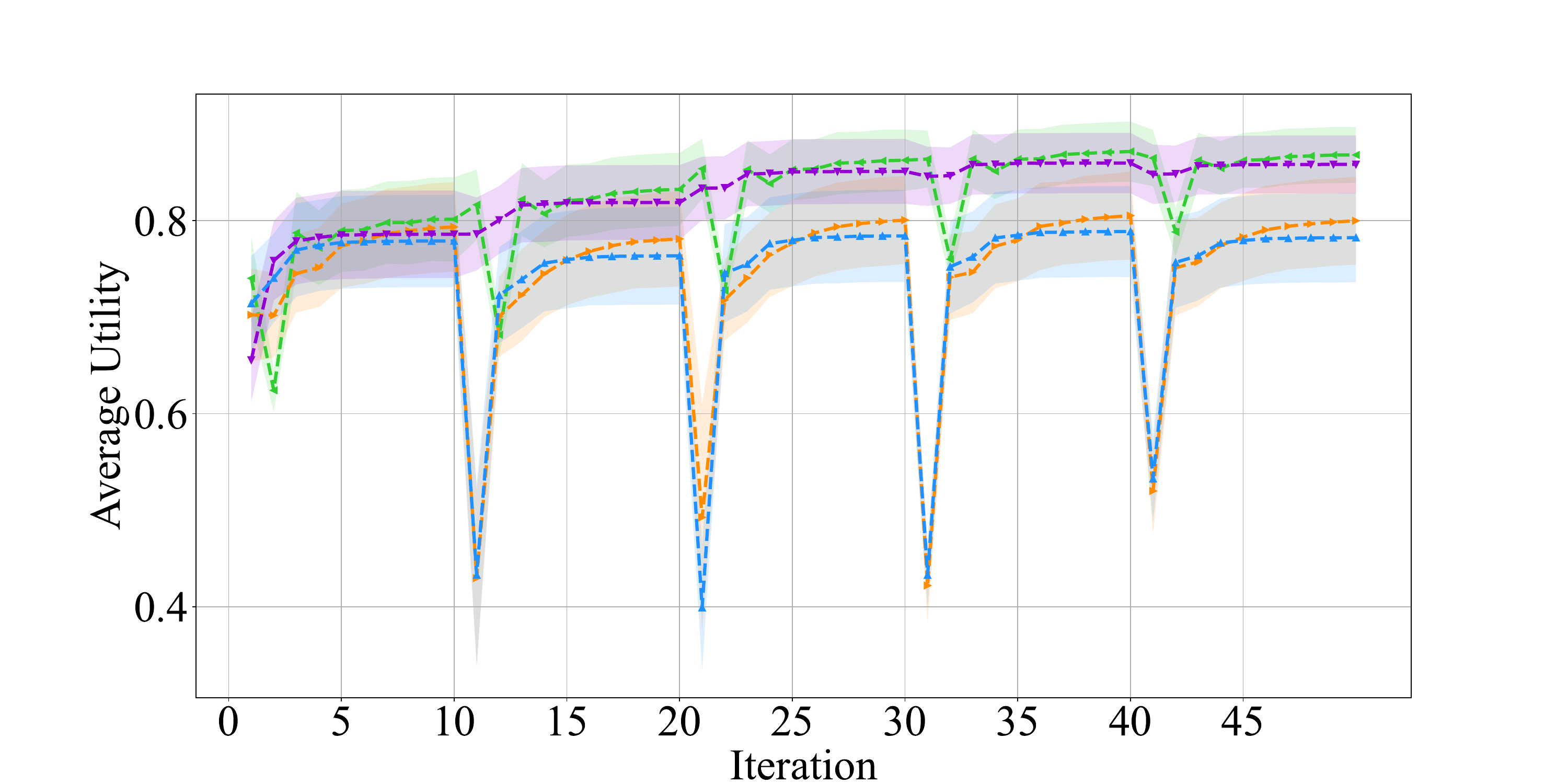}
    \caption{Satisfaction \% of iterative algorithms for the Planet constellation.} \label{fig:planet-iter}
  \end{subfigure}%
    \hspace*{\fill}   
  \begin{subfigure}{0.4\textwidth}
    \includegraphics[width=\linewidth]{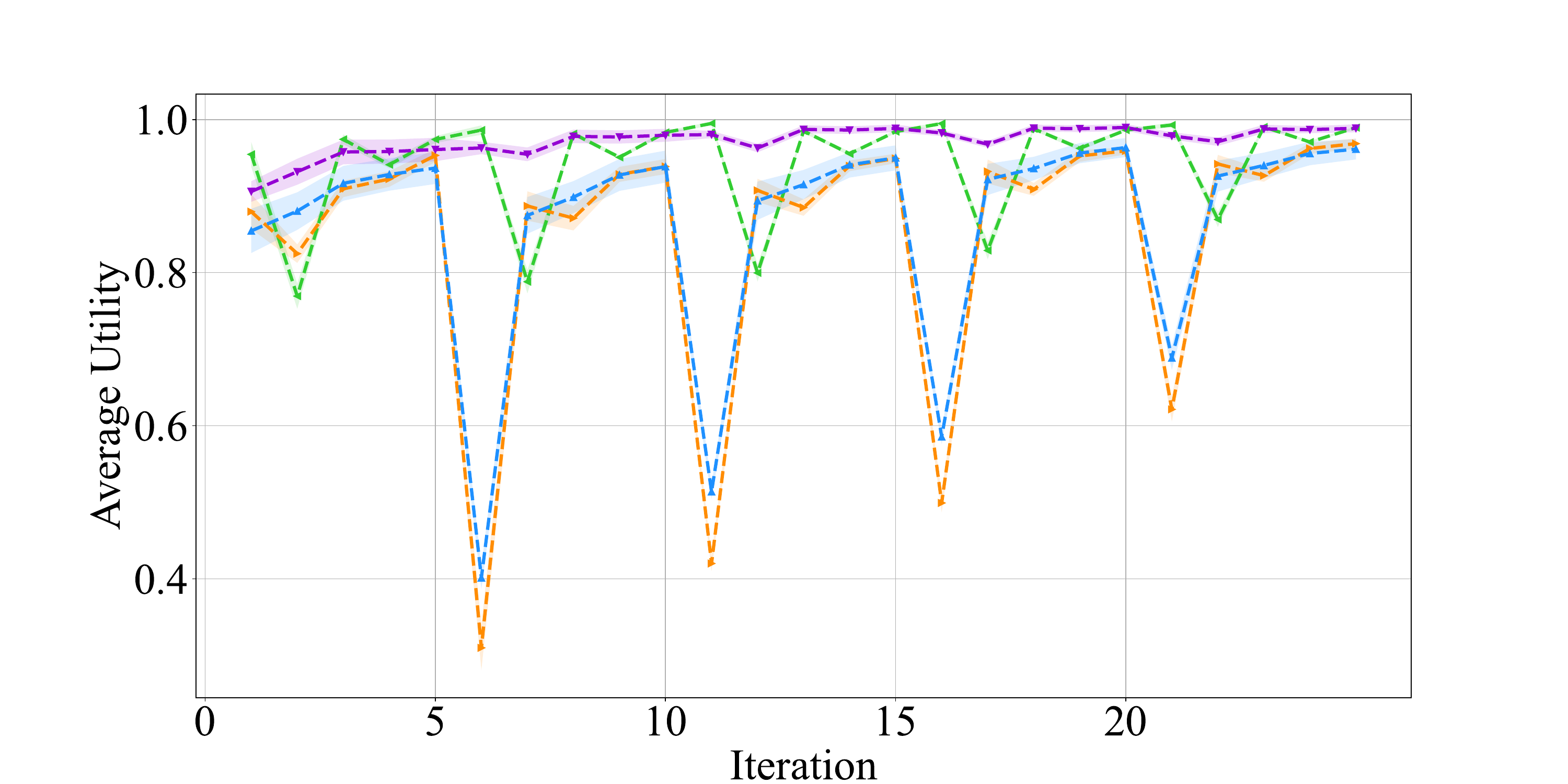}
    \caption{Satisfaction \% of iterative algorithms for the Walker constellation.} \label{fig:walker-iter}
  \end{subfigure}%

\caption{Results of large-scale simulations for the Planet (left) and Walker (right) constellations. We show the solution quality across a fixed number of iterations for iterative algorithms for instances with $v=5$. Shaded region shows standard deviation.}
 \label{fig:iters}
\end{figure*}

\begin{figure*}[h!]
   
  \begin{subfigure}{0.4\textwidth}
    \includegraphics[width=\linewidth]{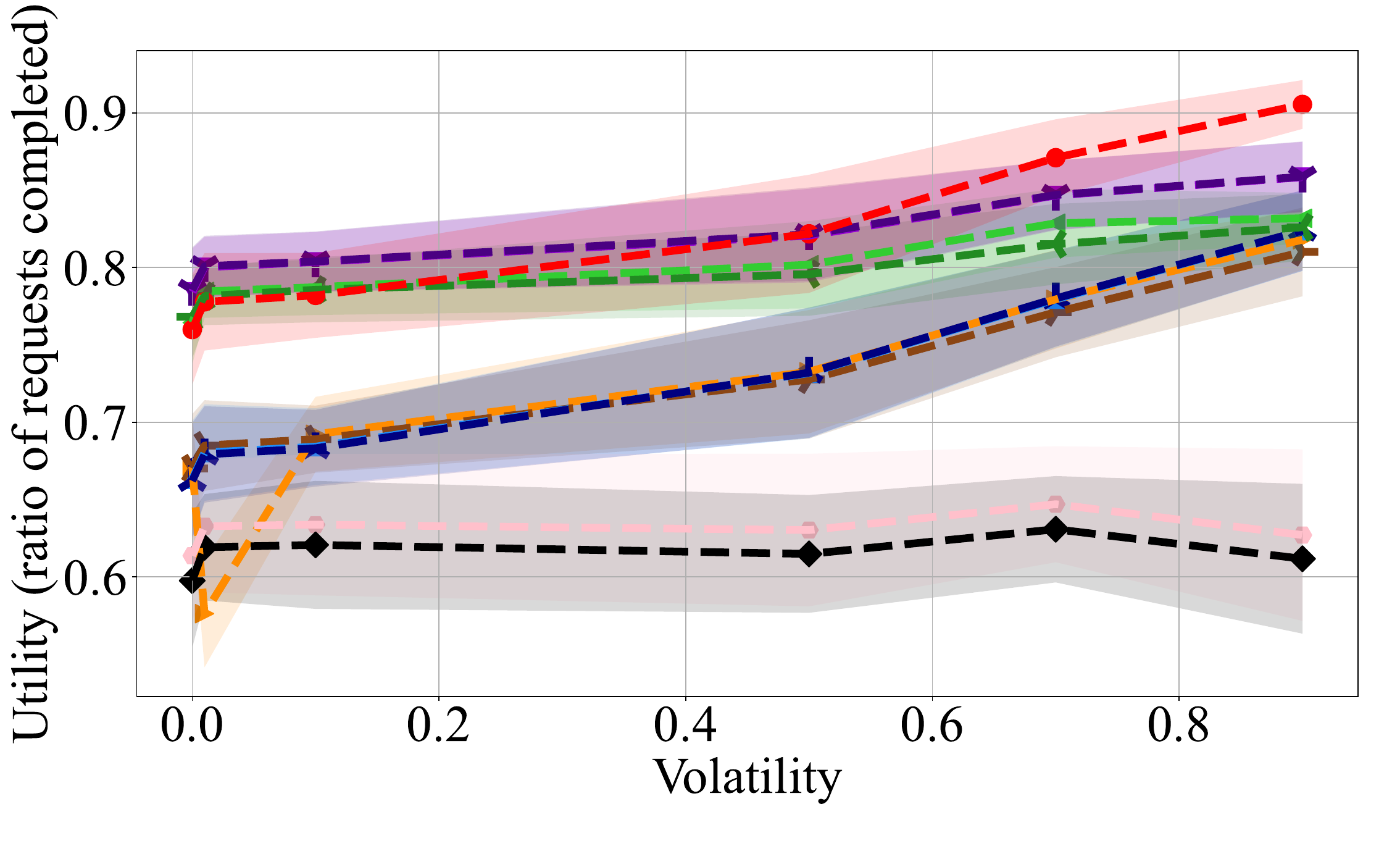}
    \caption{Planet satisfaction ratio as a function of volatility.} \label{fig:va}
  \end{subfigure}%
  \hspace*{\fill}   
  \begin{subfigure}{0.4\textwidth}
    \includegraphics[width=\linewidth]{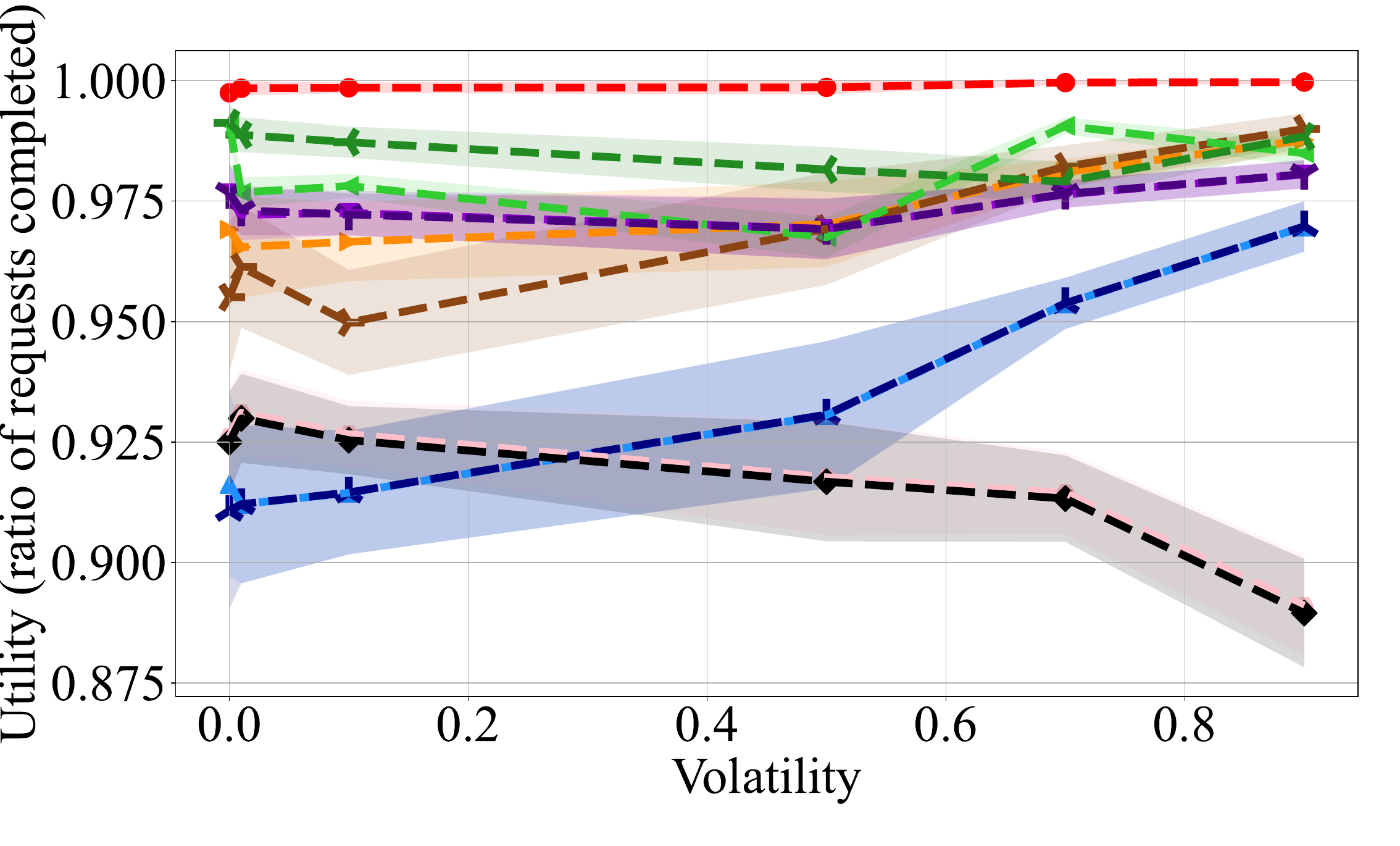}
    \caption{Walker satisfaction ratio as a function of volatility.} \label{fig:vb}
  \end{subfigure}%

  \begin{subfigure}{0.4\textwidth}
    \includegraphics[width=\linewidth]{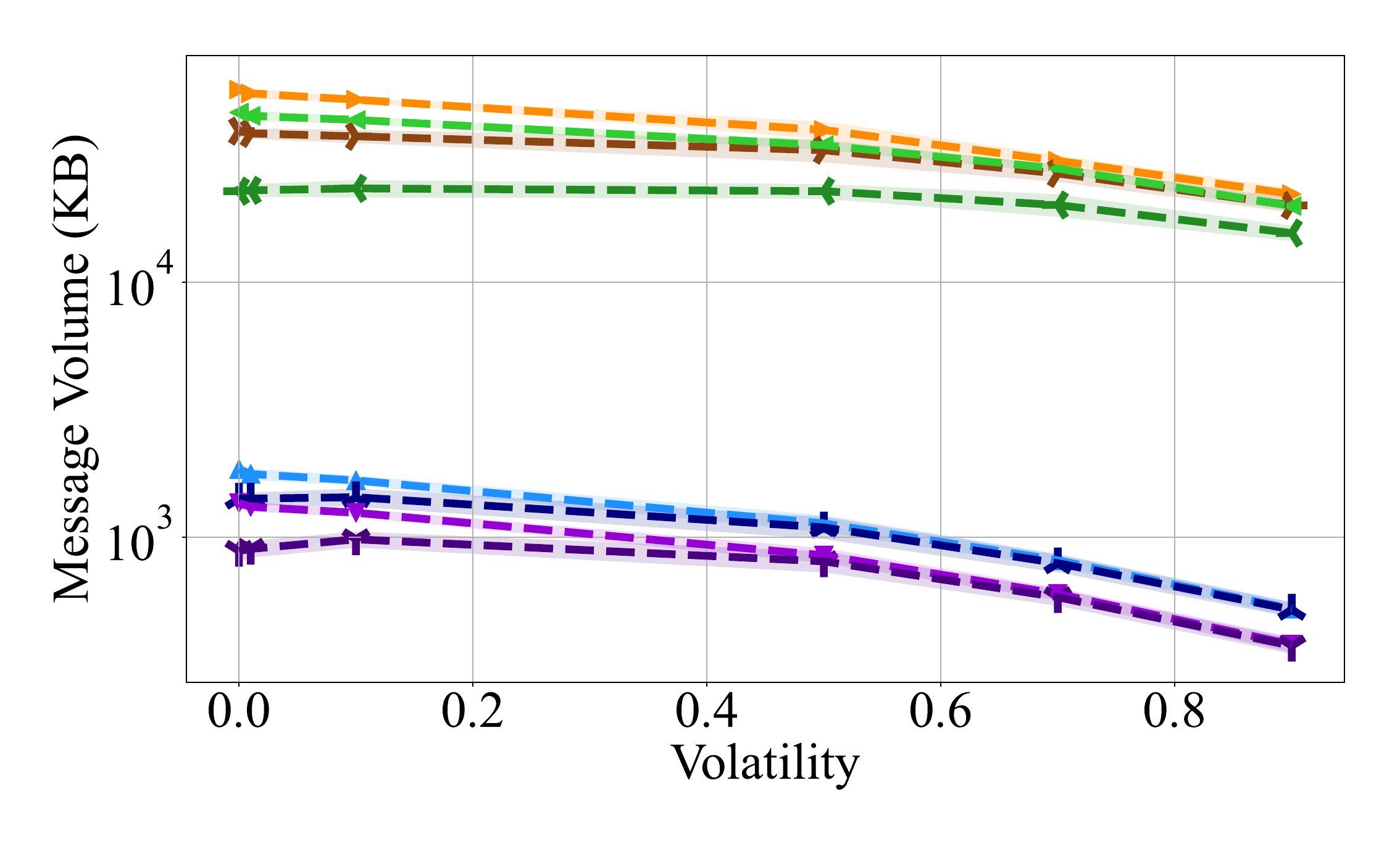}
    \caption{Planet message volume as a function of volatility.\\} \label{fig:vc}
  \end{subfigure}
\hspace*{\fill}   
  \begin{subfigure}{0.4\textwidth}
    \includegraphics[width=\linewidth]{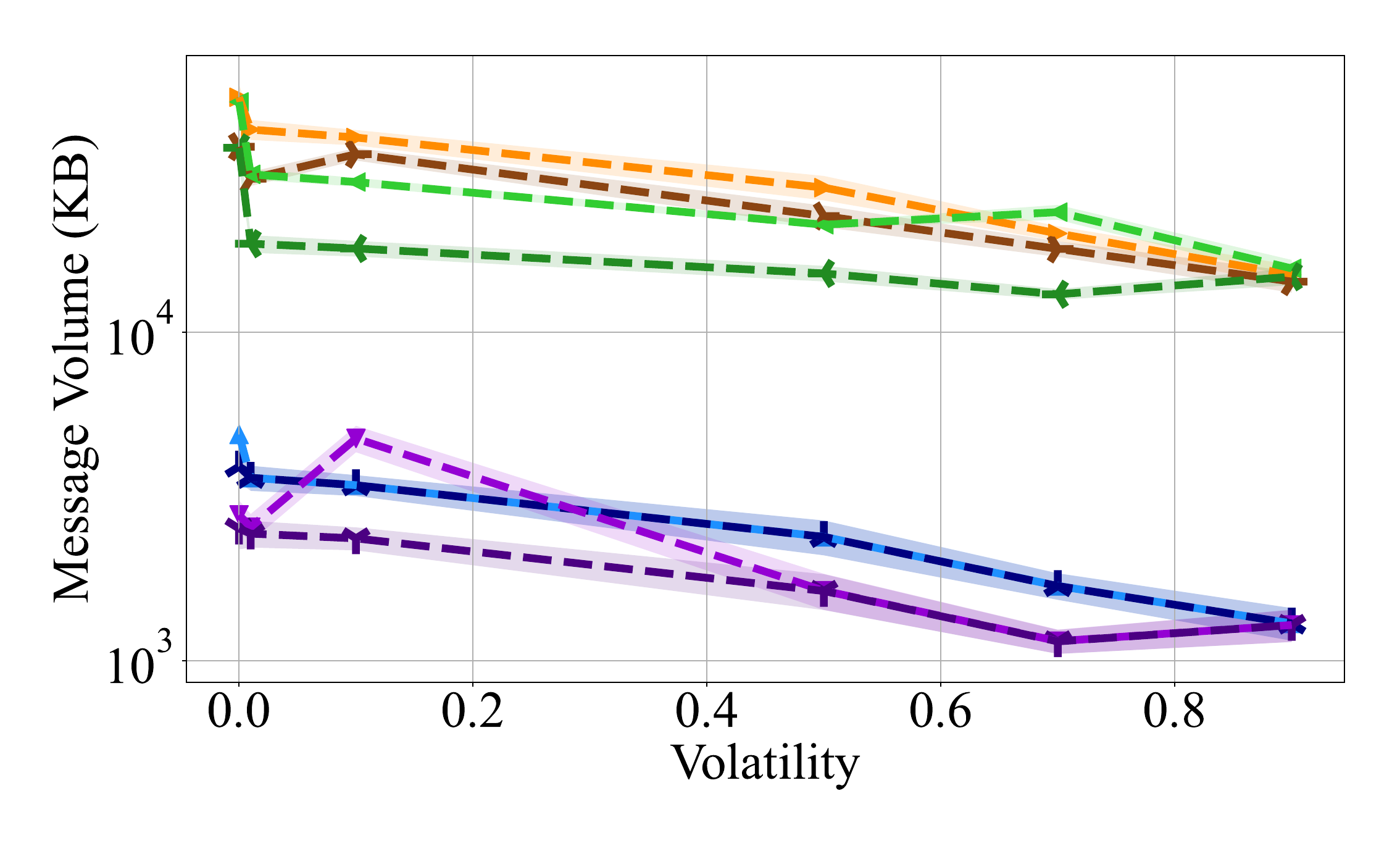}
    \caption{Walker message volume as a function of volatility.} \label{fig:vd}
  \end{subfigure}%

  \begin{subfigure}{0.4\textwidth}
    \includegraphics[width=\linewidth]{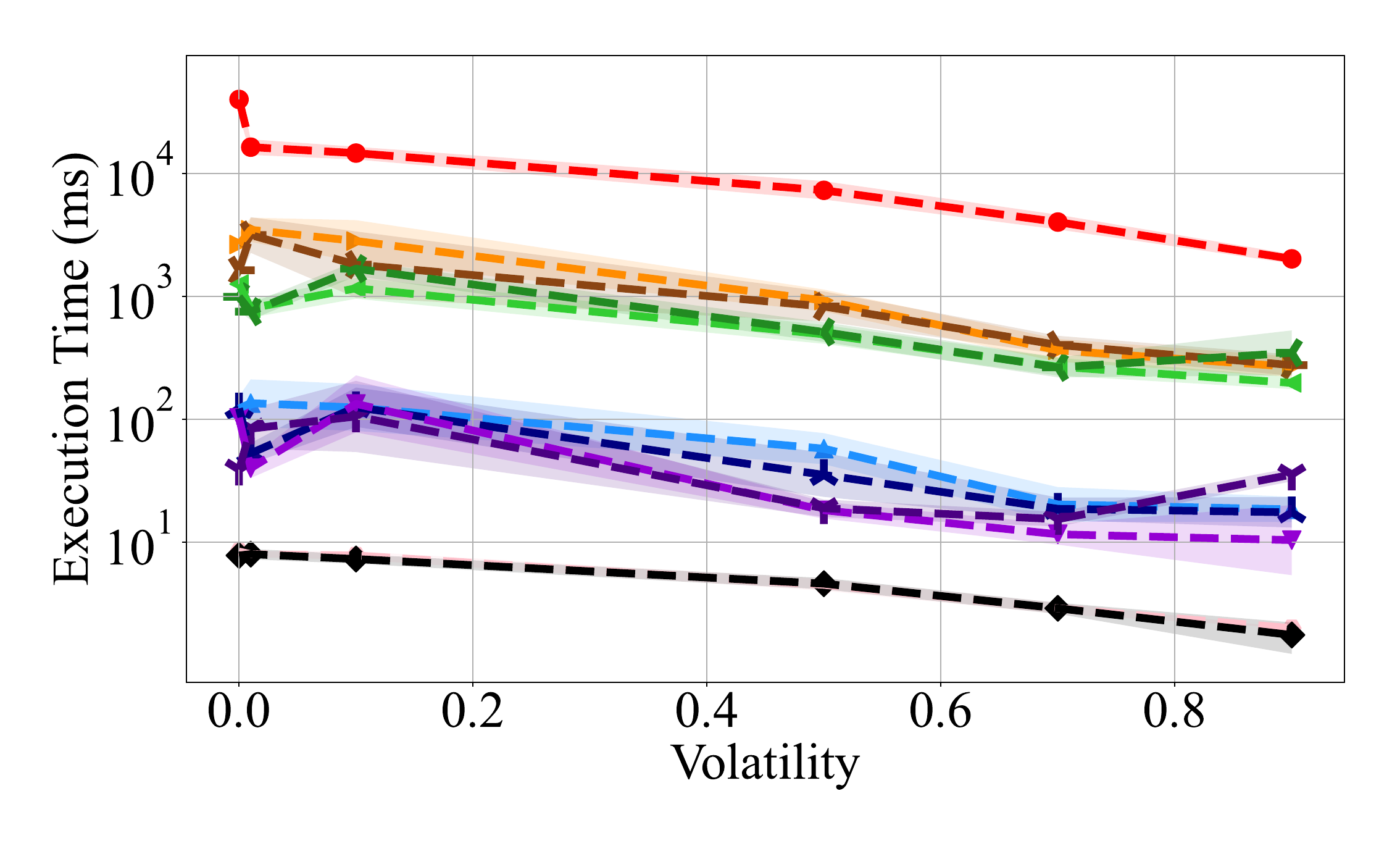}
    \caption{Planet execution time as a function of volatility.} \label{fig:ve}
  \end{subfigure}
\hspace*{\fill}   
  \begin{subfigure}{0.4\textwidth}
    \includegraphics[width=\linewidth]{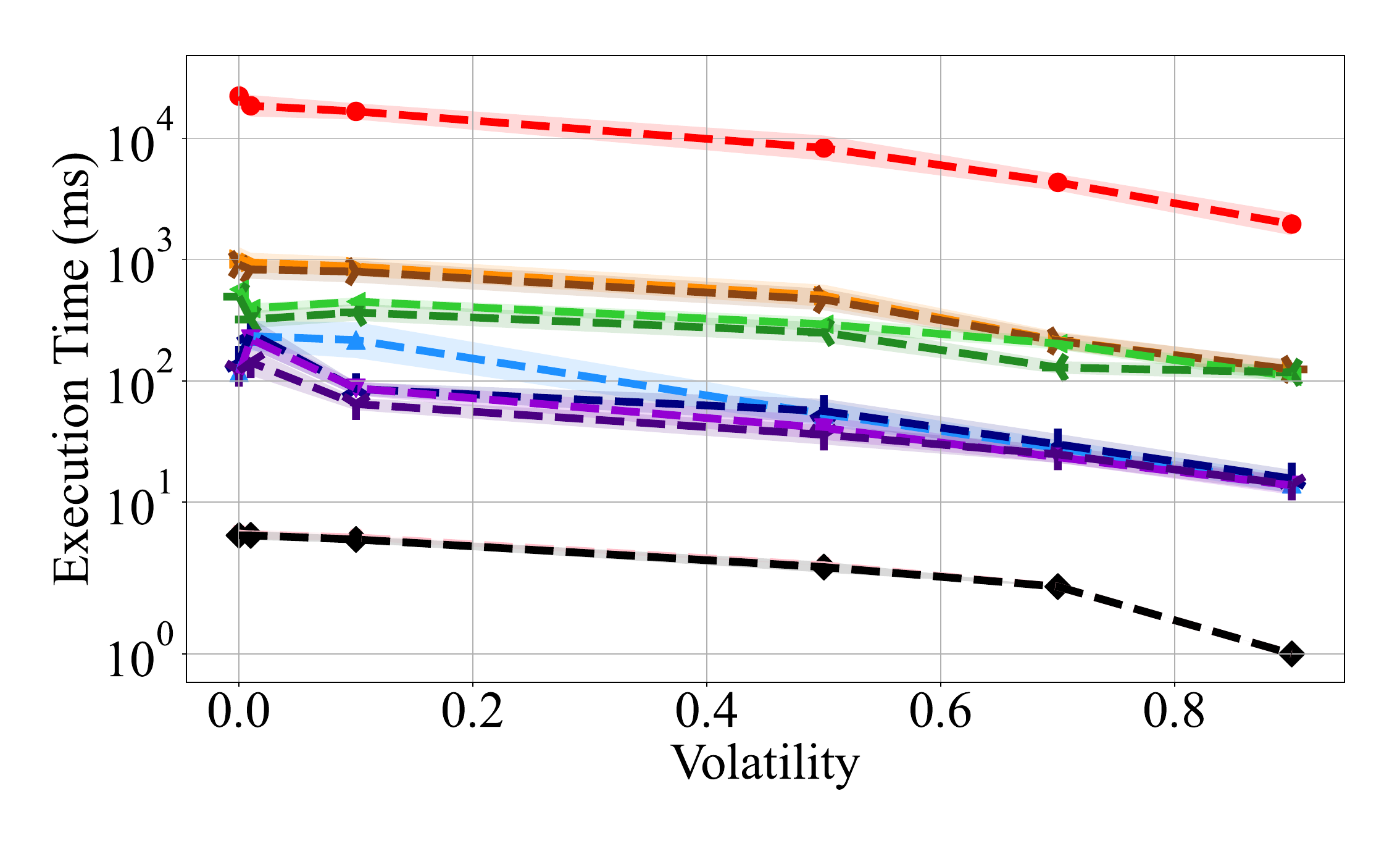}
    \caption{Walker execution time as a function of volatility.} \label{fig:vf}
  \end{subfigure}%

  \begin{center}
    \includegraphics[width=0.4\linewidth]{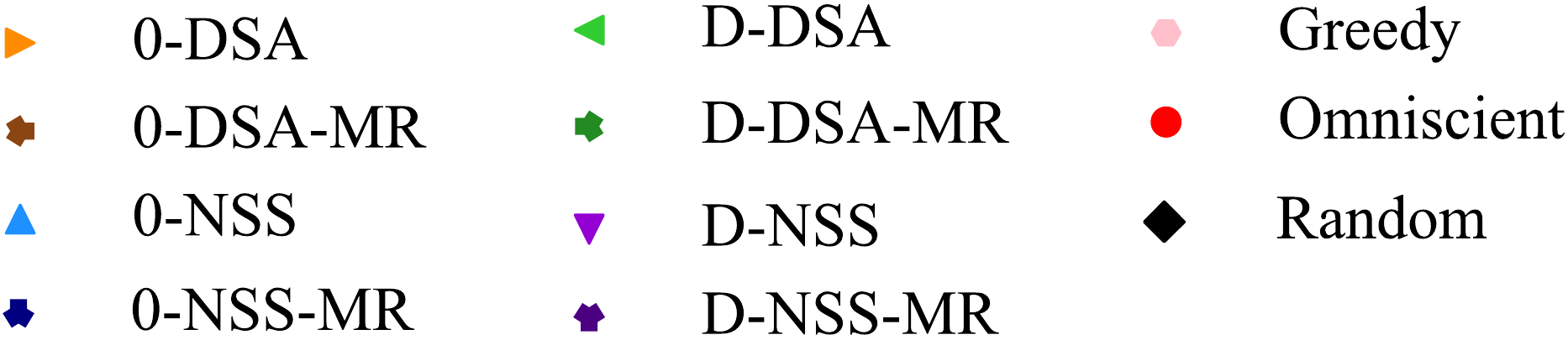}
\end{center}

\caption{Results of large-scale simulations for the Planet (left column) and Walker (right column) constellations with increasing volatility. We report the ratio of request satisfaction, total message volume, and per-agent runtime. Note the log scale for message volume and runtime. For NSS and DSA variants, we evaluate each with and without metareasoning. Shaded region shows standard deviation.}
 \label{fig:volatility}
\end{figure*}

\begin{figure*}[htbp!]
   
  \begin{subfigure}{0.4\textwidth}
    \includegraphics[width=\linewidth]{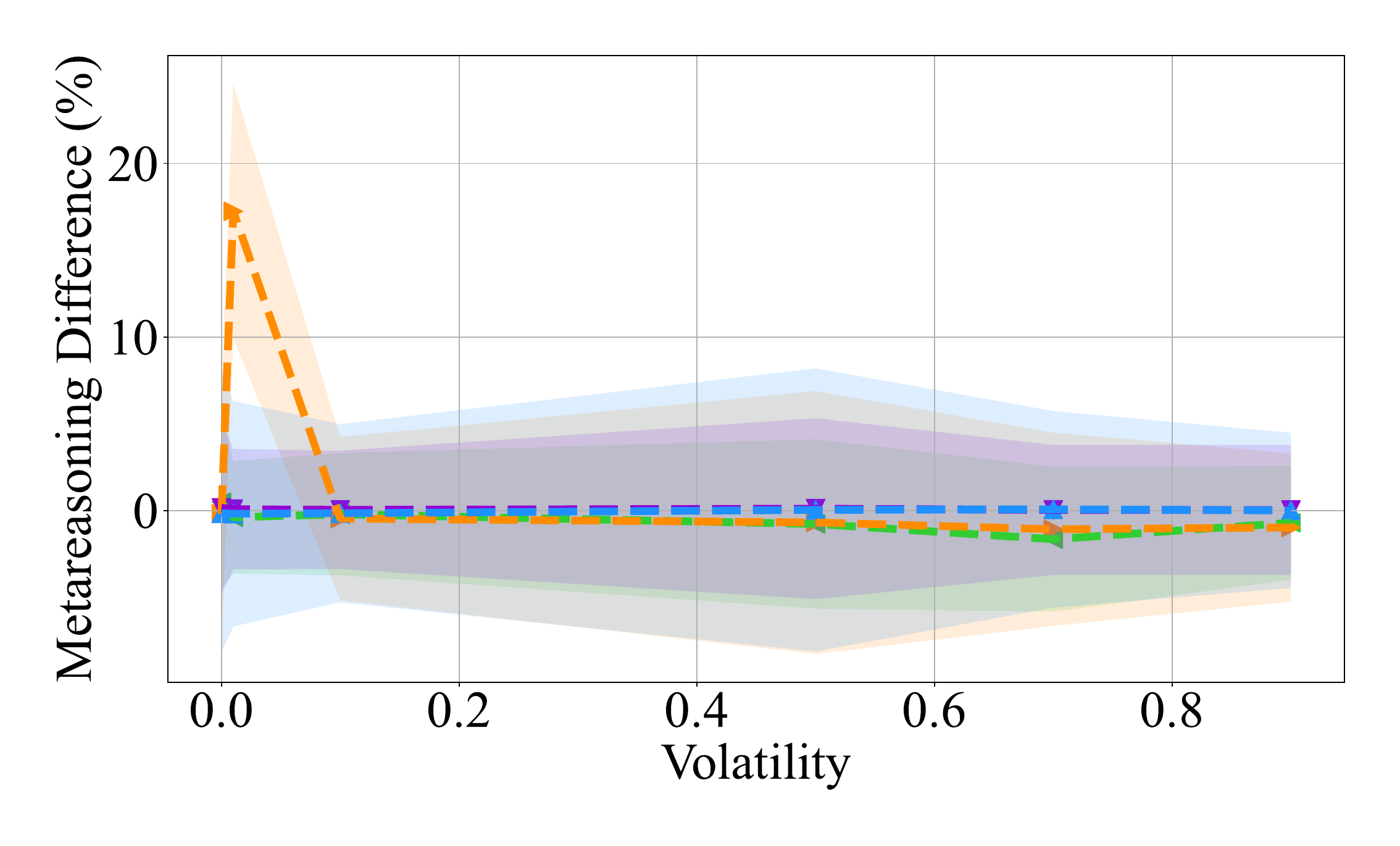}
    \caption{Planet percent difference utility when using metareasoning as a function of volatility.} \label{fig:vda}
  \end{subfigure}%
  \hspace*{\fill}   
  \begin{subfigure}{0.4\textwidth}
    \includegraphics[width=\linewidth]{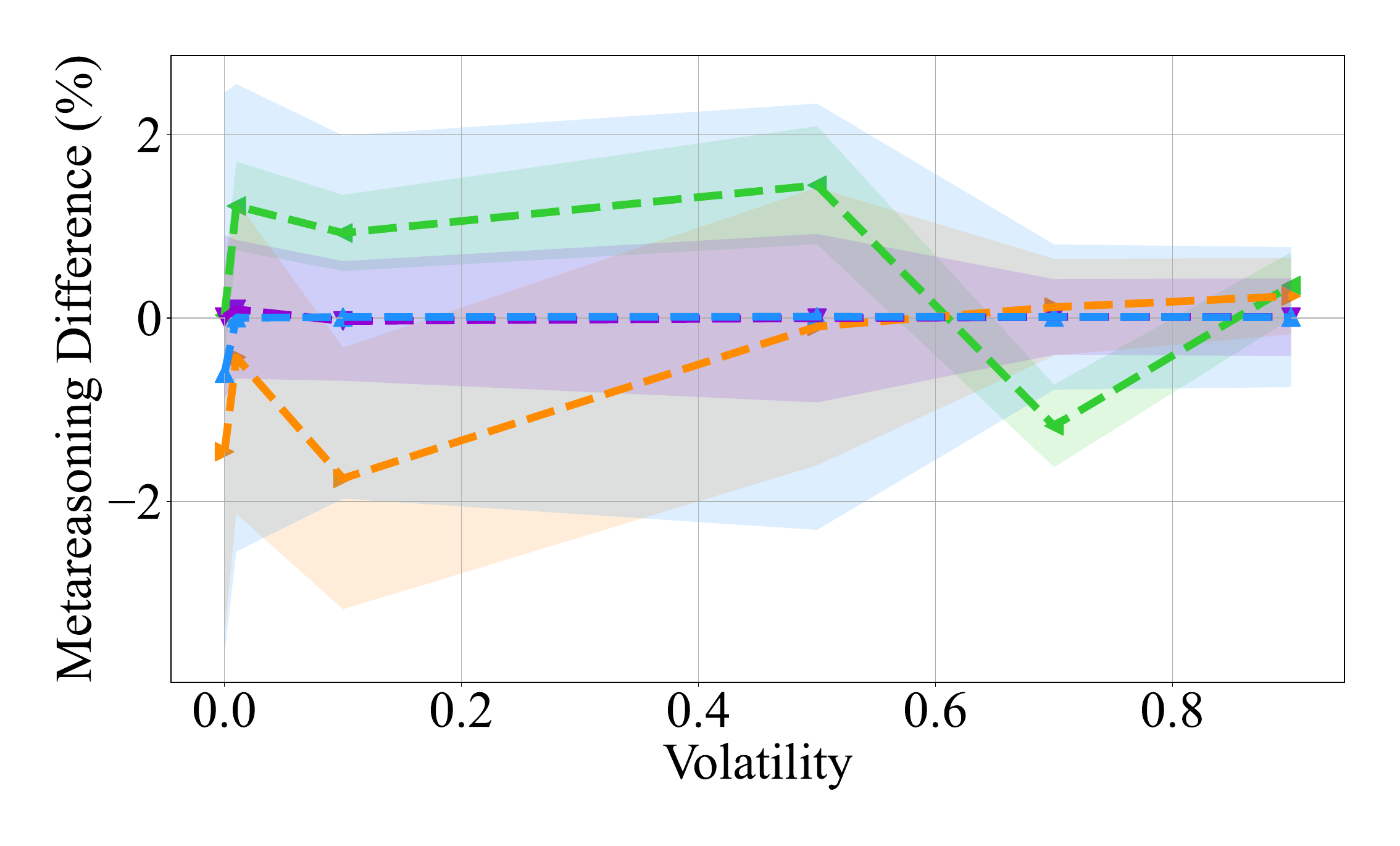}
    \caption{Walker percent difference utility when using metareasoning as a function of volatility.} \label{fig:vdb}
  \end{subfigure}%

  \begin{subfigure}{0.4\textwidth}
    \includegraphics[width=\linewidth]{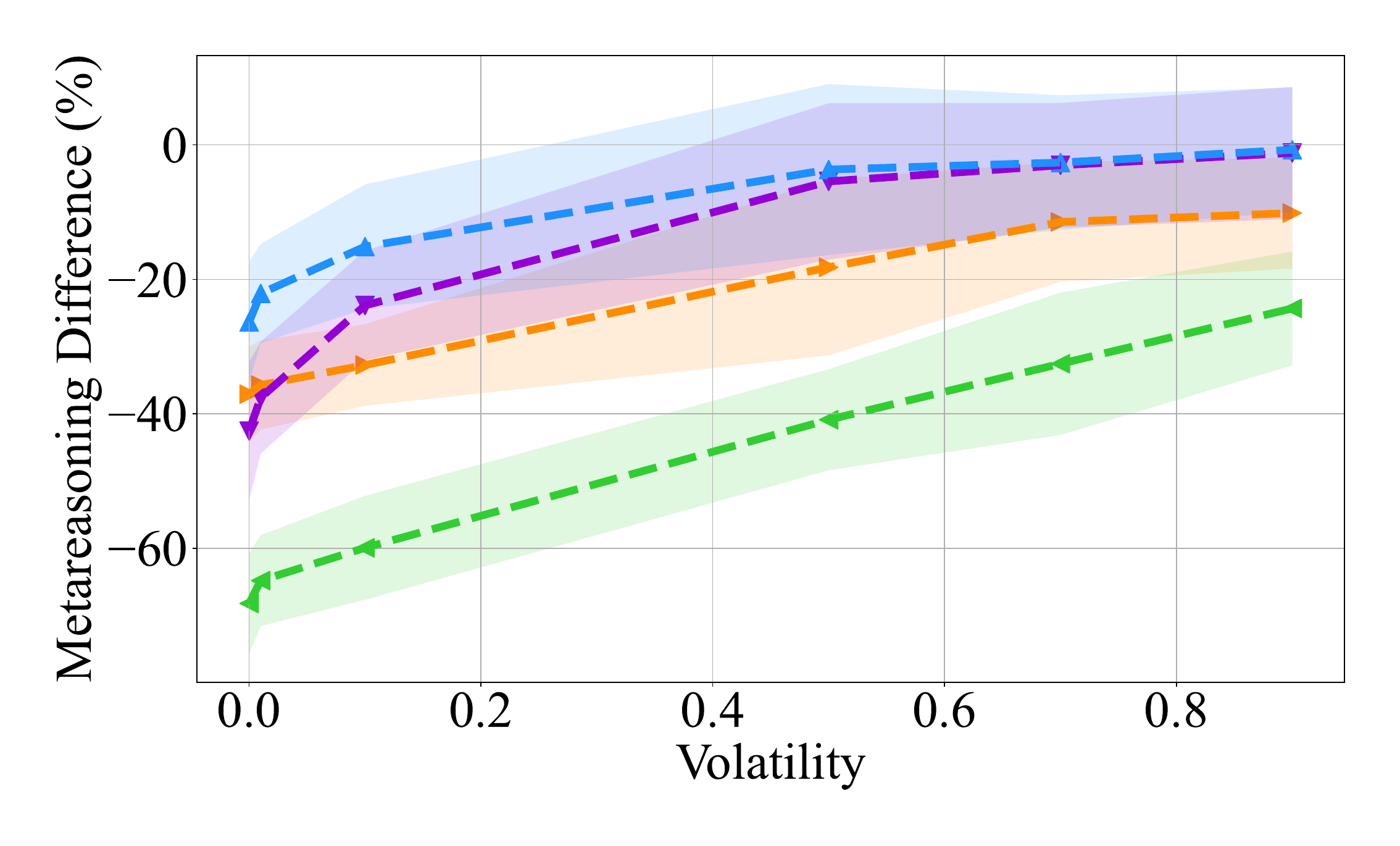}
    \caption{Planet percent difference message volume when using metareasoning as a function of volatility} \label{fig:vdc}
  \end{subfigure}
\hspace*{\fill}   
  \begin{subfigure}{0.4\textwidth}
    \includegraphics[width=\linewidth]{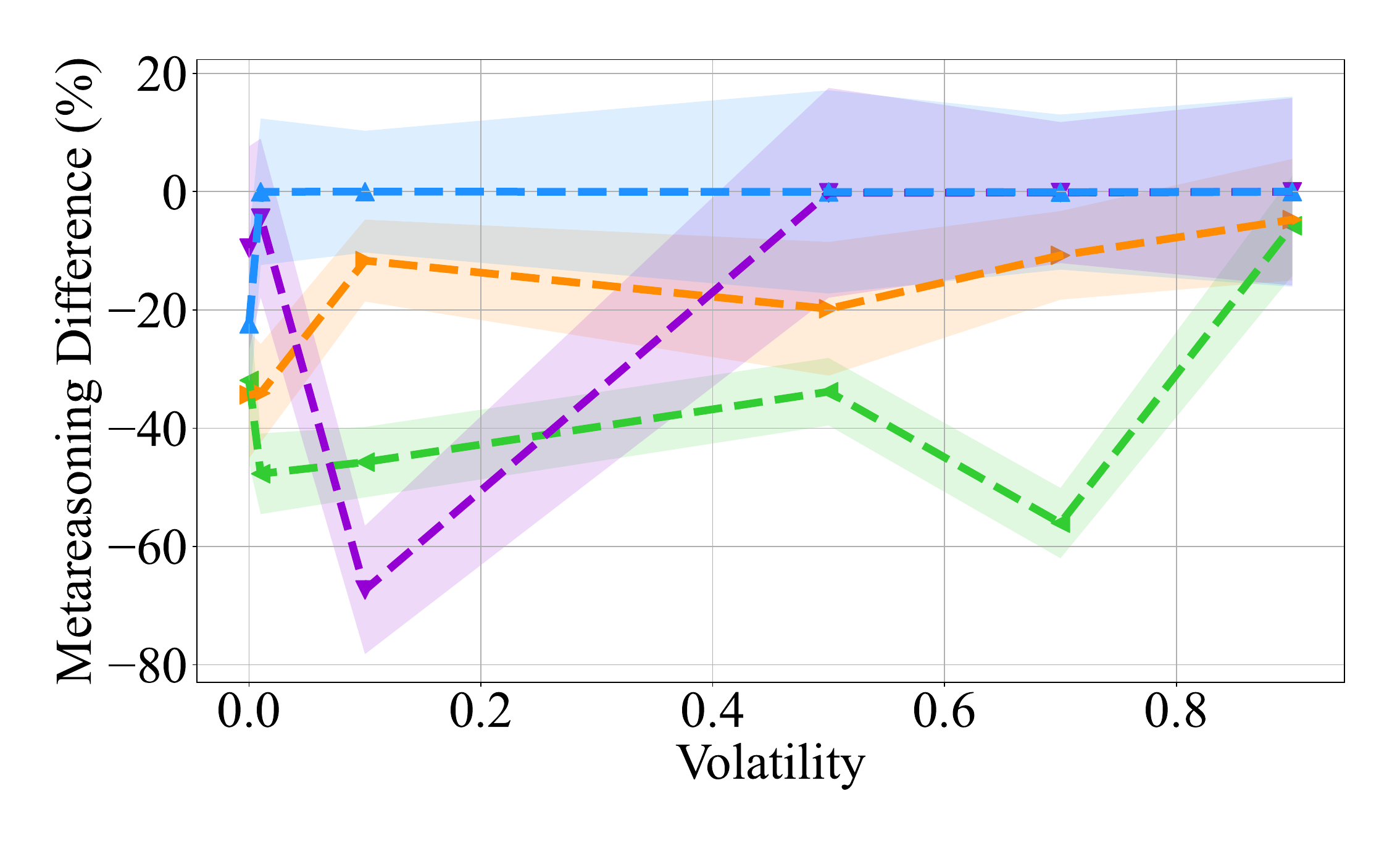}
    \caption{Walker percent difference message volume when using metareasoning as a function of volatility} \label{fig:vdd}
  \end{subfigure}%

  \begin{subfigure}{0.4\textwidth}
    \includegraphics[width=\linewidth]{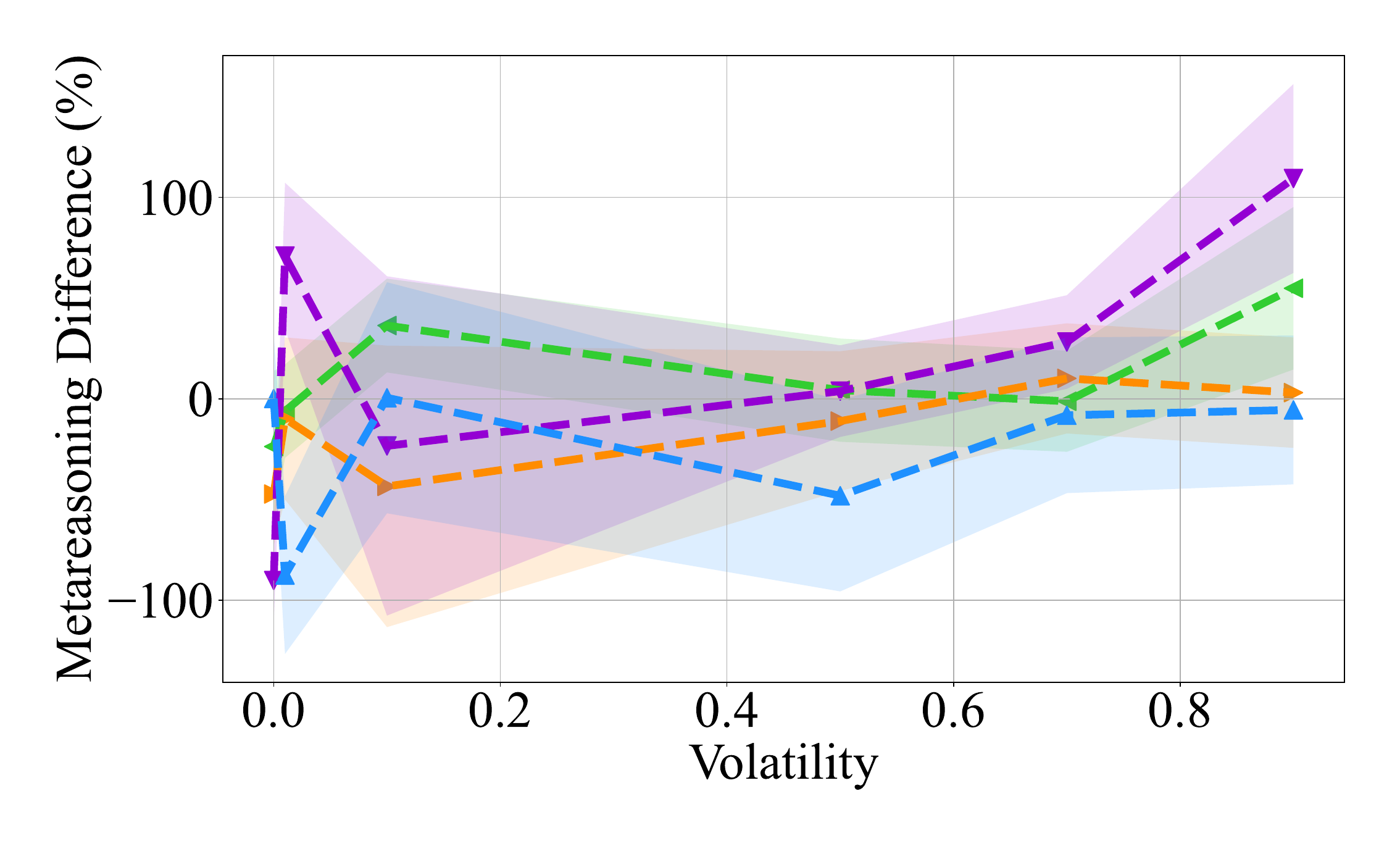}
    \caption{Planet percent difference execution time when using metareasoning as a function of volatility} \label{fig:vde}
  \end{subfigure}
\hspace*{\fill}   
  \begin{subfigure}{0.4\textwidth}
    \includegraphics[width=\linewidth]{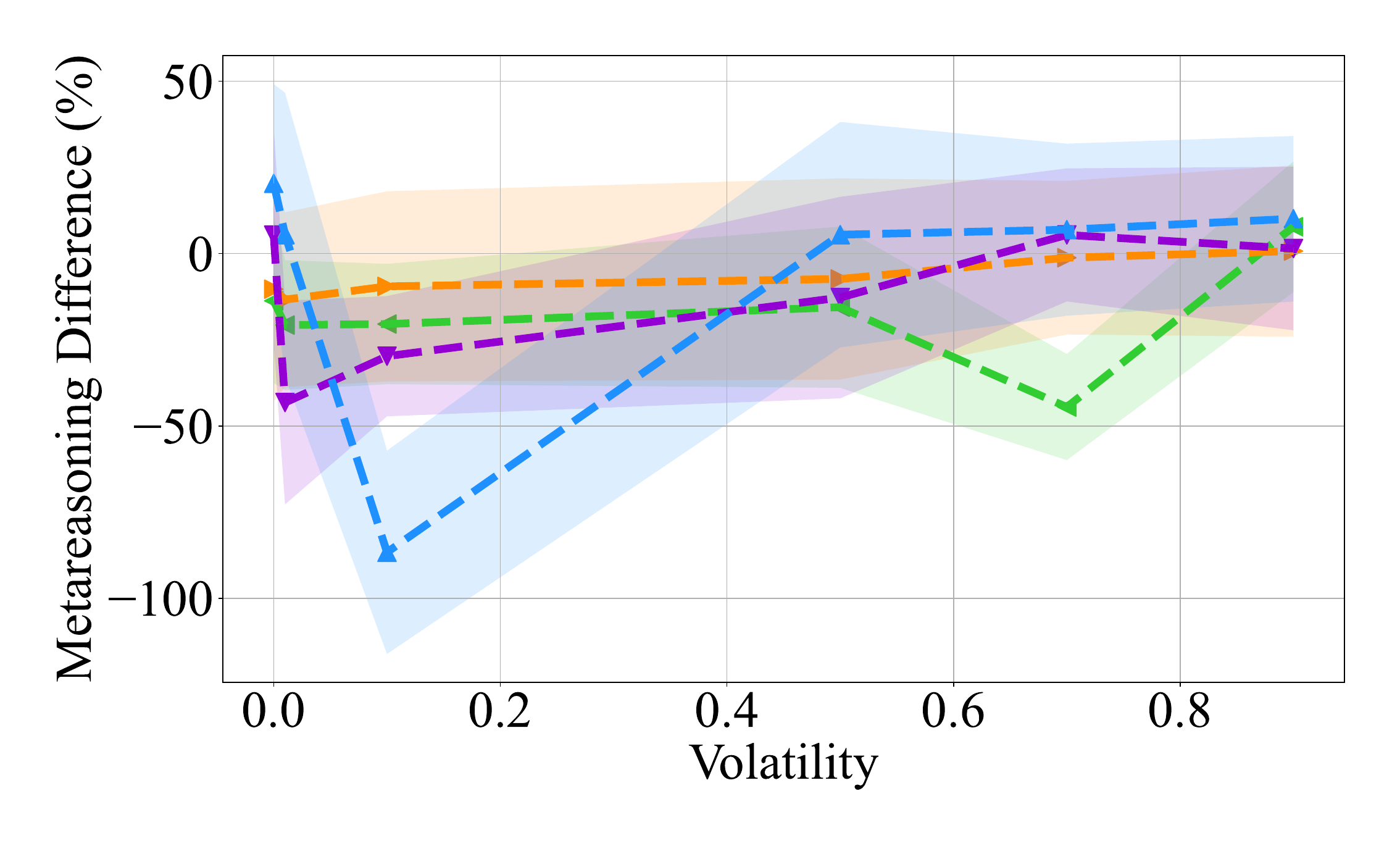}
    \caption{Walker percent difference execution time when using metareasoning as a function of volatility} \label{fig:vdf}
  \end{subfigure}%

\begin{center}
    \includegraphics[width=0.3\linewidth]{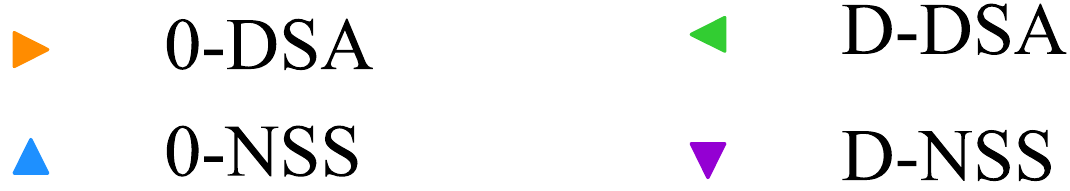}
\end{center}

\caption{Percent difference when using metareasoning for NSS and DSA variants for the Planet (left column) and Walker (right column) constellations with increasing volatility from Figure \ref{fig:volatility}. We report the percent difference in utility (top),  total message volume (middle), and per-agent runtime (bottom). A positive percent means the metareasoning variant achieved a higher value, while a negative percent corresponds to the metareasoning variant achieving a lower value. Shaded region shows standard deviation.}
 \label{fig:volatility-diff}
\end{figure*}

We evaluate the algorithms on realistic, large-scale, dynamic scenarios with thousands of requests. Figure \ref{fig:largeresults} shows the results for the Planet and Walker constellations. We report the total utility of each algorithm in Figures \ref{fig:1a} and \ref{fig:1b}, the total message volume in Figures \ref{fig:1c} and \ref{fig:1d}, and the average per-agent execution time in Figures \ref{fig:1e} and \ref{fig:1f}. Note the log scale in the figures for message volume and runtime. 

In terms of solution quality, D-NSS and D-DSA achieve close to the offline baseline utility. D-NSS outperforms both 0-NSS and 0-DSA as well as the greedy and random baselines. Crucially, D-NSS achieves high solution quality while having a lower total message volume and execution time compared to D-DSA, 0-DSA, and 0-NSS. Both D-NSS and 0-NSS use an order of magnitude fewer messages and computation times than their DSA counterparts. Since the lower bound algorithm is computed centrally, the message volume corresponds to the schedules a ground station would have to uplink.

We also evaluate the stability and convergence of D-NSS compared to D-DSA, 0-DSA, and 0-NSS. Figures \ref{fig:planet-iter} and \ref{fig:walker-iter} show the average solution quality over iterations of the algorithms during large problem instances with $v=5$. We fix the number of iterations for all algorithms to show the relative performance. Clearly, D-NSS and D-DSA are much more stable, exhibiting substantially less sensitivity to the dynamics, illustrated by the sharp drops in solution quality of 0-NSS and 0-DSA. Notably, D-NSS is very stable; even in the iteration directly after problem dynamics, D-NSS repairs solutions effectively to maintain or improve solution quality. This supports computing and repairing sub-problems rather than global solutions like D-DSA. In addition, the stability of D-NSS leads to a quick convergence in practice, which drives the efficiency of the algorithm.

\subsection{Evaluating metareasoning}
We explore aspects of the problem space, including the volatility of problem dynamics and the cost of planning. We evaluate how these parameters that define problem instances influence the value of metareasoning and dynamic algorithm variations. In the following sections, D-NSS-MR, D-DSA-MR, 0-NSS-MR, and 0-DSA-MR refer to the corresponding algorithm with a metareasoning oracle defined by the practical local metareasoning heuristic defined in Section \ref{sec:meta-heuristic}. 

\subsubsection{Effects of volatility}

We investigate how varying magnitudes of problem dynamics affect the return of metareasoning. For DCOSP instances, we define the volatility to be a real number in $\rho \in [0,1]$ that captures the ratio of requests that are added and removed during dynamic updates. A volatility of $0$ corresponds to a static problem, whereas a volatility of $1$ corresponds to a problem instance where an entirely new, disjoint request set is introduced at each time step. We evaluate each algorithm on $10$ randomly generated problem instances with a volatility of $0, 0.01, 0.1, 0.5, 0.7$, and $0.9$. Figure $\ref{fig:volatility}$ shows the results of these experiments. We report the utility achieved by each algorithm, the communication volume, and the per-agent runtime. In these experiments, we fix the cost of planning to be $300$ seconds, $p=8$, and $v=3$.

First, we notice that as the volatility increases, the from scratch and dynamic variations of NSS and DSA naturally converge, as seen in Figures \ref{fig:va} and \ref{fig:vb}. Disjoint request sets cause the dynamic variations to fail to repair previous solutions and instead resort to scheduling from scratch. However, when volatility is lower, the D-NSS and D-DSA variants achieve higher utility than 0-NSS and 0-DSA. These algorithms are also comparable in performance to the omniscient algorithm. 

We again see the orders of magnitude improvement of NSS variants over DSA in terms of message volume and runtime in Figures \ref{fig:vc}, \ref{fig:vd}, \ref{fig:ve}, and \ref{fig:vf}. We omit from Figures \ref{fig:vc} and \ref{fig:vd} the algorithms that do not require communication.

To highlight the value of metareasoning, Figure~\ref{fig:volatility-diff} reports the relative percent difference between the standard and metareasoning variants from Figure~\ref{fig:volatility}. We compute the relative percent difference using the midpoint of the two values as the denominator. We report the percent difference in utility, total message volume, and per-agent runtime. A positive percent means the metareasoning variant achieved a higher value, while a negative percent corresponds to the metareasoning variant achieving a lower value. In terms of utility, the metareasoning variants achieve nearly identical utility to their non-metareasoning counterparts as seen in Figures \ref{fig:vda} and \ref{fig:vdb}. However, we see that the metareasoning variants are able to achieve this utility while utilizing significantly lower message volume. When problem volatility is low, the metareasoning variants use between $20\%$ and $70 \%$ fewer messages. As problem volatility increases, metareasoning has diminishing returns. However, even at $0.9$ volatility, D-DSA-MR uses over $20\%$ fewer messages for the Planet constellation. In terms of runtime, we again see that metareasoning primarily reduces the amount of computation time per-agent. However, when volatility is high and replanning is always desirable, we see that there is some added computation time for D-NSS and D-DSA variants due to problem repair and metareasoning. 

Overall, the results show the value of metareasoning with respect to volatility; metareasoning has substantial benefits when problem volatility is low and diminishing returns as volatility increases. Metareasoning achieves similar utility as non-metareasoning variants while utilizing fewer computational resources. This behavior is consistent with the criterion developed in Section \ref{sec:meta}. Agents only expend resources if replanning is estimated to improve the current solution. The results show that the estimation of the scheduling cost is effective in practice and metareasoning can save key resources.

\subsubsection{Effects of scheduling costs}

\begin{figure*}[h!]
   
  \begin{subfigure}{0.4\textwidth}
    \includegraphics[width=\linewidth]{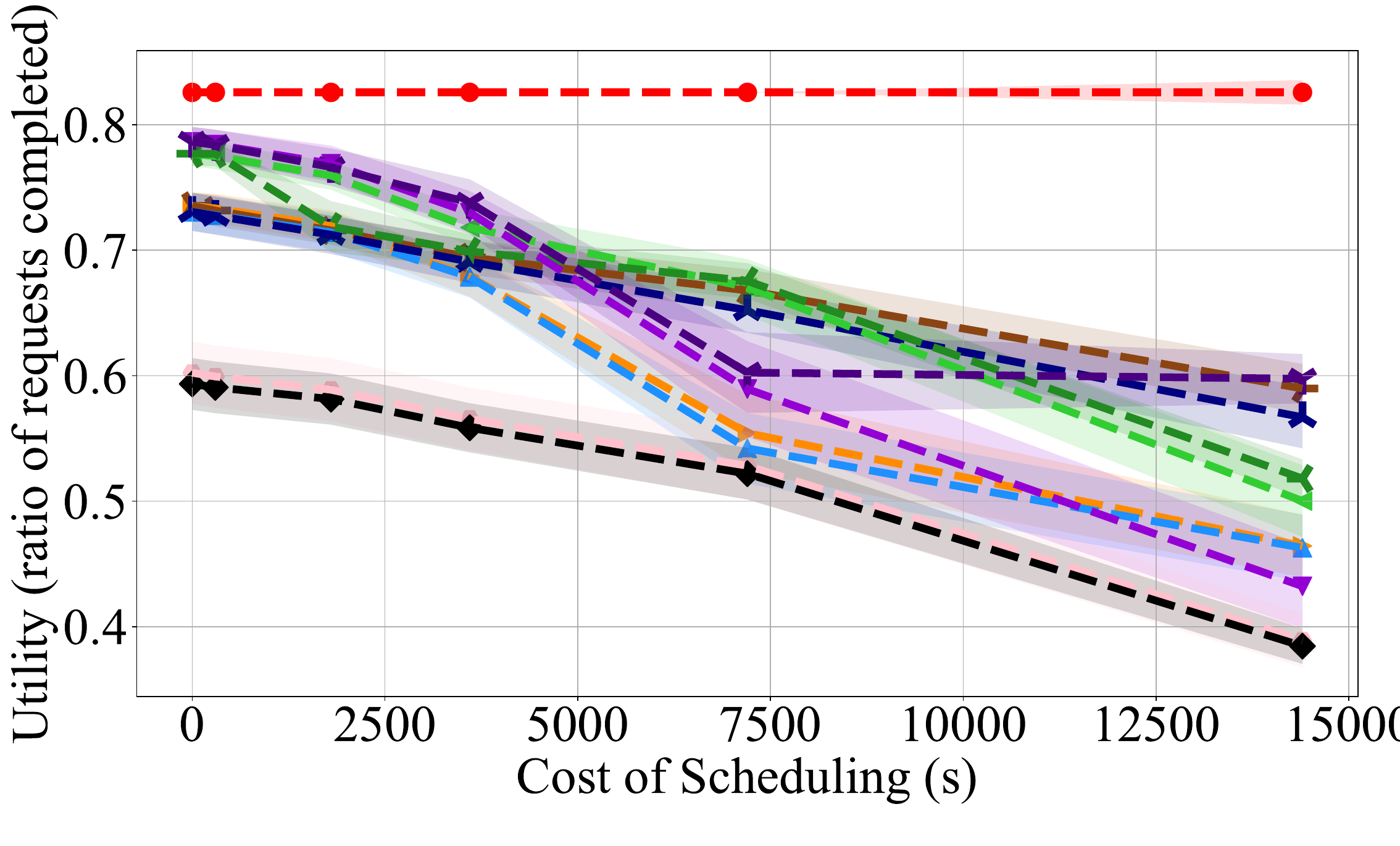}
    \caption{Planet satisfaction ratio as a function of scheduling cost.} \label{fig:ca}
  \end{subfigure}%
  \hspace*{\fill}   
  \begin{subfigure}{0.4\textwidth}
    \includegraphics[width=\linewidth]{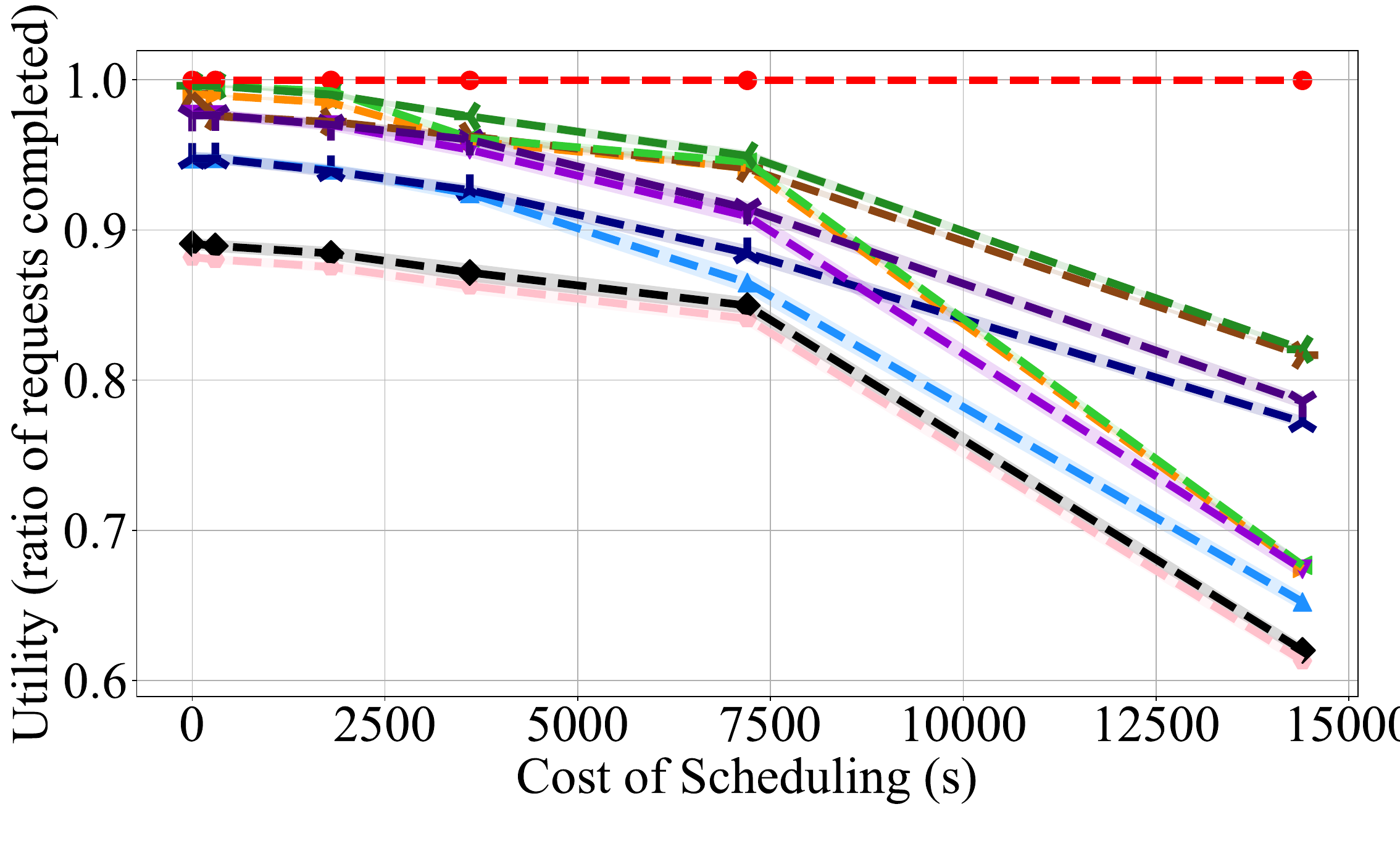}
    \caption{Walker satisfaction ratio as a function of scheduling cost.} \label{fig:cb}
  \end{subfigure}%

  \begin{subfigure}{0.4\textwidth}
    \includegraphics[width=\linewidth]{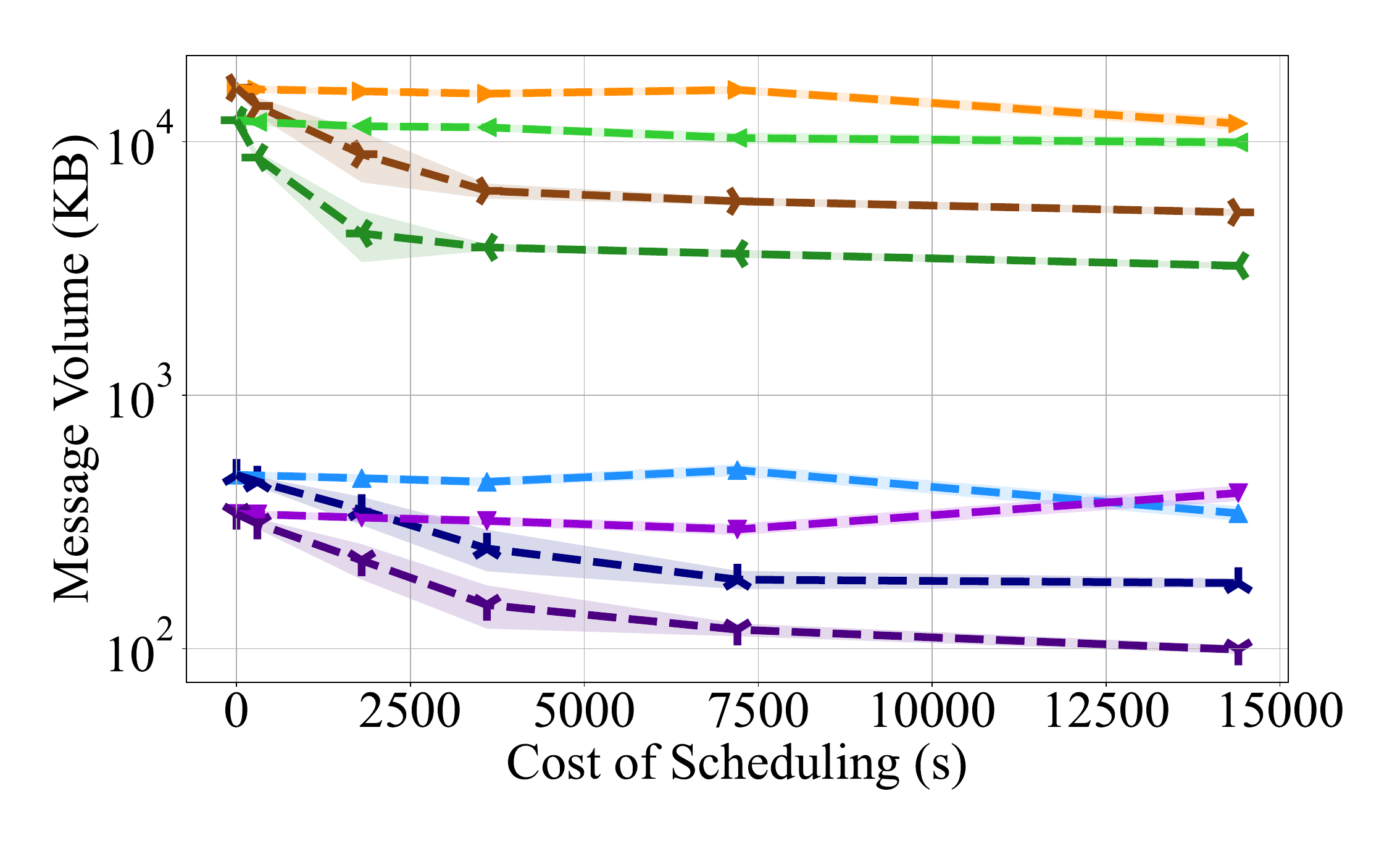}
    \caption{Planet message volume as a function of scheduling cost.} \label{fig:cc}
  \end{subfigure}
\hspace*{\fill}   
  \begin{subfigure}{0.4\textwidth}
    \includegraphics[width=\linewidth]{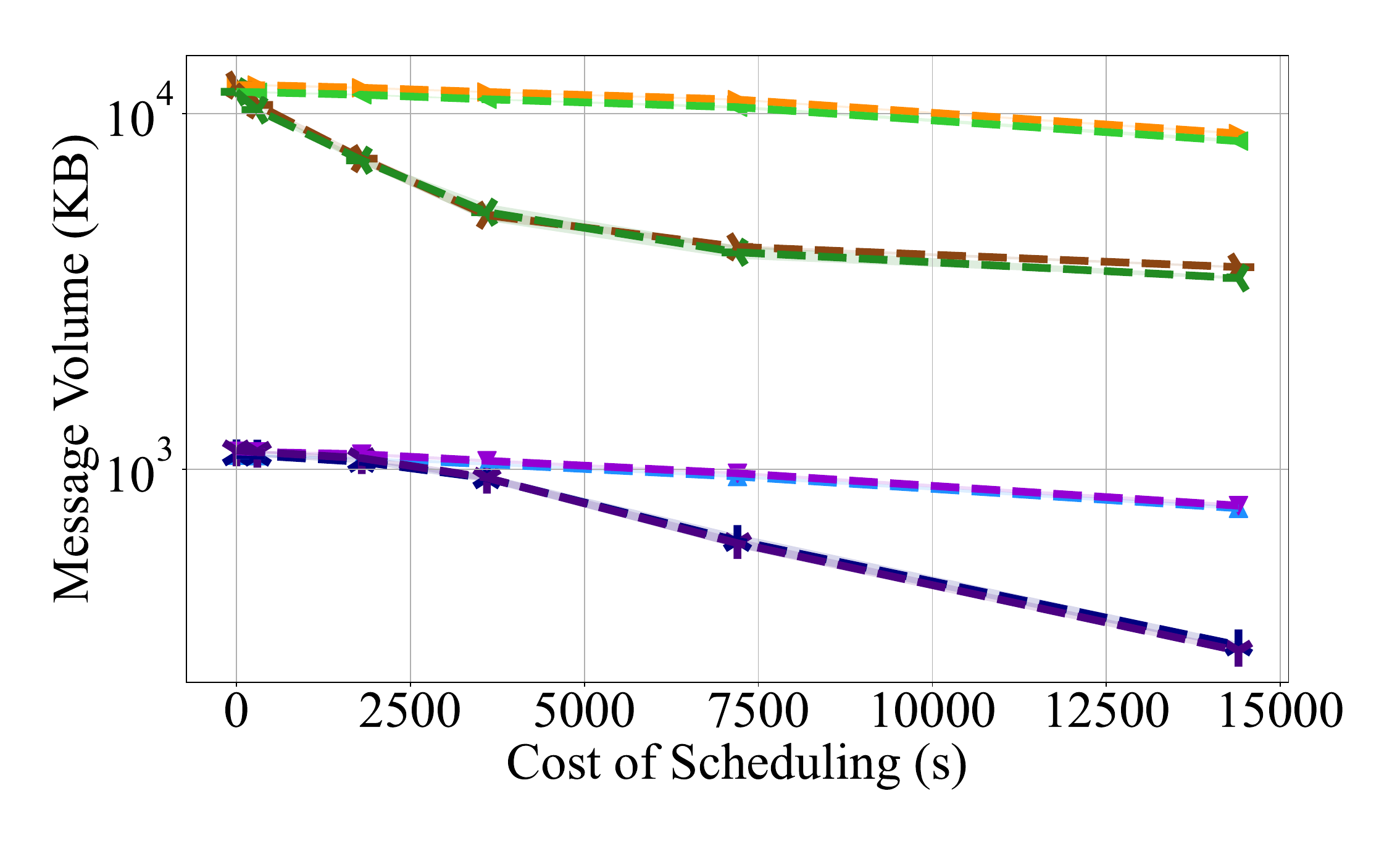}
    \caption{Walker message volume as a function of scheduling cost.} \label{fig:cd}
  \end{subfigure}%

  \begin{subfigure}{0.4\textwidth}
    \includegraphics[width=\linewidth]{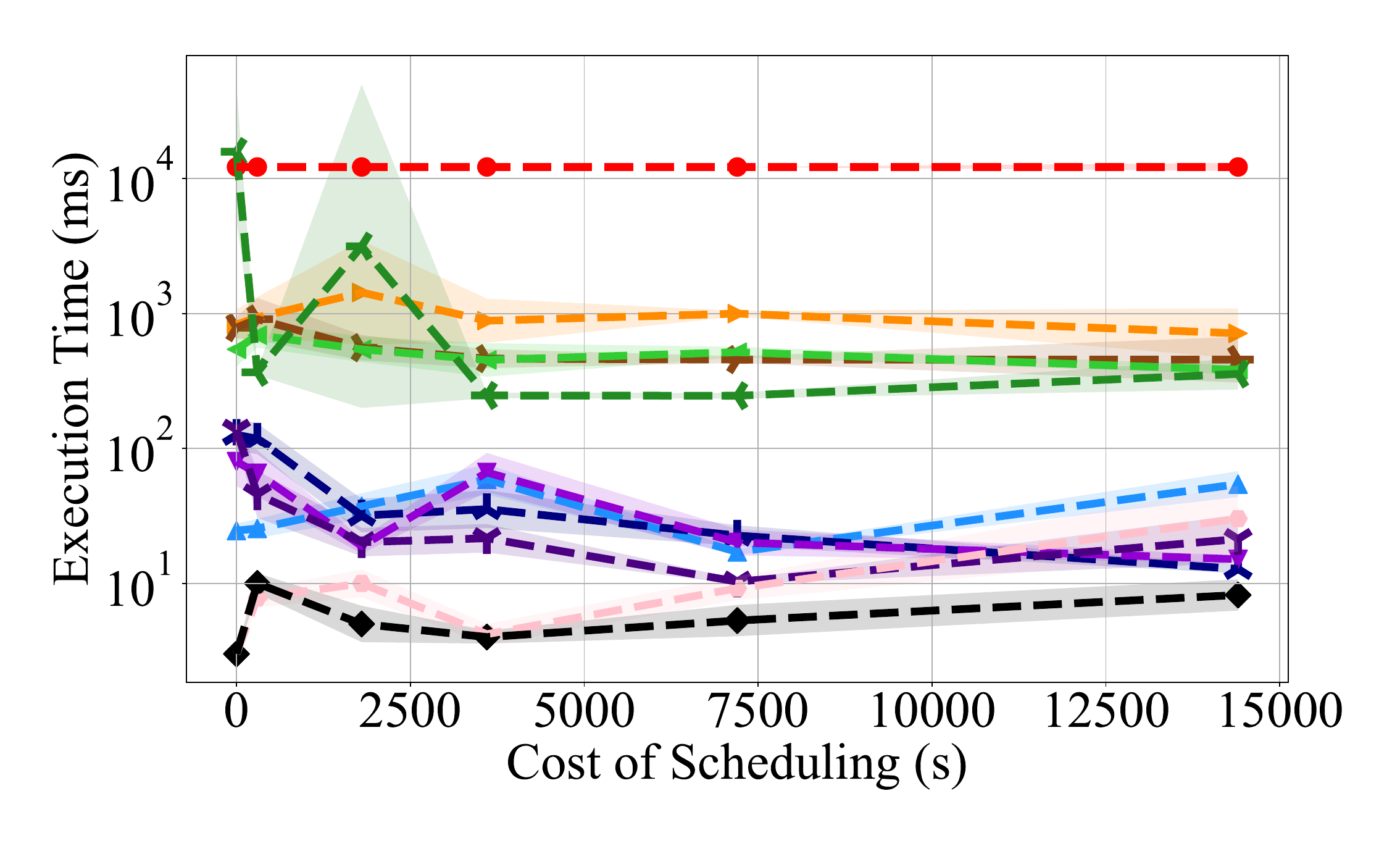}
    \caption{Planet execution time as a function of scheduling cost.} \label{fig:ce}
  \end{subfigure}
\hspace*{\fill}   
  \begin{subfigure}{0.4\textwidth}
    \includegraphics[width=\linewidth]{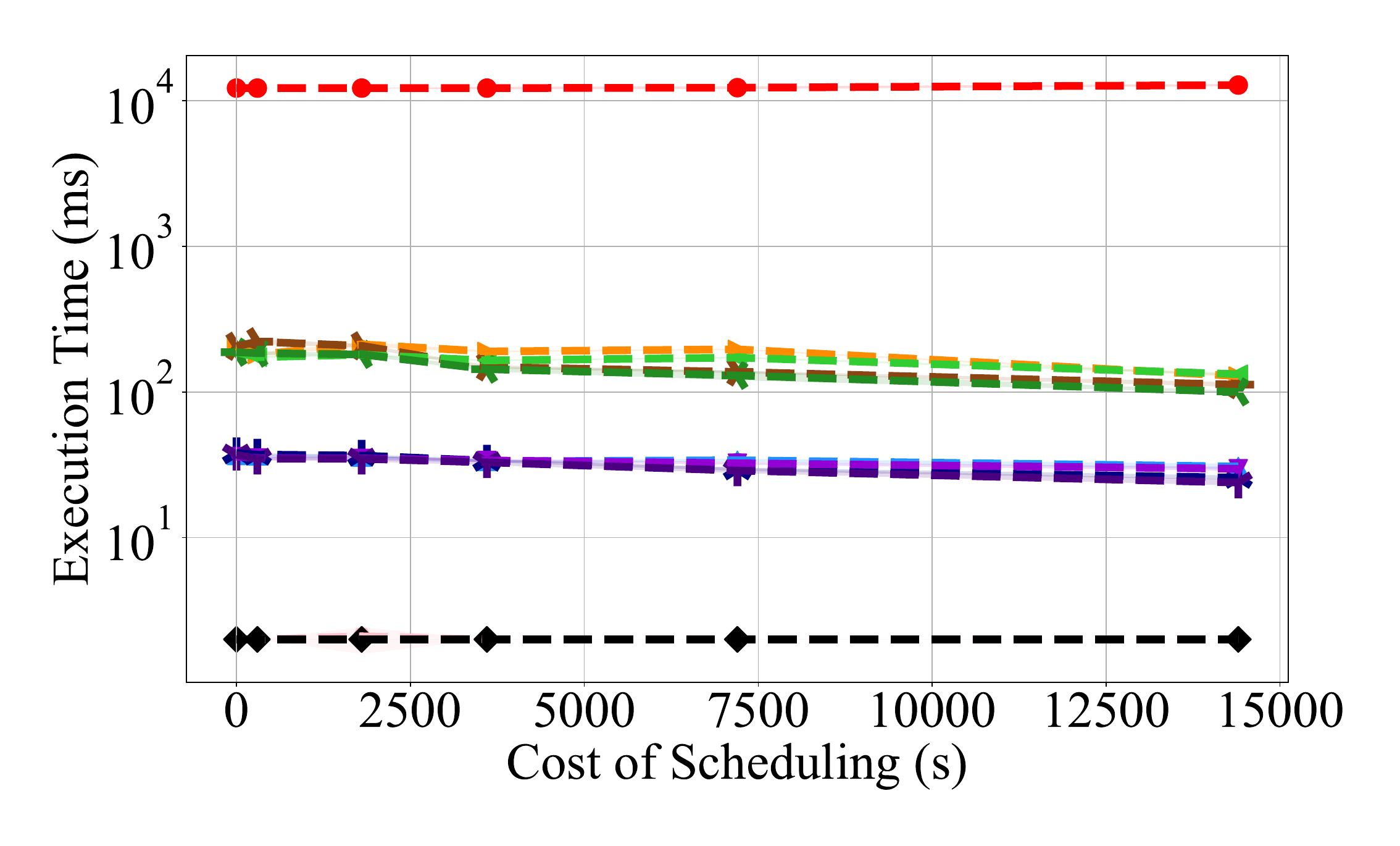}
    \caption{Walker execution time as a function of scheduling cost.} \label{fig:cf}
  \end{subfigure}%

  \begin{center}
    \includegraphics[width=0.4\linewidth]{figures/mr-legend.pdf}
\end{center}

\caption{Results of large-scale simulations for the Planet (left column) and Walker (right column) constellations with increasing scheduling cost. We report the ratio of request satisfaction, total message volume, and per-agent runtime. Note the log scale for message volume and runtime. For NSS and DSA variants, we evaluate each with and without metareasoning. Shaded region shows standard deviation.}
 \label{fig:sched-cost}
\end{figure*}

We explore how increasing the computational scheduling time impacts the efficacy of metareasoning. While the per-agent runtime of algorithms on the ground may be on the order of seconds, when run on a spacecraft, the time to execute scheduling may increase drastically. There are a number of factors that influence this runtime, including hardware, contention among other processes, and communication latencies. While investigating the impact of these factors is of interest, it is beyond the scope of this work. Previous benchmarking has shown a range of scaling factors between execution time on ground-based systems and flight hardware \citep{dunkel2023benchmarking}. These factors range from constant to orders of magnitude slower and depend heavily on the processors used. To simulate this, we vary a fixed interval of time that upper bounds how long planning will take (\textit{e.g.}, 30 seconds). We evaluate each algorithm on $10$ randomly generated problem instances with a planning time of $0, 5, 30, 60, 120$, and $240$ minutes. While a planning time of $30$ minutes is already large, it is plausible on constrained flight hardware for sufficiently expensive scheduling procedures. In contrast, a planning time of $4$ hours would not be practical, yet we include these higher costs for both completeness and theoretical analysis; a planning time of $4$ hours represents a cost equivalent to the request interval, $h(r)$. While $h(r)$ is large for DCOSP instances, in other domains the request interval may be much shorter, therefore it is of interest to examine these values. 
We report the utility achieved by each algorithm, the communication volume, and the per-agent runtime in Figure \ref{fig:sched-cost}. For these experiments, we fix the volatility at $0.1$, $p=8$, and $v=3$. Due to its centralized execution, the omniscient algorithm does not incur any scheduling cost and upper bounds utility. 

Figures \ref{fig:ca} and \ref{fig:cb} illustrate how as the cost of scheduling increases, the utility achieved by the system decreases. This demonstrates that scheduling costs influence agent resources and can even impact the utility of algorithms. 

We again see that D-NSS and D-DSA achieve higher utility than 0-NSS and 0-DSA respectively while expending less time and messages. We also find that NSS variations utilize orders of magnitude less runtime and fewer messages compared to DSA variants, and that D-NSS is among the highest-performing algorithms. These results are consistent with the results of the prior sections.

\begin{figure*}[htbp!]
   
  \begin{subfigure}{0.4\textwidth}
    \includegraphics[width=\linewidth]{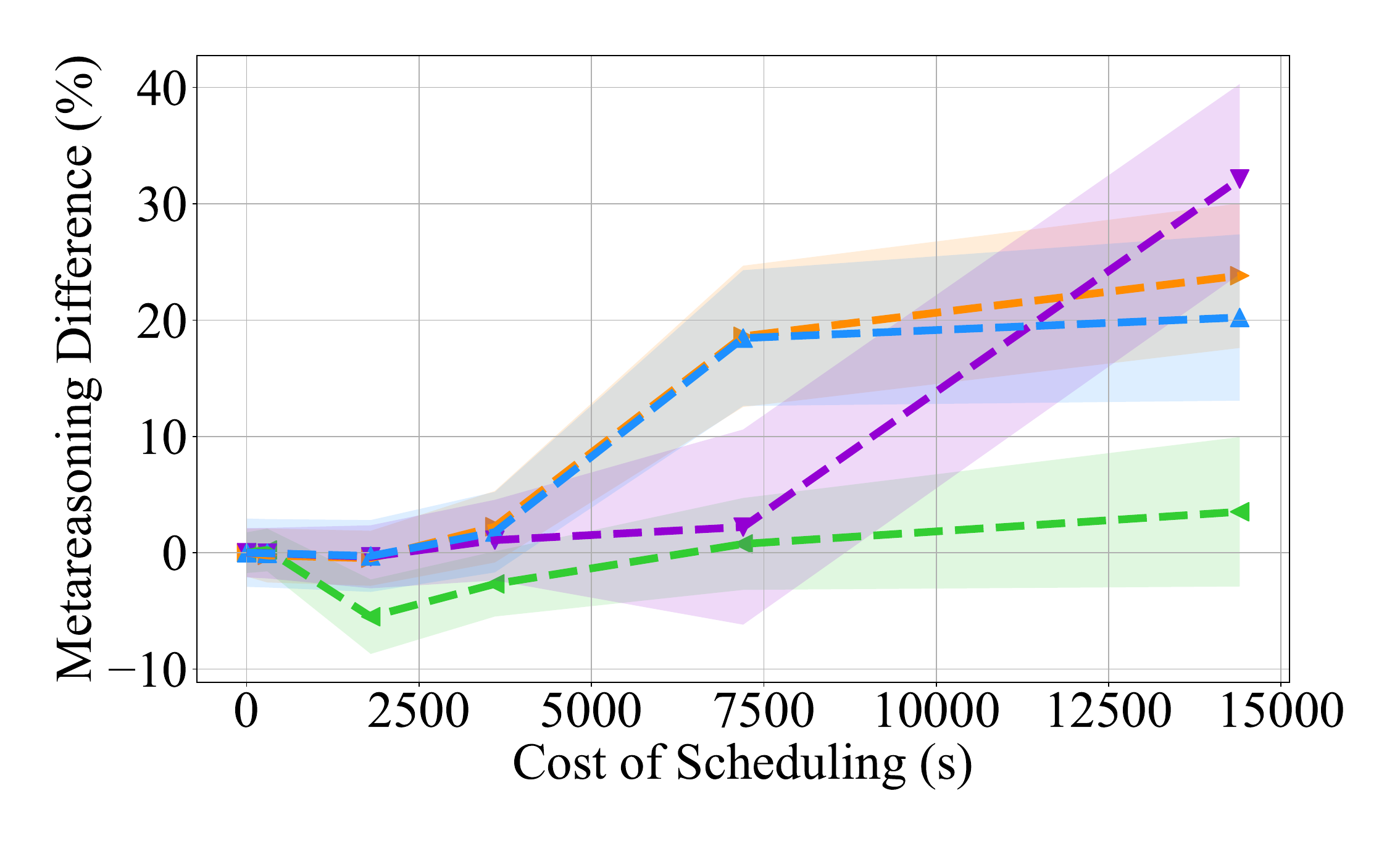}
    \caption{Planet percent difference utility when using metareasoning as a function of scheduling cost.} \label{fig:cda}
  \end{subfigure}%
  \hspace*{\fill}   
  \begin{subfigure}{0.4\textwidth}
    \includegraphics[width=\linewidth]{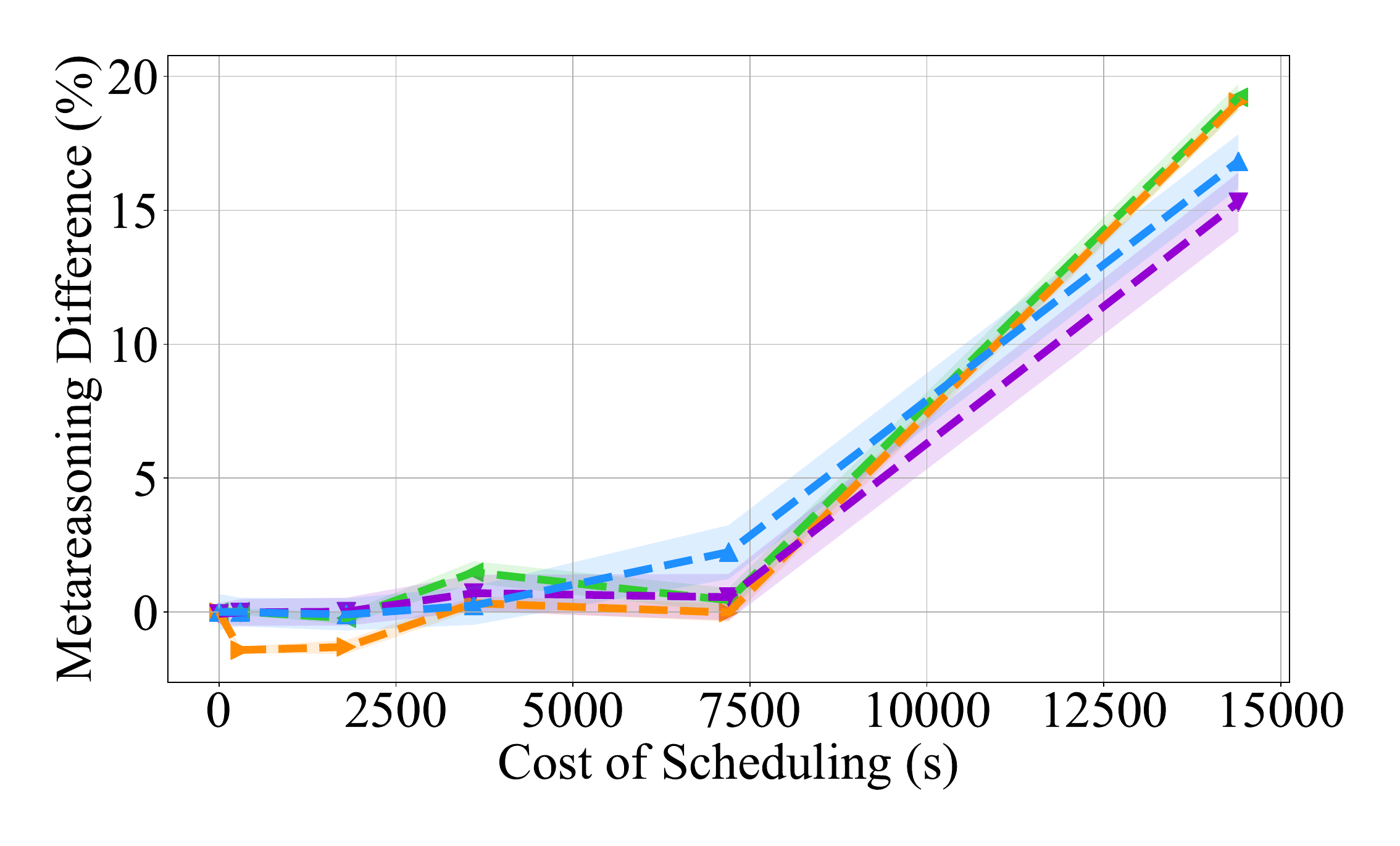}
    \caption{Walker percent difference utility when using metareasoning as a function of scheduling cost.} \label{fig:cdb}
  \end{subfigure}%

  \begin{subfigure}{0.4\textwidth}
    \includegraphics[width=\linewidth]{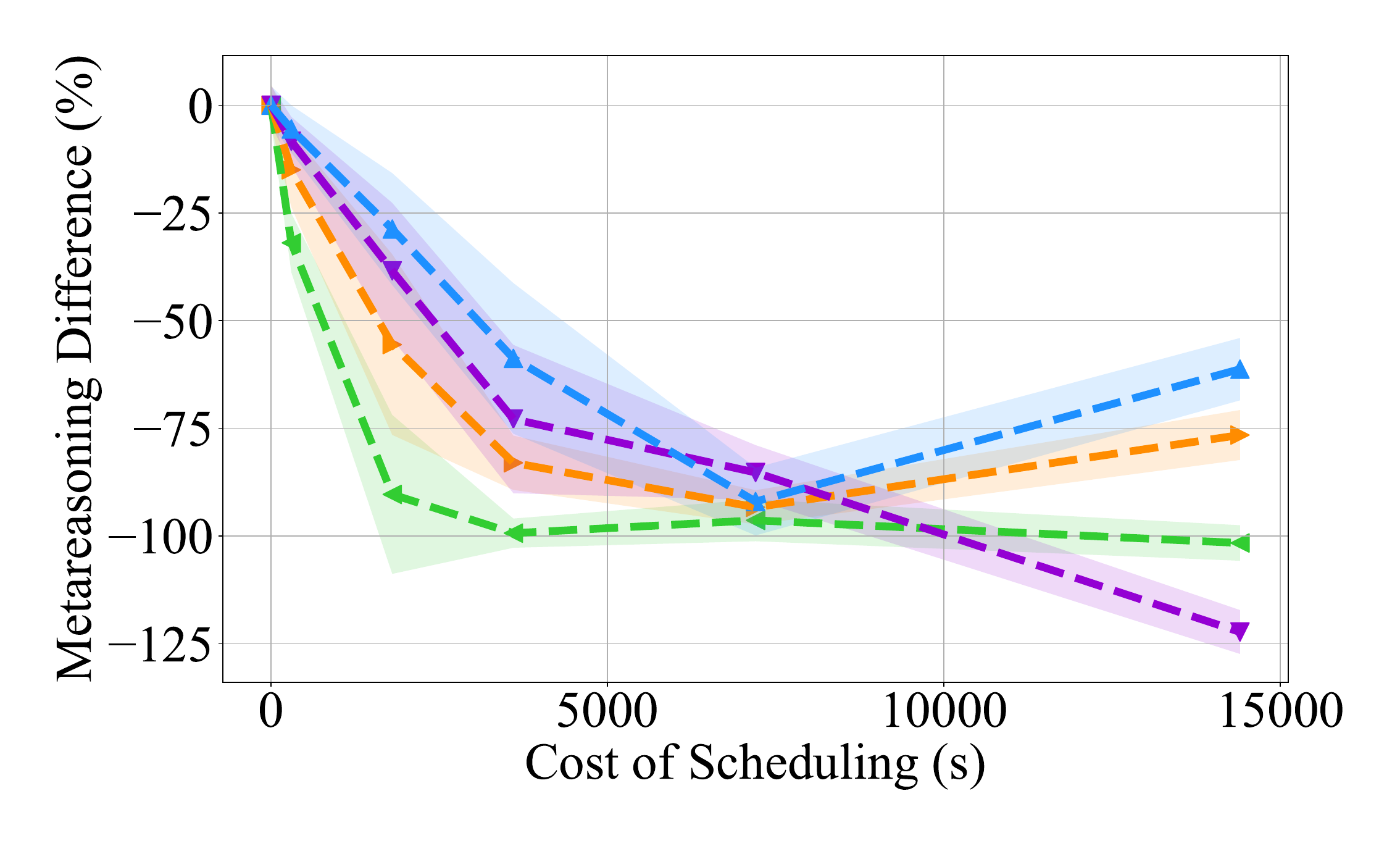}
    \caption{Planet percent difference message volume when using metareasoning as a function of scheduling cost} \label{fig:cdc}
  \end{subfigure}
\hspace*{\fill}   
  \begin{subfigure}{0.4\textwidth}
    \includegraphics[width=\linewidth]{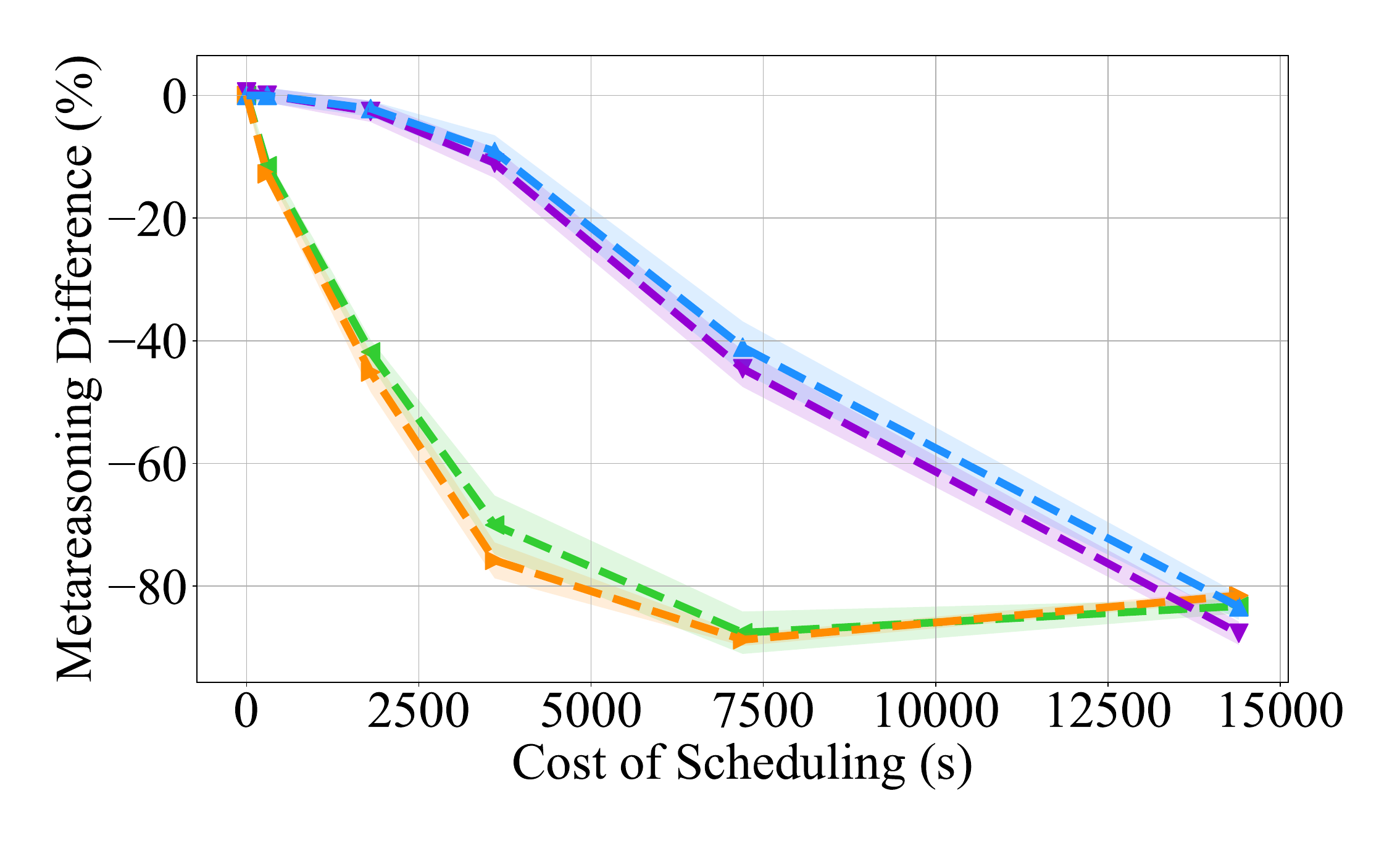}
    \caption{Walker percent difference message volume when using metareasoning as a function of scheduling cost} \label{fig:cdd}
  \end{subfigure}%

  \begin{subfigure}{0.4\textwidth}
    \includegraphics[width=\linewidth]{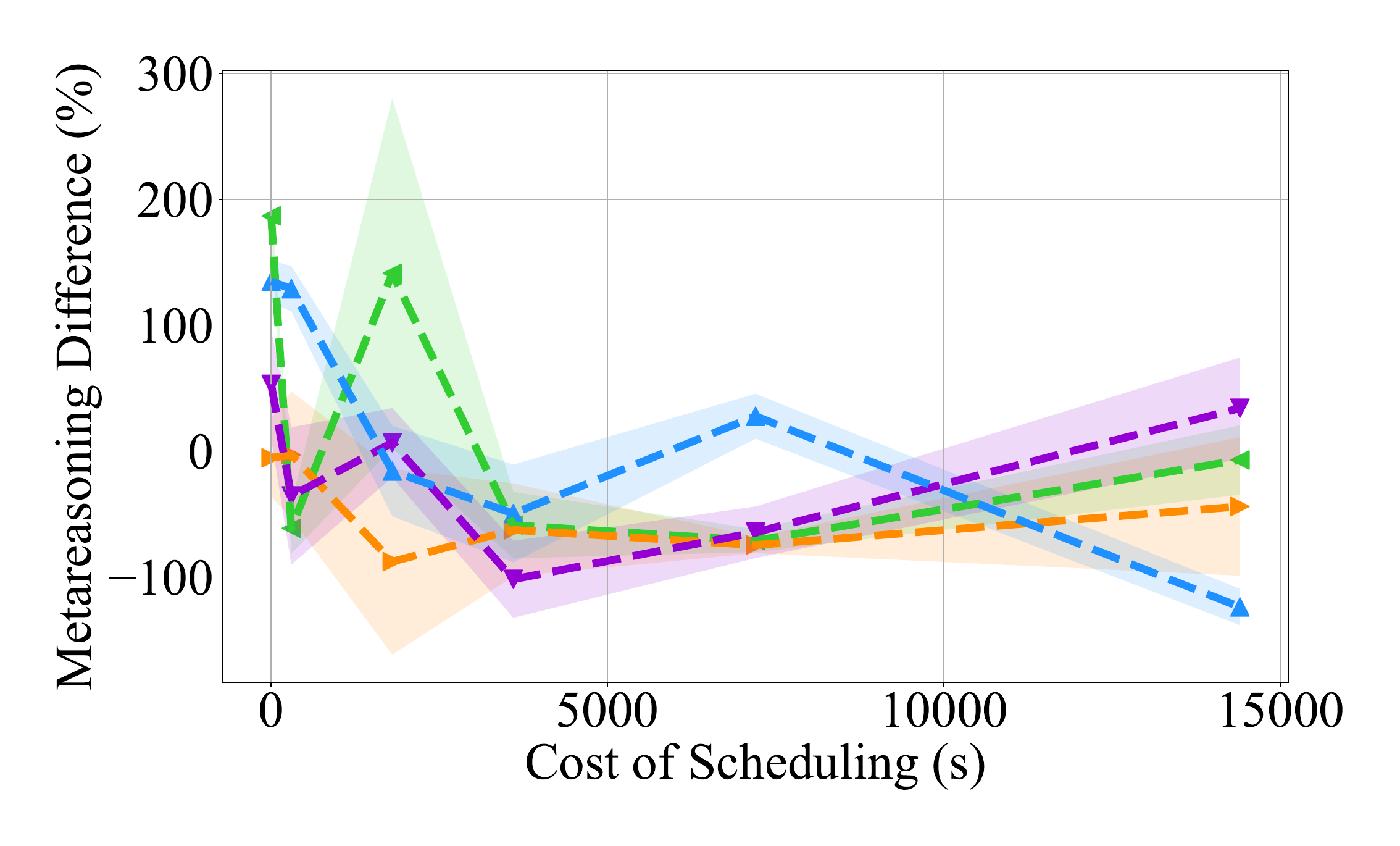}
    \caption{Planet percent difference execution time when using metareasoning as a function of scheduling cost} \label{fig:cde}
  \end{subfigure}
\hspace*{\fill}   
  \begin{subfigure}{0.4\textwidth}
    \includegraphics[width=\linewidth]{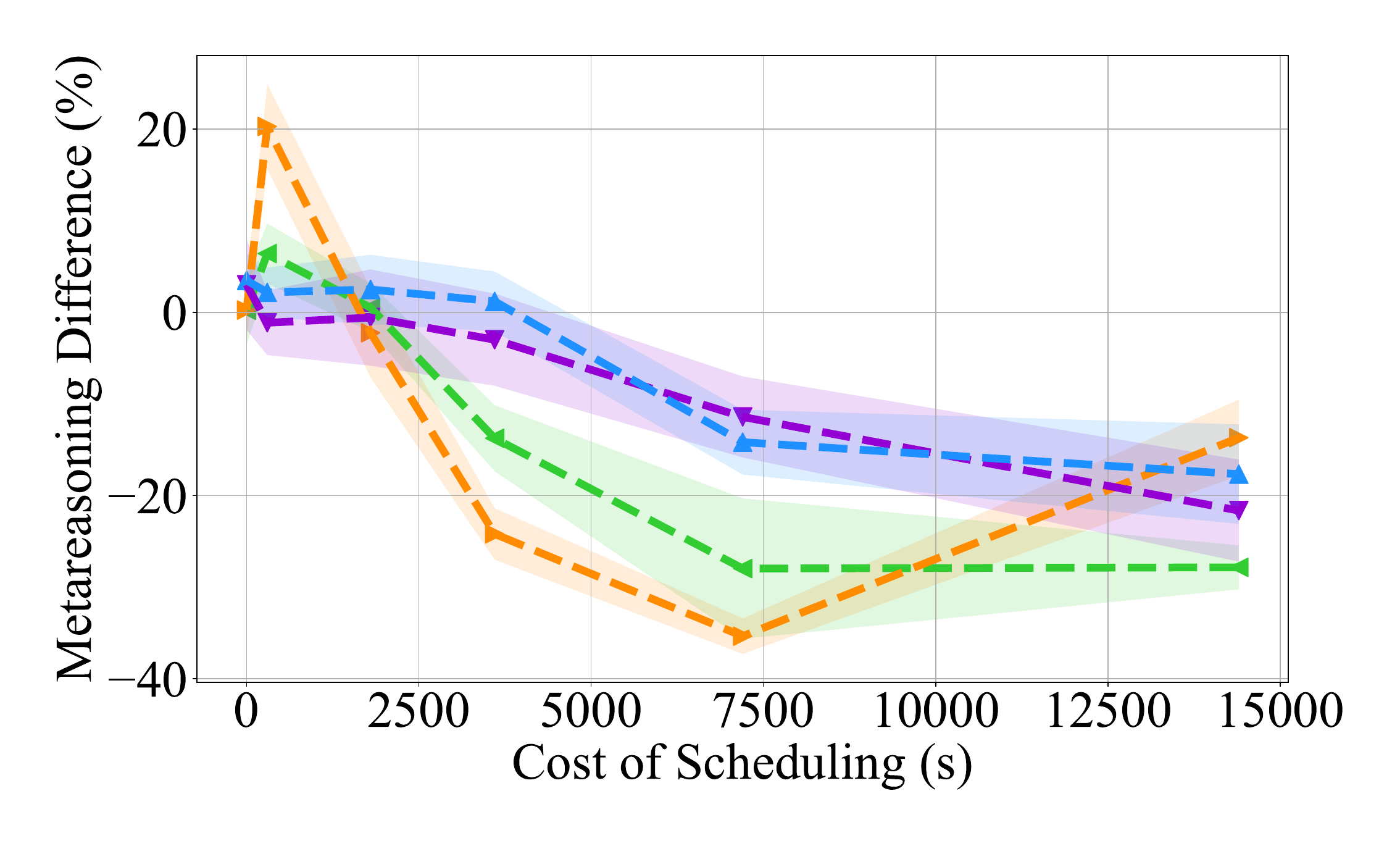}
    \caption{Walker percent difference execution time when using metareasoning as a function of scheduling cost} \label{fig:cdf}
  \end{subfigure}%

\begin{center}
    \includegraphics[width=0.3\linewidth]{figures/mr-diff-legend.pdf}
\end{center}

\caption{Percent difference when using metareasoning for NSS and DSA variants for the Planet (left column) and Walker (right column) constellations with increasing scheduling cost from Figure \ref{fig:sched-cost}. We report the percent difference in utility (top),  total message volume (middle), and per-agent runtime (bottom). A positive percent means the metareasoning variant achieved a higher value, while a negative percent corresponds to the metareasoning variant achieving a lower value. Shaded region shows standard deviation.}
 \label{fig:sched-cost-diff}
\end{figure*}

We plot the percent difference between the standard and metareasoning variants from Figure \ref{fig:sched-cost} in Figure \ref{fig:sched-cost-diff} to highlight the effect of metareasoning.
As the scheduling cost decreases, metareasoning has a diminishing influence. Both the utility achieved and the message volume converge for metareasoning and non-metareasoning algorithms when the cost of scheduling goes to $0$. This is highlighted in Figures \ref{fig:cda}, \ref{fig:cdb},  \ref{fig:cdc}, and \ref{fig:cdd}. In terms of runtime, when the cost of scheduling is low, metareasoning slightly increases per-agent runtime, as shown in Figures \ref{fig:cde} and \ref{fig:cdf}. However, as the cost of scheduling grows, we see the value of metareasoning. Not only do metareasoning variants expend fewer computational resources, but they also achieve higher utility when the cost of scheduling becomes large. In the best case, D-NSS-MR achieves around $30\%$ more utility than D-NSS while the percent difference in messages is $-120\%$. The runtime results also confirm a general decrease in computation time per-agent as the cost of scheduling grows. 

Intuitively, when the cost of scheduling becomes comparable to a request interval, then scheduling may block agents from completing tasks. Metareasoning allows agents to identify these scenarios, and preserve both computational resources and utility. 

As the cost of scheduling grows, metareasoning becomes increasingly powerful. Metareasoning not only drastically reduces the message volume and computation time, but can improve utility when computation is costly.  Our approximation of optimal metareasoning is highly effective. It achieves equal, or better, utility than non-metareasoning while saving resources.

\section{Conclusions and future work}

This work extends the DDCOP framework to better model and solve continual optimization problems in which there is overlapping optimization and execution, resources persist across time, and computation incurs a cost. We introduced an execution-aware formulation of DDCOPs through DCOSP, established a general framework for metareasoning within DDCOPs, and developed D-NSS, a scalable DDCOP algorithm. 

Our empirical evaluation demonstrated that D-NSS computes high-quality solutions while substantially reducing computation time and communication compared to baselines. Although motivated by satellite scheduling, D-NSS is a general decomposition-based algorithm that can incorporate alternative decomposition heuristics, including learned or adaptive ones, enabling broader applicability across DDCOP applications.

Beyond introducing a new algorithm, we formulated metareasoning as a general component of the DDCOP framework. By explicitly reasoning about the value of replanning, agents can balance estimated solution improvements against computational and communication costs. Our experiments demonstrated that metareasoning substantially reduces resource expenditure while maintaining---and in some settings improving---overall solution quality. We believe this framework will be valuable for a broad range of computationally constrained multi-agent systems, including emerging agentic AI systems~\citep{Picard26:Agentic}.

Future work includes extending execution-aware DDCOPs to proactively reason about future dynamics, such as through the \textit{proactive DCOP (PDDCOP)} model~\citep{pddcop}. Within DCOSP, we can utilize prior distributions over problem dynamics to generate robust offline schedules and construct or learn heuristics that prioritize certain tasks that are more likely to yield utility when executed. Another promising direction for future work is developing new variations of common DDCOP algorithms that are compatible with disjoint decisions from agents using metareasoning.

Our methods are being deployed as part of \textit{NASA’s Federated Autonomous Measurement} (FAME) mission~\citep{fame-spaceops-2025,zilberstein-spaceops-2025}. FAME is the largest demonstration of distributed multi-agent autonomy in space,  consisting of more than 60 Earth-observing spacecraft that will coordinate their measurements to optimize observation completion across dynamic scenarios. Beyond this application, we believe the ideas developed in this paper provide a foundation for future work on continual dynamic distributed optimization across a broad range of decentralized AI systems.

\section*{Acknowledgments}
Portions of this research were carried out at the Jet Propulsion Laboratory, California Institute of Technology, under a contract with the National Aeronautics and Space Administration (80NM0018D0004). Government sponsorship acknowledged. Itai Zilberstein is also supported by the NSF Graduate Research Fellowship Program under grant DGE2140739. Any opinions, findings, and conclusions or recommendations expressed in this material are those of the author(s) and do not necessarily reflect the views of the funding agencies. 

\bibliography{bib}

\end{document}